\documentclass[journal]{IEEEtran}
\usepackage{graphicx}

\usepackage{makecell,multirow,diagbox}
\usepackage{cite}
\usepackage{color} 
\usepackage{adjustbox}
\usepackage{url}
\usepackage{arydshln}

\newcommand{\ie}{{\em i.e.}}
\newcommand{\eg}{{\em e.g.}}
\newcommand{\etc}{{\em etc}}

\begin{document}
\title{Computational Imaging and Artificial Intelligence: The Next Revolution of Mobile Vision}
\author{Jinli~Suo,
        Weihang~Zhang,
        Jin~Gong,
        Xin~Yuan,~\IEEEmembership{Senior~Member,~IEEE,}
        David J.~Brady,~\IEEEmembership{Fellow,~IEEE,}
        and~Qionghai~Dai,~\IEEEmembership{Senior~Member,~IEEE}%
\thanks{\noindent{\bf Jinli Suo, Weihang Zhang, Jin Gong} and {\bf Qionghai Dai} are with the Department of Automation, and the Institute for Brain and Cognitive Sciences, Tsinghua University, Beijing 100084, China.
E-mails: jlsuo@tsinghua.edu.cn, \{zwh19, gongj20\}@mails.tsinghua.edu.cn, qhdai@tsinghua.edu.cn.}%
\thanks{\noindent {\bf Xin Yuan} is with Bell Labs, Murray Hill, NJ 07974, USA.
E-mail: xyuan@bell-labs.com.}%
\thanks{\noindent {\bf David J. Brady} is with Wyant College of Optical Sciences, University of Arizona, Tucson, AZ 85721, USA. E-mail:  djbrady@arizona.edu.}
\thanks{{\em Corresponding authors: Xin Yuan and Qionghai Dai.}}
}
\maketitle

\begin{abstract}
Signal capture stands in the forefront to perceive and understand the environment and thus imaging plays the pivotal role in mobile vision. Recent explosive progresses in Artificial Intelligence (AI) have shown great potential to develop advanced mobile platforms with new imaging devices. Traditional imaging systems based on the "capturing images first and processing afterwards" mechanism cannot meet this unprecedented demand. {Differently, Computational Imaging (CI)} systems are designed to capture high-dimensional data in an encoded manner to provide more information for mobile vision systems.
Thanks to AI, CI can now be used in real systems by integrating deep learning algorithms into the mobile vision platform to achieve the closed loop of intelligent acquisition, processing and decision making, thus leading to the next revolution of mobile vision.
Starting from the history of mobile vision using digital cameras, this work first introduces the advances of CI in diverse applications and then conducts a comprehensive review of current research topics combining CI and AI. Motivated by the fact that most existing studies only loosely connect CI and AI (usually using AI to improve the performance of CI and only limited works have deeply connected them), in this work, we propose a framework to deeply integrate CI and AI by using the example of self-driving vehicles with high-speed communication, edge computing and traffic planning. 
Finally, we outlook the future of CI plus AI  by investigating new materials, brain science and new computing techniques to shed light on new directions of mobile vision systems.
\end{abstract}


\section{Introduction}
\label{sec:introduction}
Significant changes have taken place in mobile vision during the past few decades.
Inspired by the recent advances of Artificial Intelligence (AI) and the emerging field of Computational Imaging (CI), we are heading to a new era of mobile vision, where we anticipate to see a deep integration of CI and AI.
New mobile vision devices equipped with intelligent imaging systems will be developed and widely deployed in our daily life.

The history of mobile vision using digital cameras~\cite{Boyle_Smith_CCD} can be traced back to 1990s, when the digital event data recorders {(EDRs) have started to be installed} on vehicles~\cite{Chidester1999Recording}.

Interestingly, almost during the same time, cellphones with built-in cameras were invented~\cite{firstcamphoneweb}. 
Following this, the laptop started to embed cameras inside~\cite{Dalrymple2006Apple}.
{By then, mobile vision} started emerging and the digital data became exploding, which paved the way of us coming to the ``big data" era.
EDR, computer and cellphone are the three pillars of mobile vision that plays in the battlefront to capture digital data. 
We term this as the {\bf first revolution}, \ie, the inception of modern mobile vision. 
In this generation, the mode of mobile vision was to {\em capture and save on device}. Apparently, the main goal was to capture data.

Armed with the internet and mobile communication, mobile vision is developing rapidly and revolutionizing communications and commerce. Representative applications include web conferencing, online shopping, social network, \etc. We term this as the {\bf second revolution} of mobile vision, where {the chain} of capture, encoding, transmission and decoding is complete.  
This has been further advanced by the wide deployment of the fourth generation (4G) of broadband cellular network technology starting in 2009~\cite{chen2014the}. The mode of mobile vision has then been changed to {{\em capture, transmit and share images}}.
Meanwhile, new mobile devices such as tablets and wearable devices have also been invented.
Based on these, extensive social media have been developed and advanced. Sharing becomes the main role in this generation of mobile vision.

\begin{table*}[t]\centering
\small
\caption{Three revolutions of mobile vision systems.}
\label{Tab:3revolution}
 \resizebox{1\textwidth}{!}
 {
\begin{tabular}{c|l|c}
\hline
\rule{0pt}{10pt}{ }&\makecell[c]{\textbf{Figures}}& \makecell[c]{\textbf{Events}} \\
\hline

\rule{0pt}{45pt}\makecell[c]{\textbf{First revolution:} \\\texttt{capture}, save on device}& \makecell[c]{\includegraphics[height=55pt]{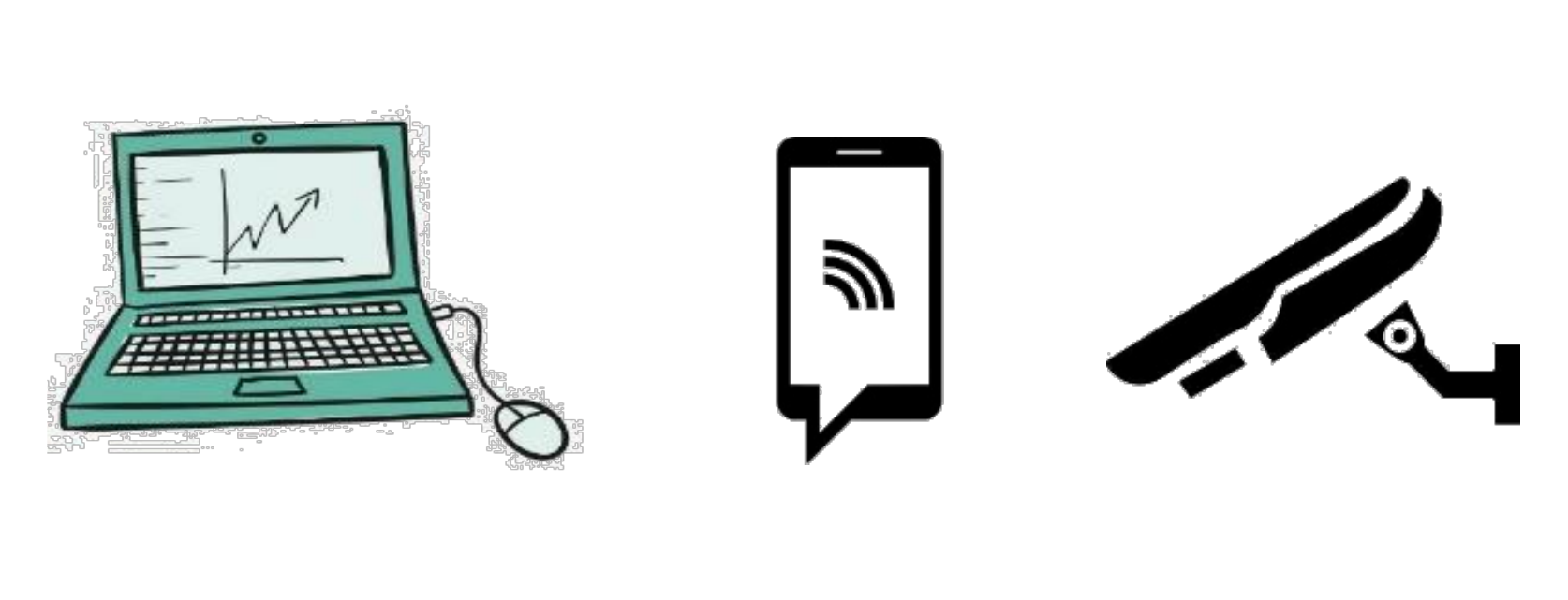}}& \makecell[l]{1990s\\ $\bullet$ Popularization of digital event data recorders \\1999-2000\\ $\bullet$ Invention of cell phones with built-in camera \\ \quad (Kyocera/Samsung SCH-V200/Sharp J-SH04)\\2006\\$\bullet$ Invention of laptop with built-in webcam \\ \quad (Apple MacBook Pro)}  \\
\hline

\rule{0pt}{45pt}\makecell[c]{\textbf{Second revolution:}\\{capture, transmit and \texttt{share} images}}&\makecell[c]{\includegraphics[height=55pt]{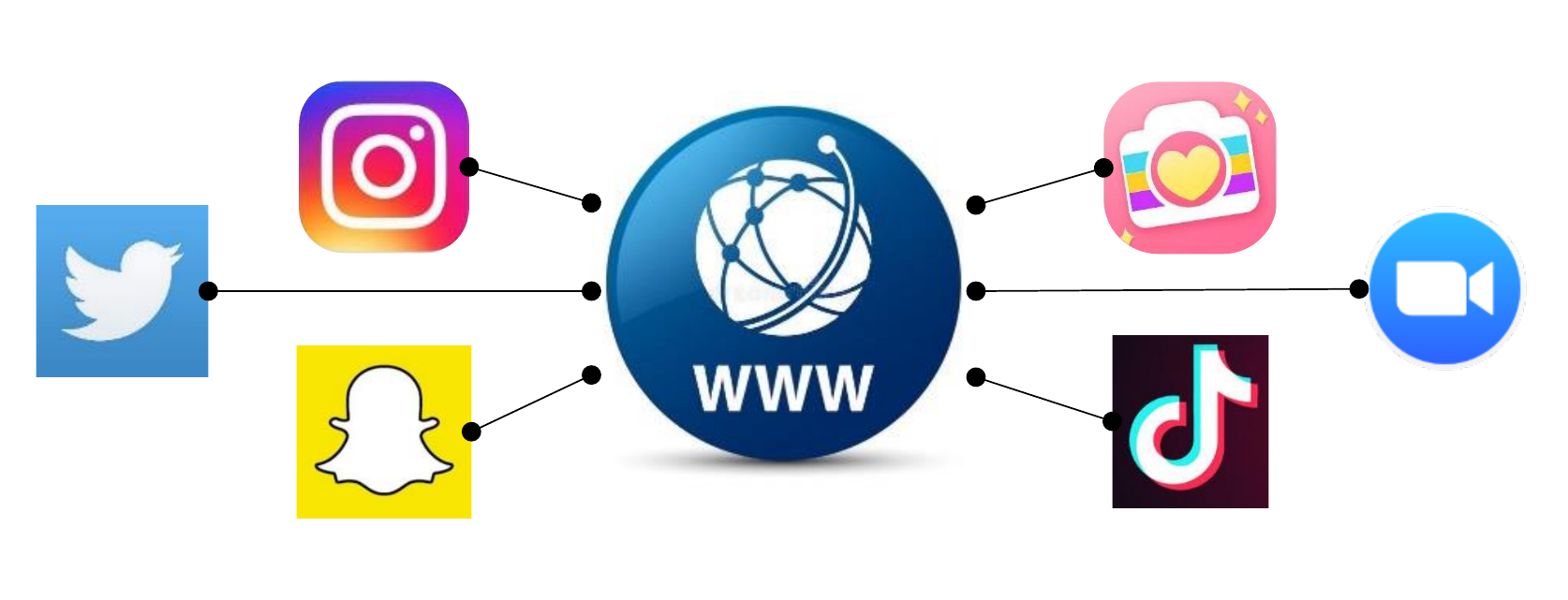}}& \makecell[l]{2000s\\$\bullet$ Booming of video conferences and instant video \\ \quad messaging (Skype/iChat)\\$\bullet$ Popularization of social networks sharing \\ \quad images and videos \\ \quad (Friendster/MySpace/Twitter/Instagram, \etc.)\\ \quad} \\
\hline

\rule{0pt}{45pt}\makecell[c]{\textbf{Third revolution:}\\{\texttt{intelligent} vision}}&\makecell[c]{\includegraphics[height=55pt]{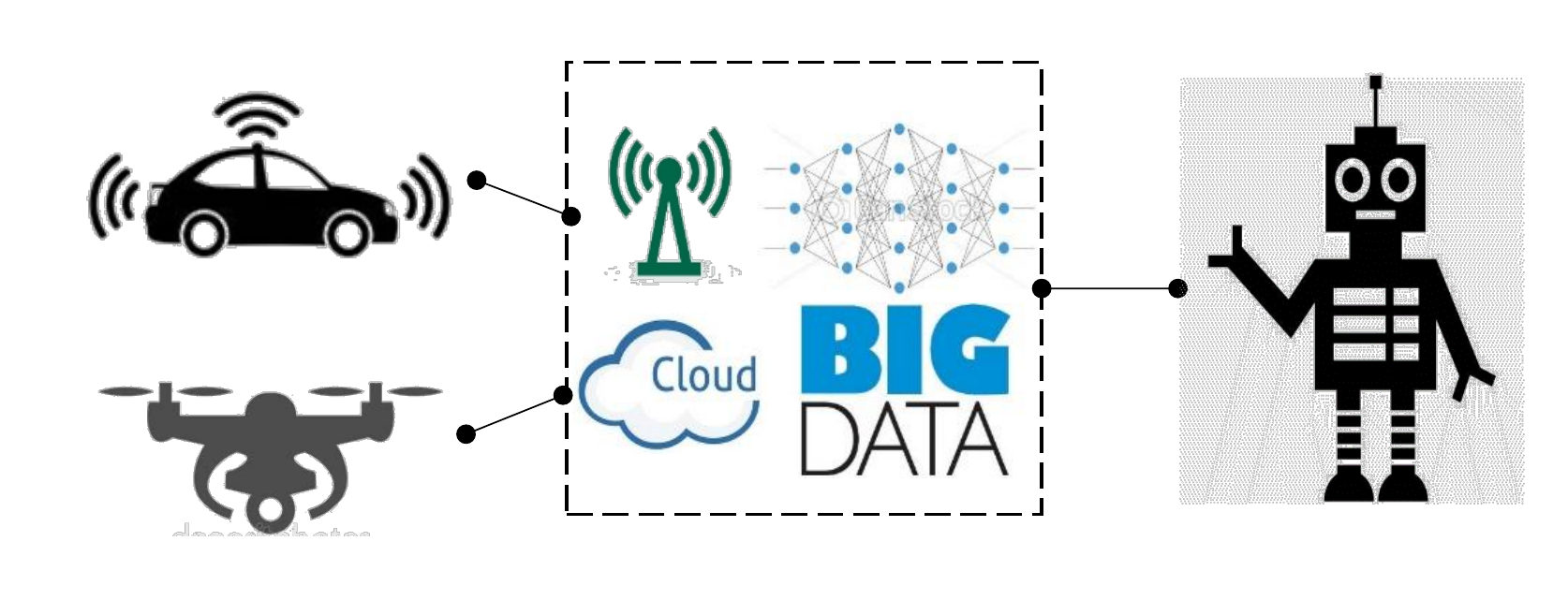}}& \makecell[l]{Ongoing process, with preliminary research studies:\\
$\bullet$ Connected and autonomous vehicles (CAVs) \\ \quad  using computational imaging systems  \\ 
$\bullet$ Robotic vision using computational cameras\\
\quad}\\

\hline

\end{tabular}
}
\label{tab1}
\end{table*}

While the second revolution of mobile vision has changed our life style significantly, grand challenges have also been raised in an unexpected way.
For instance, the bandwidth is usually limited while the end users need to capture more data.
On the other hand, though a significant amount of data are captured, the information increase is limited. For instance, most of the videos and images shared on the social media are barely watched in detail, especially in the pixel level. For another example, although the resolution of the mobile phone camera is already very high, in the same photo with the size possibly about several Mbs or even more than 10Mbs, the area we are interested in will not be clearer or contain more details than the broad background where we pay no attention, which, on the other hand, is often occupying the storage space of the mobile phone.
This poses two questions:

\begin{itemize}
    \item How to capture more information?
    \item How to extract useful information from the data?
\end{itemize}
These two questions can be addressed by CI and AI, respectively, which are the two main topics in this study. 
Moreover, we take one step further to answer both questions in one shot:
\begin{itemize}
    \item Can we build new devices to capture more useful information and directly use the information to perform the desired task and to maximize the performance?
\end{itemize}
The answer of this question leads to the {\bf next (third) revolution} of mobile vision, which will integrate CI and AI deeply and make the imaging systems be a combination of intelligent acquisition, processing and decision-making in high-speed modes. Mobile vision will now be in the mode of {\em intelligent vision}. 

{In this third revolution}, using AI we can build imaging systems that are optimized to functional metrics such as detection accuracy and recognition rate, {not just the amount of information they capture.}
One function of the third revolution will still be to communicate and share images, but {rather than sharing the actual
physical image, the data is mined and transformed by AI to {be optimized} for communication}. Similarly, {the imaging system
can understand the intent} of its use in different situations, for example to measure objects in a scene, to help quantify
objects, to identify an object, to help with safety, \etc. The system tries to intelligently use resources for task-specific
rewards. Rather than maximizing data or information, it maximizes performance.
These new imaging devices are dubbed {\em computational imaging systems} in this study.  

Equipped with this new revolution, and with the rapid development of machine vision and artificial intelligence, we anticipate that intelligent mobile vision platforms, such as self-driving vehicles, drones, and autonomous robots, are gradually becoming reality. 
With the advances of AI, the boundary between mobile vision and intelligent imaging is getting blurry. When applied to mobile platforms, the intelligent imaging systems need to take load, available resources and granted response time into consideration.
These mobile vision platforms aim to replace or even improve humans to perform specific tasks such as navigation, transportation, and tracking. In a nutshell, the operations of mobile intelligent platforms mainly  include five stages: {\em acquisition, communication, storage, processing and decision-making}. In other words, these systems capture real-time visual information from external environment, fuse more information by communication and sharing, digest and process  the information, make timely decisions and take corresponding actions successively.
In the meantime, emerging topics and techniques would support and promote the development of new mobile vision platforms. For instance, big data technology can help process and analyze large amount of collected data within allocated timing period. Edge (cloud) computing technology can be utilized to access powerful and reliable distributed computing resources for data storage and processing. Besides, the 5G communication technology being applied to smartphones would support high-speed and high-bandwidth data transmission. 

Table~\ref{Tab:3revolution} summarizes the representative events of the three revolutions in mobile vision described above.
Notably, though the intelligent vision has recently been advanced significantly by AI in the third revolution, automation and intelligent systems have been long time desires of human beings.
For example, the idea of using AI and cameras to control mobile robots can be dated back to 1970s~\cite{nilsson1984shakey}, and using video cameras and computer software to develop self-driving cars has been initiated in 1986~\cite{Jochem1995PANS}.

\subsection{Current Status}

Most existing mobile vision systems are basically a combination of camera and digital image processor. The former maps the high-dimensional continuous visual information into a 2D digital image, during which only a small portion of information is recorded such that the image systems are of low throughput. The latter extracts semantic information from the recorded measurements, which are intrinsically redundant and limited for decision making and following actions. 
Both image acquisition (cameras) and the subsequent processing algorithms face big challenges in real applications, including insufficient image quality, {unstable algorithms}, long running time and limited resources such as memory, processing speed and transmission bandwidth. Fortunately, we have witnessed the rapid development in camera industry and artificial intelligence. The progresses in other related fields such as cloud computing and 5G are potentially providing solutions {for} these resource limitations.

\subsection{Computational Imaging (CI)}
Different from the above ``capturing images first and processing afterwards" regime, the emerging computation imaging technique~\cite{Altmann18Science,Mait18_AOP_CI} provides a new architecture integrating these two steps (capturing and processing) to improve the perception capability of imaging systems (Fig.~\ref{fig:CI-comparision}).
Specifically, CI combines optoelectronic imaging with algorithmic processing, and {\em designs new imaging mechanisms} for optimizing the performance of computational analysis. Incorporating computing into the imaging setup requires the parameters settings such as focus, exposure, and illumination, being controlled by complex algorithms developed for image estimation \cite{brady2020smart}.
Benefited from the joint design, CI can better extract the scene information by optimized encoding and corresponding decoding. Such architecture might help mobile vision in the scenarios challenging conventional cameras, such as low light imaging, non-line-of-sight object detection, high frame rate acquisition, \etc.

\subsection{Incubation of Joint CI and AI}
{In CI systems, the sensor measurements are usually} recorded in an {\em encoded} manner, and so some complex decoding algorithms are required to retrieve the desired signal.
AI is widely used in the past decade due to its high performance and efficiency, which can be a good option for fast high-quality decoding. Besides, AI has achieved big successes in various computer vision tasks, and might even inspire researchers to design new CI systems for reliable perception with compact structures.
Therefore, the convergence of CI and AI is expected to play significant roles in each stage (such as acquisition,  communication,  storage,  processing  and decision-making) of the next revolution of mobile vision platforms.

Recently, various AI methods represented by deep neural networks that use large training datasets to solve the inverse problems in imaging \cite{lucas2018using}\cite{rick2017one}\cite{Ongie2020_JSAIT} have been proposed to conduct high-level vision tasks, such as detection, recognition, tracking, \etc. 
On one hand, this has advanced the field of CI to maximize the performance for specific tasks, rather than just capture more information as mentioned earlier.
On the other hand, this scheme is different from human vision system. As the fundamental objective of human vision system is to perceive and interact with the environment, the vision cortex extracts semantic information from encoded visual information and the intermediate image might be skipped/unnecessary, leading to a higher processing efficiency. Hence, it is of great significance to develop new visual information encoding, and decoding informative semantics from the raw data, which is also the main goal of CI. This also shares the same spirit with smart cameras \cite{brady2020smart}, \ie, to intelligently use resources of task-specific rewards to optimize the desired performance. 
As an additional point, the CI systems reviewed in this study is more general than conventional cameras used for photography in the visible wavelength.


In this comprehensive survey, we will review the early advances and ongoing  topics integrating CI and AI, and outlook the trends and future work in mobile vision, hoping to provide highlights for the research of the next generation mobile vision systems.
Specifically, after reviewing existing representative works, we put forward the prospect of next generation mobile vision. This prospect is based on the fact that the combination of CI and AI has made great progress, but meanwhile it is not close to the mobile vision, and thus cannot be widely used on mobile platforms currently.  CI systems start being used in many fields of our daily life (Section~\ref{Sec:CI}), and AI algorithms also play key roles in each stage of CI process (Section~\ref{Sec:AIforCI}). 
It is promising to leverage the developments of these two fields to advance the next generation mobile vision systems.
Based on this, in Section~\ref{Sec:Outlook}, we propose a new mobile vision framework to deeply connect CI and AI using autonomous driving as an example. The premise of this framework is that each mobile platform is equipped with CI systems and adopts efficient AI algorithms for image perception and decision-making. Within this framework, we present some AI methods that may be used, such as big data analysis and reinforcement learning to assist decision-making, which are mostly based on the input from CI systems. Finally, in Section~\ref{Sec:X}, we introduce related emerging research disciplines that can be used in CI and AI to revolutionize the mobile vision systems in the future.

\section{Emerging and Fast Developing CI \label{Sec:CI}}

\begin{figure}[ht]
\centering
\includegraphics[width=1.0\linewidth]{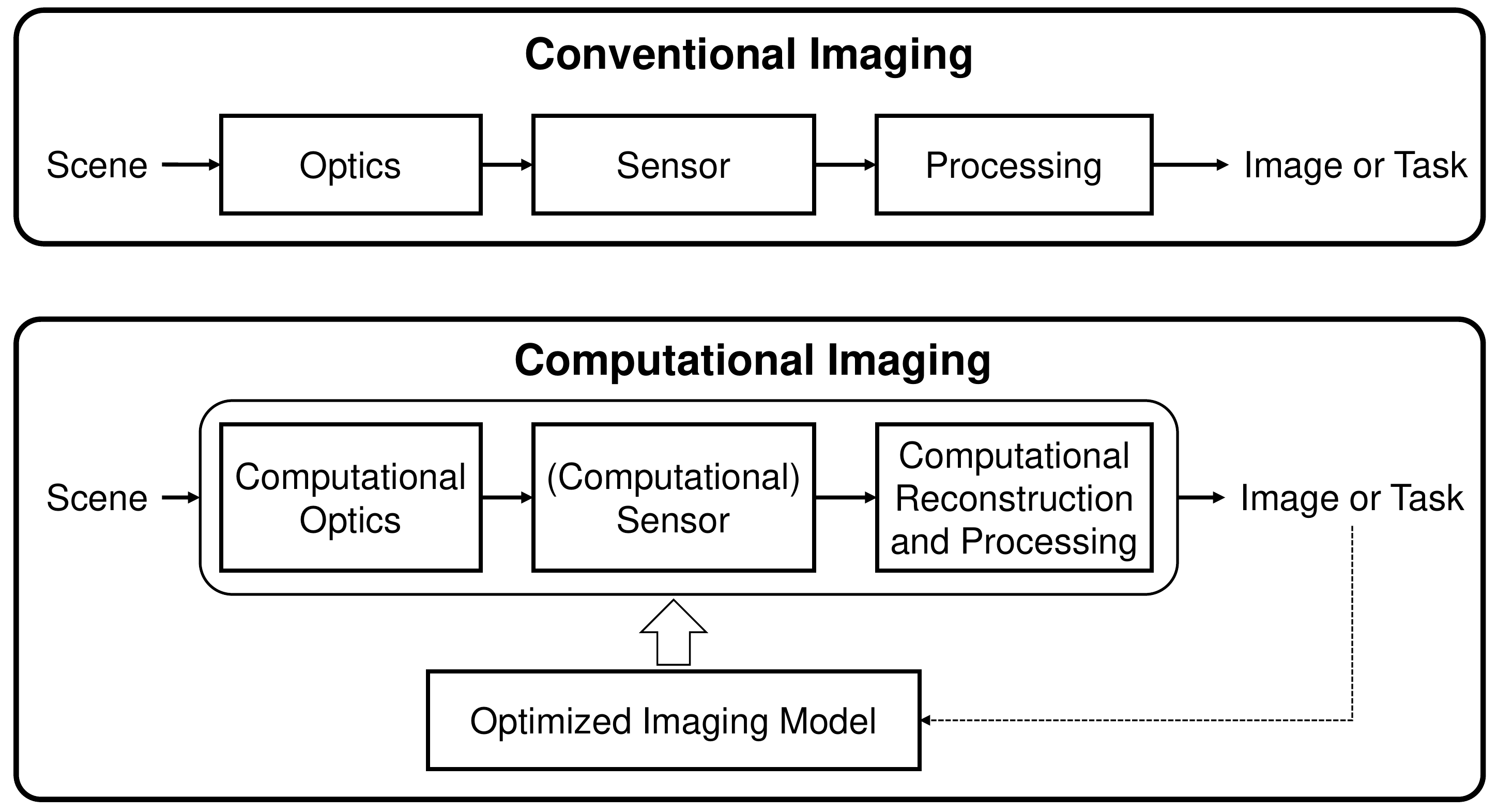}
\caption{Comparison of conventional imaging (top) and computational imaging (bottom). }
\label{fig:CI-comparision}
\end{figure}

Conventional digital imaging maps high-dimensional continuous information to two-dimensional discrete measurements, via projection, analog-to-digital conversion and quantization.
After decades of development, digital imaging has achieved a great success and promoted a series of related research fields, such as computer vision, digital image processing, \etc~\cite{agrawal2016signal}.
This processing line is shown in the top part of Fig.~\ref{fig:CI-comparision} and the related image processing technologies, such as pattern recognition have significantly advanced the development of machine vision. However, due to the inherent limitation of this ``capturing images first and processing afterwards" scheme, machine vision has encountered bottlenecks in both imaging and successive processing, especially in real applications.
The dilemma is mainly due to the limited information capture capability, since during the (image) capture a large amount of information is filtered out (\eg, spectrum, depth, dynamic range, \etc.) and the subsequent processing is thus highly ill-posed.

Computational imaging, by contrast, considers the imaging system and the consequent processing in an integrated manner. This new scheme blurs the boundary between optical acquisition and computational processing, and moves the calculation forward to the imaging procedure, \ie, introduces computing to design new task-specific systems. Specifically, as shown in the bottom part of Fig.~\ref{fig:CI-comparision}, in some cases, CI changes the illumination and optical elements (including apertures, light paths, sensors, \etc.) to encode the visual information and then decodes the signal from the coded measurements computationally. {The encoding process can encompass more visual information than conventional imaging systems or task-oriented information for better post-analysis.}
In some cases, CI generalizes multiplexed imaging to high dimensions of the plenoptic function~\cite{wetzstein2013plenoptic}.

\begin{table*}[t]\centering
\footnotesize
\caption{Representative works of computational imaging research. We emphasize that this list is by no means exhaustive.}
\resizebox{1\textwidth}{!}{
\begin{tabular}{c|l|l|c}
\hline
\rule{0pt}{10pt}{ }& \makecell[c]{\textbf{Methods}} & \makecell[c]{\textbf{Applications}}&\textbf{References}\\
\hline
\rule{0pt}{40pt}\makecell[c]{\includegraphics[width=45pt,height=40pt]{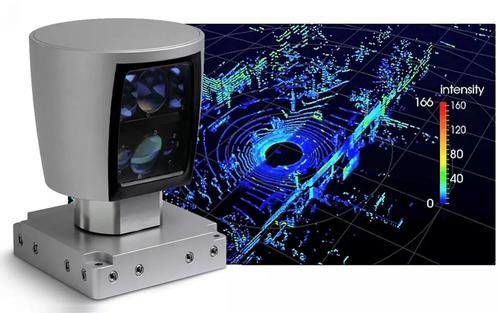}\\ \textbf{\scriptsize{Depth}}} & \begin{minipage} [t] {0.65\textwidth} 
\vspace{-12mm}
\begin{itemize}
    \item {\bf Binocular system} obtains depth information through the parallax of dual cameras.
    \item {\bf Structured light} projects given coding patterns to improve feature matching effect.
    \item {\bf Light-field imaging} calculates the depth from the four dimensional light field.
    \item {\bf Depth from focus/defocus} estimates the depth from relative blurring.
    \item {\bf Time of flight and LiDAR} continuously transmit optical pulses and detect the flight time.
\end{itemize} 
\end{minipage}
& \makecell[l]{1. Self-driving\\2. Entertainment \\3. Security system\\4. Remote sensing}&{\cite{sciammarella1982moire,grinberg1994geometry,Sun17OE,chen2012depth,Sun16OE,Willomitzer_OE17,wu2020freecam3d,ng2005light,chang2016variable,tambe2013towards,lehtinen2011temporal,jayasuriya2015depth,nayar1994shape,el2019solving,el2019closed,xiong1993depth,subbarao1994depth,chaudhuri2012depth,Zhou2009_ICCV,levin2007image,kazmi2014indoor,heide2015doppler,shrestha2016computational,marco2017deeptof,su2018deep,gong2016three,kirmani2014first,o2017reconstructing,lindell2018towards,heide2018sub,wu2018learning,sun2020spadnet,chang2019deep,wu2019phasecam3d,ba2019deep,zhang2020multiscale,zhao2012ghost,levoy1996light,Llull15Optica,kadambi2017rethinking}} \\
\hline

\rule{0pt}{30pt}\makecell[c]{\includegraphics[width=45pt,height=40pt]{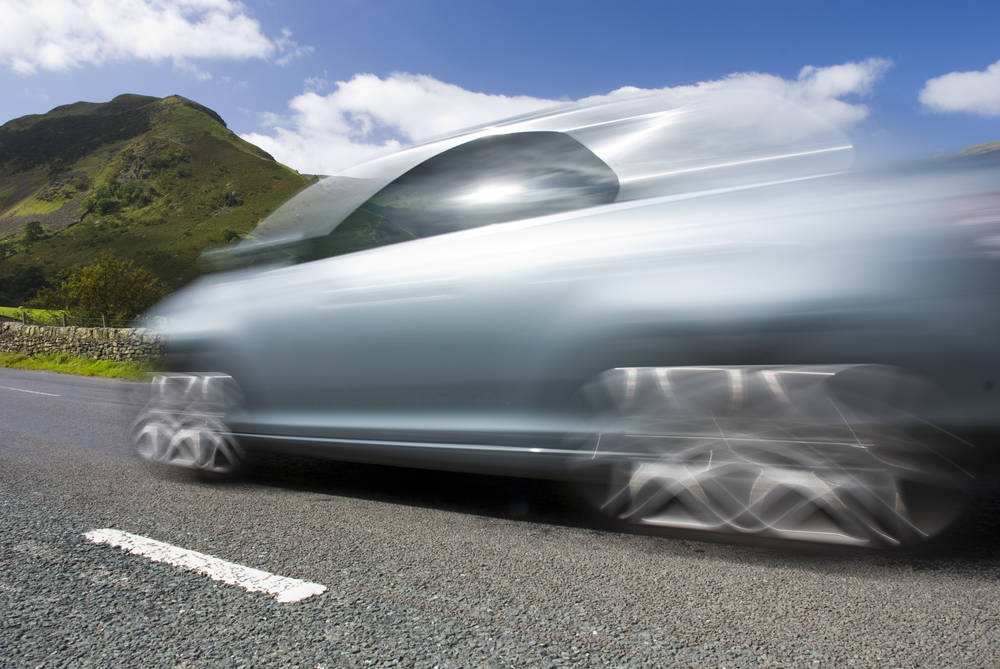}\\ \textbf{\scriptsize{Motion}}}& 
\begin{minipage} [t] {0.65\textwidth} 
\vspace{-7mm}
\begin{itemize}
\item {\bf Compressive sensing} is widely applied to capture fast motions by recovering a sequence of \\frames from their encoded combination.
\item {\bf Snapshot imaging} uses a spatial light modulator to modulate the high-speed scenes and develops\\ end-to-end neural networks or Plug-and-Play frameworks.
\end{itemize} 
\end{minipage}
& \makecell[l]{1. Surveillance \\2. Sports \\3. Industrial inspection \\4. Visual navigation}&\makecell[c]{{\cite{hitomi2011video,reddy2011p2c2,llull2013coded,deng2019sinusoidal,liu2013efficient,gao2014single,qi2020single,liang2020single,velten2013femto,antipa2019video,agrawal2010optimal,Qiao2020_CACTI,Yuan14CVPR,Yuan16BOE,Sun17OE}}}\\
\hline

\rule{0pt}{30pt}\makecell[c]{\includegraphics[width=45pt,height=40pt]{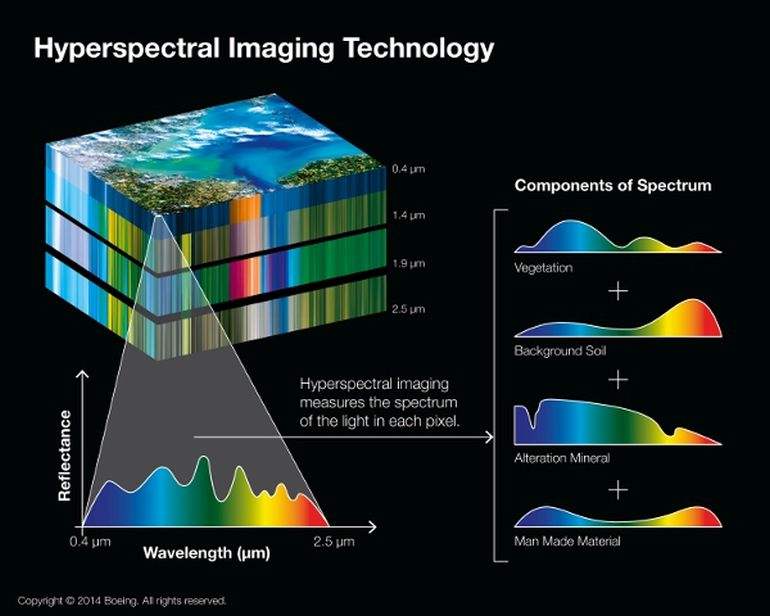}\\ \textbf{\scriptsize{Spectrum}}}& 
\begin{minipage} [t] {0.65\textwidth} 
\vspace{-8mm}
\begin{itemize}
\item {\bf Hyperspectral imaging} combines imaging technology and spectral technology to detect the two-dimensional geometric space and one-dimensional spectral information of the target and obtain the continuous and narrow band image data of hyperspectral resolution. In the spectral dimension, the image is segmented among not only R, G, B, but also many channels in the spectral dimension, which can feed back continuous spectral information.
\end{itemize}
\end{minipage}
& \makecell[l]{1. Agriculture\\2. Food detection\\3. Mineral testing\\4. Medical imaging}&\makecell[c]{{\cite{donoho2006compressed,Candes06TIT,zhao2019hyperspectral,cao2011prism,ma2014acquisition,cao2016computational,arad2018ntire,goodfellow2020generative,wagadarikar2008single,mohan2008agile,lin2014dual,Yuan15JSTSP,Meng2020_OL_SHEM}}}\\
\hline

\rule{0pt}{30pt}\makecell[c]{\includegraphics[width=45pt,height=40pt]{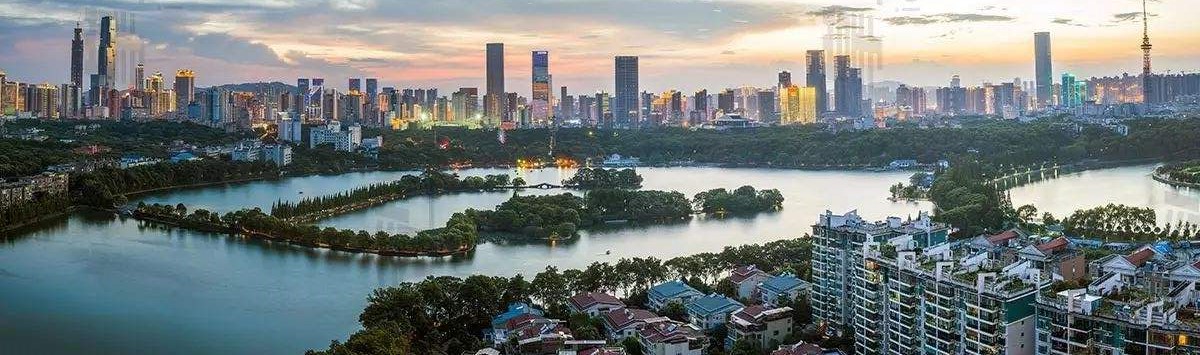}\\ \textbf{\scriptsize{Field-of-view}}}&
\begin{minipage} [t] {0.65\textwidth} 
\vspace{-5mm}
\begin{itemize}
\item {\bf Large field-of-view imaging} methods design a multi-scale optical imaging system, in which multiple small cameras are placed in different fields of view to segment the image plane of the whole field and the large field-of-view image is stitched through subsequent processing.
\end{itemize}
\end{minipage}
& \makecell[l]{1. Security\\2. Agriculture\\3. Navigation \\4. Environment monitoring}&\makecell[c]{{\cite{lohmann1996space,cossairt2011gigapixel,brady2012multiscale,yuan2017multiscale,Qiao2020_CACTI}}}\\
\hline

\rule{0pt}{30pt}\makecell[c]{\includegraphics[width=45pt,height=40pt]{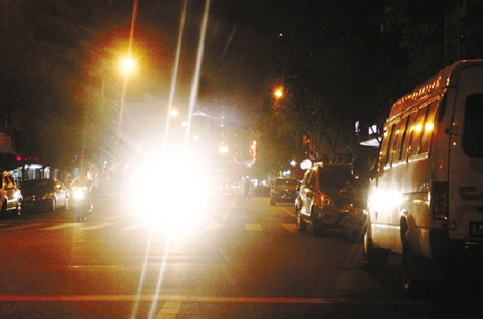}\\ \textbf{\scriptsize{Dynamic range}}}& \begin{minipage} [t] {0.65\textwidth} 
\vspace{-8mm}
\begin{itemize}
\item The {\bf spatial light modulator based method} performs multiple-length exposures of images in the time domain.
\item {\bf Encoding-based methods} encode and modulate the exposure intensity of different pixels in the spatial domain. Different control methods have been used to improve the dynamic range through feedback control.
\end{itemize}
\end{minipage}
& \makecell[l]{1. Digital photography\\2. Medical imaging\\3. Remote sensing\\ 4. Exposure control}&\makecell[c]{{\cite{nayar2000high,Metzler_2020_CVPR,sun2020learning,reinhard2010high,cvetkovic2010automatic,Nayar03_HDR,banterle2017advanced,rahman2011using,barakat2008minimal,martel2020neural,zhao2015unbounded} }}\\
\hline

\rule{0pt}{30pt}\makecell[c]{\includegraphics[width=45pt,height=40pt]{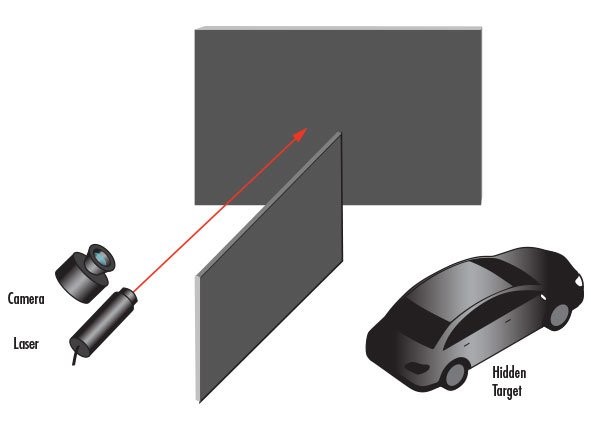}\\ \textbf{\scriptsize{NLOS}}}& \begin{minipage} [t] {0.65\textwidth} 
\vspace{-8mm}
\begin{itemize}
\item {\bf Non-Line-Of-Sight} (NLOS) imaging  can recover images of the scene from the indirect light. 
\item {\bf Active NLOS} uses laser as source and single photon avalanche diode (SPAD) or streak camera as the receiver.
\item {\bf Passive NLOS} uses a normal camera to capture the scattered  light and uses algorithms to reconstruct the scene.
\end{itemize}
\end{minipage}
& \makecell[l]{1. Self-deriving\\2. Military reconnaissance}&\makecell[c]{{\cite{faccio2020non,velten2012recovering,o2018confocal,lindell2019wave,ahn2019convolutional,musarra2019non,musarra20193d,la2018error,heide2019non,tsai2019beyond,thrampoulidis2018exploiting,rapp2020seeing,tanaka2020polarized,ye2021compressed,wu2021non,maeda2019thermal,lindell2019acoustic,satat2017object,isogawa2020optical,metzler2019keyhole,tancik2018nlos,pandharkar2011estimating,katz2012looking,liu2019non,raskar2020seeing,bedri2014seeing,mao2018aim,su2015acoustic,raskar2014looking,katz2014non,kirmani2011looking,buttafava2015non,xin2019theory,reza2019phasor,reza2018physical,lei2019direct,reza2018imaging,kiran2018non,la2020non,liu2020role,guillen2020effect,kirmani2009looking,lindell2020computational,liu2020phasor,tancik2018data,pandharkar2011hidden,seidel2020two,geng2021recent}}}\\
\hline

\rule{0pt}{30pt}\makecell[c]{\includegraphics[width=45pt,height=40pt]{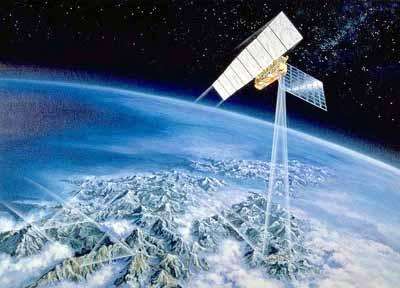}\\ \textbf{\scriptsize{Radar}}}& \begin{minipage} [t] {0.65\textwidth} 
\vspace{-8mm}
\begin{itemize} 
\item {\bf Synthetic Aperture Radar} (SAR) uses the reflection of the ground target on the radar beam as the image information formed by the backscattering of the ground target. 
SAR is an active side-view radar system and its imaging geometry is based on oblique projection.
\item {\bf Through-the-wall radar} can realize motion detection without cameras.
\item {\bf WiFi} has been used for indoor localization.
\end{itemize}
\end{minipage}
& \makecell[l]{1. Drones\\2. Military inspection \\ 3. Elder care}&\makecell[c]{{\cite{moses2004wide,sun2020mimo,mansour2018sparse,sun20214d,amin2014through,amin2016radar,wu2014compressive,wu2014through,wang2019human,zhang2018latern,zhang2019udeep,hazra2018robust,kotaru2015spot,wang2017biloc,wang2017csi,tomiyasu1978tutorial,willett2013sparsity,holloway2017savi}}}\\
\hline

\end{tabular}
}
\label{tab2}
\end{table*}


The innovation of CI (by integrating sensing and processing) has strongly promoted the development of imaging researches and produced a number of representative research areas, which have significantly improved the imaging capabilities of the vision systems, especially mobile vision.
In the following, we show some representative examples: depth, high speed, hyperspectral, wide field-of-view, high dynamic range, non-line-of-sight and radar imaging.
All these progresses demonstrate the advantageous of CI and the large potential of using CI in mobile vision systems.
As mentioned before, CI systems are not limited to visible wavelength for the photography purpose. In this study, we consider diverse systems aiming to capture {\em images} for mobile vision for can be used in mobile vision in the future.
We summarize the recent advances of representative CI in Table~\ref{tab2} and describe some research topics below, however, this list is by no means exhaustive.

\subsection{Depth Imaging}
Depth image refers to an image with intensity describing the distance from the target scene to the camera, which directly reflects the geometry of the (usually visible) surface of the scene. As an extensively studied topic, there exist various depth imaging methods, {including LiDAR ranging, stereo vision imaging, structured illumination method{\cite{sciammarella1982moire}}, \etc. }
In the following, we list some examples.
\begin{itemize}
    \item Stereo vision or binocular systems might be the first passive and well studied depth imaging technique. It has a solid  theory based on geometry~\cite{grinberg1994geometry} and is also widely used in current CI {systems}~\cite{Sun17OE}. 
    \item {Structured illumination, as an active method,} has been widely used in our daily life for high-precision and high-speed  depth imaging~\cite{chen2012depth}. Different structured illumination methods are being actively developed in CI systems~\cite{Sun16OE,Willomitzer_OE17,wu2020freecam3d}.
    \item Light-field imaging {is} another way to perform depth imaging. In this case, depth is calculated from the four-dimensional light field, in addition to confining the scene information to three-dimensional spatial coordinates~\cite{levoy1996light}. Ng et al. {\cite{ng2005light}} proposed a plenoptic camera that captures 4D light field with a single exposure, which is of compact structure, portable, and can also help refocusing. With CI, light field camera has also been employed to overcome the spatio-angular resolution  trade-off \cite{chang2016variable}, and extended to capture dynamic and transient scenes~\cite{tambe2013towards,lehtinen2011temporal,jayasuriya2015depth}.
    \item Depth from focus~/~defocus estimates the 3D surface of a scene from a set of two or more images of that scene. It can be implemented by changing the camera parameters~\cite{nayar1994shape,el2019solving,el2019closed,xiong1993depth,subbarao1994depth} or using coded apertures~\cite{chaudhuri2012depth,Zhou2009_ICCV,levin2007image}.
    \item {Time-of-Flight (ToF) cameras obtain the distance of the target object by detecting the round-trip time of the light pulse. Recently, Kazmi et al.~\cite{kazmi2014indoor} have verified through experiments that ToF can provide accurate depth data with high frame rate under suitable conditions.}
    Doppler ToF imaging~\cite{heide2015doppler} and CI with multi-camera ToF systems~\cite{shrestha2016computational} have also been built recently. The end-to-end optimization of imaging pipelines~\cite{marco2017deeptof}\cite{su2018deep} for ToF further overcomes the interference using deep learning. Based on the ToF principle but with sensor arrays, LiDAR has been widely used in autonomous driving and remote sensing; it can be categorized into two classes according to the measurement modes: scanning based and non-scanning based. The former one obtains the real-space image of the target by point-by-point (or line-by-line) scanning with a pulsed laser, while the latter one covers the whole field of the target scene and obtains the 3D map in a single exposure using a pulsed flash laser imaging system~\cite{gong2016three}. Kirmani et al. \cite{kirmani2014first} propose to use the first photon received by the detector for high-quality three-dimensional image reconstruction, applying computational algorithms to LiDAR imaging in the low-flux environment. 
\end{itemize}

In addition to the above categorized depth imaging approaches, recently, the single-photon senors have also been used in CI for depth estimation~\cite{o2017reconstructing,lindell2018towards,heide2018sub}.
In computer vision, single image 3D is another research topic~\cite{wu2018learning,sun2020spadnet,chang2019deep,wu2019phasecam3d}. By adapting local monocular cameras to the global camera, deep learning helps solve the joint depth estimation of multi-scale camera array and panoramic object rendering, in order to provide higher 3D image resolution and better quality virtual reality (VR) experience \cite{zhang2020multiscale}. Another idea is depth estimation and 3D imaging based on polarization. As the light reflected off the object has the polarization state corresponding to the shape, Ba et al. \cite{ba2019deep} input the polarization image into neural network to estimate the surface shape of the object, and establish a polarization image dataset.
Regarding the applications, depth imaging has been widely used in our daily life such as self-driving vehicles and entertainments. Recently, face identify systems start using the depth information for the security consideration. With varying resolutions ranging from meters to micrometers, depth imaging systems have been widely used in remote sensing and industrial inspection. 

\subsection{High Speed Imaging}
For an imaging system, the frame rate is a trade-off between the illumination level and sensor sensitivity. An insufficient frame rate would result in missing events or motion blur. Therefore, high-speed imaging is crucial for recording of highly dynamic scenes. In CI, the temporal resolution can be improved by temporal {compressive} coding.

Hitomi et al.~\cite{hitomi2011video} employed a digital micromirror device to modulate the high-speed scene and used dictionary learning method to reconstruct a video from a single compressed image with high spatial resolution.
A liquid crystal on silicon (LCoS) device was used for modulation in~\cite{reddy2011p2c2} to achieve temporal compressive imaging.
Llull et al.~\cite{llull2013coded} used mechanical translation of a coded aperture for encoded acquisition of a sequence of time stamps, dubbed Coded Aperture Compressive Temporal Imaging (CACTI).
Deng et al.~\cite{deng2019sinusoidal} used the limited {spatio-spectral} distribution of the above coded measurements, and further increased the frame rate by a novel compressive video acquisition technique, termed Sinusoidal Sampling Enhanced Compression Camera (S2EC2). This method enhanced widely-used random coding pattern with a sinusoidal one and successfully multiplexed a groups of coded measurements within a snapshot.
Another representative work is by Liu et al. {\cite{liu2013efficient}}, who implemented coded exposure on the complementary metal–oxide–semiconductor (CMOS) image sensor with an improved control unit instead of a spatial light modulator (SLM), and conducted reconstruction by utilizing the sparse representation of the video patches using an over-complete dictionary.
All these methods increased the imaging speed while maintaining a high spatial resolution. Solely aiming to capture high-speed dynamics,
streak camera is a key technology to achieve femto-second photograph~\cite{velten2013femto}. Combining streak camera with compressive sensing, compressed ultrafast photography (CUP)~\cite{gao2014single} was developed to achieve billions frames per second~\cite{qi2020single} and has been extended to other dimensions~\cite{liang2020single}. However, there might be a long way to implement CUP into mobile platforms.

High-speed imaging has wide applications in sports, surveillance, industrial inspections and visual navigation.

\subsection{Hyperspectral Imaging}
Hyperspectral imaging technology aims to capture the spatio-spectral data cube of the target scene, which combines imaging with spectroscopy. 
Conventional hyperspectral imaging methods include spectrum splitting by grating, {acousto-optic tunable filter} or prism, and then conduct separate recording via scanning, which limits the imaging speed so they are inapplicable to capture dynamic scenes.
Since the hyperspectral data is intrinsically redundant,  {compressive} sensing theory~\cite{donoho2006compressed}\cite{Candes06TIT} provides a solution for {compressive} spectral sampling. The Coded Aperture Snapshot Spectral Imaging (CASSI) system proposed by Wagadarikar et al.~{\cite{wagadarikar2008single}} realized using a two-dimensional detector to compressively sample the three-dimensional ({spatio-spectral}) information by encoding, dispersing and integrating the spectral information within a single exposure. 
Following this, other type of methods achieves hyperspectral imaging by designing and improving masks. For a compact design, Zhao et al. {\cite{zhao2019hyperspectral}} proposed a method acquiring coded images using a random-colored mask by a consumer-level printer in front of the camera sensor, and computationally reconstructed the hyperspectral data cube using a {deep learning based} algorithm. {In addition, the hybrid camera systems~\cite{cao2011prism}\cite{ma2014acquisition} which integrated high-resolution RGB video and low-resolution multispectral video have also been developed. For a more comprehensive review of the hyperspectral CI systems, please refer to the survey by Cao et al.~\cite{cao2016computational}. }
Another related work is to recover the hyperspectral images from an RGB image by learning based methods~\cite{arad2018ntire}, which is inspired by the deep generative models~\cite{goodfellow2020generative}.

Equipped with rich spectral information, hyperspectral imaging has wide applications in agriculture, food detection, mineral detection and medical imaging~\cite{Meng2020_OL_SHEM}.

\subsection{Wide Field-of-View High Resolution Imaging}
Wide field-of-view (FOV) computational imaging with both high spatial and high temporal resolutions is an indispensable tool for vision systems working in a large scale scene.
On one hand, high resolution helps observe details, distant objects and small changes, while a wide FOV helps observe the global structure and connections in-between objects. On the other hand, conventional imaging equipment has long been constrained by the trade-off between FOV and resolution and the bottleneck of {\em spatial bandwidth product} \cite{lohmann1996space}. With the development of high-resolution camera sensors, geometric aberration fundamentally limits the resolution of the camera in a wide FOV. In order to solve this problem, Cossairt et al. \cite{cossairt2011gigapixel} proposed a method to correct the aberrations by calculations. Specifically, they used a simple optical system containing a ball lens cascaded by hundreds of cameras to achieve a compact imaging architecture, and realized gigapixel imaging. 
Brady et al. \cite{brady2012multiscale} proposed a multi-scale gigapixel imaging system by using a shared objective lens.
Instead of using a specific system, Yuan et al. \cite{yuan2017multiscale} {integrated} multi-scale scene information to generate gigapixel wide-field video. 
A joint FOV and temporal compressive imaging system was built in~\cite{Qiao2020_CACTI} to capture high-speed multiple field of view scenes.
%
Large FOV imaging systems have wide applications in environmental pollution monitoring, security cameras and navigation, especially for multiple object recognition and tracking.

\subsection{High Dynamic Range Imaging}
There often exists a drastic range of contrast in a scene that goes beyond a conventional camera sensor.  One widely used method is to synthesize the final HDR image according to the best details from a series of {low-dynamic range (LDR) images} with different exposure elapses, which improves the imaging quality. 
Nayar et al. {\cite{nayar2000high}} proposed to use an optical mask with spatially varying transmittance to control the exposure time of adjacent pixels before the sensor, and to reconstruct the HDR image by an effective algorithm. With the development of SLM, the authors further proposed an adaptive method to enhance the dynamic range of the camera~\cite{Nayar03_HDR}. 
Most recently, deep learning has been used in the single-shot HDR system design~\cite{Metzler_2020_CVPR,sun2020learning}. With these approaches, CI techniques are able to achieve decent recording within both ends of a large radiance range \cite{reinhard2010high}.
Researchers have used the proportional-integral-derivative (PID) method to increase the dynamic range through feedback control~\cite{cvetkovic2010automatic}. 

HDR imaging aims to address the practical imaging challenges when very bright and dark objects coexist in the scene. Therefore, it is widely used in the digital photography by exposure control and also been used in medical imaging and remote sensing.

\subsection{Non-Line-of-Sight Imaging}

Non-line-of-sight (NLOS) imaging can recover details of a hidden scene from the indirect light that has scattered multiple times~\cite{faccio2020non}. 
During the very first research period, the streak camera is used~\cite{velten2012recovering} to capture the optical signal reflected by the underlying scene illuminated by the pulsed laser, but limited to one dimensional capture in one shot. Later, in order to improve the temporal and spatial resolution, single photon avalanche diode (SPAD) has become widely used in NLOS imaging~\cite{o2018confocal,lindell2019wave,ahn2019convolutional}, enjoying the advantage of no mechanical scanning, and can easily be combined with single-pixel imaging to achieve 3D capture~\cite {musarra2019non}~\cite {musarra20193d}. After capturing the transient data, the NLOS scene can be reconstructed in the form of volume~\cite{velten2012recovering,ahn2019convolutional,o2018confocal,lindell2019wave} or surface~\cite{tsai2019beyond}. In the branch of volume-based reconstruction, Velten et al.~\cite{velten2012recovering} use higher-order light transport to model the reconstruction of NLOS scene as a least square problem based on transient measurements, and solve it approximately by filtered back projection (FBP). This paves the way of solving this problem by deconvolution. For example, O'Toole et al.~\cite {o2018confocal} represent the higher-order transport model as light cone transform (LCT) in the case of confocal, and reconstruct the scene by solving the three-dimensional signal deblurring problem. Based on the idea of back propagation, Manna et al.~\cite{la2018error} prove that the iterative algorithm can optimize the reconstruction results and is more sensitive to the optical transport model. In addition, Lindell et al.~\cite {lindell2019wave} use the 3D propagation of waves and recover the geometry of hidden objects by resampling along the time-frequency dimension in the Fourier domain. Heide et al.~\cite {heide2019non} add the estimation of surface normal vector to the reconstruction result by an optimization problem. In the branch of surface-based reconstruction, Tsai et al.~\cite{tsai2019beyond} obtain the surface parameters of scene by calculating the derivative relative to NLOS geometry and measuring the reflectivity. Using the occlusions~\cite{thrampoulidis2018exploiting} or edges~\cite{rapp2020seeing} in the scene also helps to retrive the geometric structure. Recently, the polar signal ~\cite{tanaka2020polarized}, compressive sensing~\cite {ye2021compressed} and deep learning~\cite{tancik2018nlos} have also been used in NLOS to improve the performance. Most recently, NLOS has achieved a range of kilometers~\cite{wu2021non}. 
Other than optical imaging, Maeda et al.~\cite{maeda2019thermal} conduct thermal NLOS imaging to estimate the position and pose of the occluded object, and Lindell et al.~\cite{lindell2019acoustic} implement acoustic NLOS imaging system to enlarge the distance and reduce the cost.

At present, NLOS has been used in human pose classification through scattering media~\cite{satat2017object}, three-dimensional multi-human pose estimation~\cite{isogawa2020optical} and movement-based object tracking~\cite{metzler2019keyhole}.

In the future, NLOS has potential applications in self-driving vehicles and military reconnaissance.

\subsection{Radar Imaging}


As a widely used imaging technique in remote sensing, SAR (Synthetic-Aperture Radar) is an indirect measurement method that uses a small-aperture antenna to produce a virtual large-aperture radar via motion and mathematical calculations. It requires joint numerical calculation of radar signals at multiple locations to obtain a large field of view and high resolution. 

Moses et al.\cite{moses2004wide}  propose a wide-angle SAR imaging method to improve the image quality by {combining GPS} and the {UAV} technology. On the basis of traditional SAR, InSAR (Interferometric Synthetic-Aperture Radar), PolSAR (Polarimetric Synthetic-Aperture Radar) and other technologies have further improved the performance of remote sensing, and help implement the functions of 3D positioning, classification and recognition.


%

Recently, other radar techniques have also been used for imaging. For example, multiple-input-multiple-output radar has been widely used in mobile intelligent platform due to its high resolution, low cost and small size, which synthesizes a large-aperture virtual array with only a small number of antennas to achieve a high angular resolution~\cite{sun2020mimo}. Many efforts have been made to improve the imaging accuracy of radar on mobile platform. Mansour et al.~\cite{mansour2018sparse} propose a method to map the position error of each antenna to the spatial shift operator in the image domain and performing the multi-channel deconvolution to achieve superior autofocus performance. Sun et al.~\cite{sun20214d} use random sparse step-frequency waveform to suppress high sidelobes in azimuth and elevation and to improve the capability of weak target detection. In recent years, through-the-wall radar imaging (TWRI) has attracted extensive attention because of its applications in urban monitoring and military rescue~\cite{amin2014through}. Amin et al.~\cite{amin2016radar} analyze radar images by using short-time Fourier transform and wavelet transform to perform anomaly detection (such as falling) in elder care. At present, machine learning algorithms and deep neural networks are also applied to object classification and detection in radar imaging. Wu et al. use the clustering properties of sparse scenes to reconstruct high-resolution radar images~\cite{wu2014compressive}, and further develop a practical subband scattering model to solve the multi-task sparse signal recovery problem with Bayesian compressive sensing~\cite{wu2014through}. Wang et al.~\cite{wang2019human} use the stacked recurrent neural network (RNN) with long-short-term-memory (LSTM) units with the spectrum of the original radar data to successfully classify human motions into six types.
In addition, radar has also been applied to hand gesture recognition~\cite{zhang2018latern,zhang2019udeep,hazra2018robust} and WiFi used for localization~\cite{kotaru2015spot,wang2017biloc,wang2017csi}\cite{abbas2019wideep}.

Based on its unique frequency, depending on platforms such as satellites, drones and mobile devices, radar has its advantageous applications in military, agriculture, remote sensing and elder care.

\subsection{Summary}
All these systems described above involve computation in the image capture process.
Towards this end, diverse signal processing algorithms have also been developed. 
During the past decade, the emerging of AI, especially the advances of deep neural networks has dramatically improved the efficacy of the algorithm design.
Now it is a general trend to apply AI into CI design and processing. 

Again, we want to mention that the above representative list is by no means exhaustive. Some other important CI research topics such as low light imaging, dehazing, raindrop removal, scattering and diffusion imaging, ghost imaging and phaseless imaging
potentially also have a high impact in mobile vision.


\section{AI for CI: Current Status \label{Sec:AIforCI}}
%
The CI systems introduce computing into the design of imaging setup. They are generally used in three cases: ($i$) it is difficult to directly record the required information, such as the depth and large field of view;
($ii$) there exists a dimensional mismatch between the sensor and the target visual information, such as light field, hyperspectral data  and tomography, and ($iii$) the imaging conditions are too harsh for conventional cameras, e.g., high speed imaging, high dynamic range imaging.

CI obtains the advantageous over conventional cameras in terms of the imaging quality or the amount of task-specific information.
These advantageous come at the expense of high computing cost or limited imaging quality, since computational reconstruction is required to decode the information from the raw measurements.
The introduction of artificial intelligence provided early CI with many revolutionary advantages, mainly improving the quality (higher signal-to-noise ratio, improved fidelity, less artifacts, \etc.), and efficiency of computational reconstruction.
In  some applications, with the help of AI, CI can break some performance barrier in photography~\cite{mitra2014performance}\cite{Yuan_2021_Innovation}. 
Most recently, AI has been employed into {\em the design of CI systems} to achieve adaptivity, flexibility as well as  to improve performance in high-level mobile vision tasks with a schematic shown in Fig.~\ref{fig:Fourtypes}.


\subsection{AI Improves Quality and Efficiency of CI {System}}
\label{sec:AIforCIfirst}

\begin{figure}[t]
\centering
\includegraphics[width=1.0\linewidth]{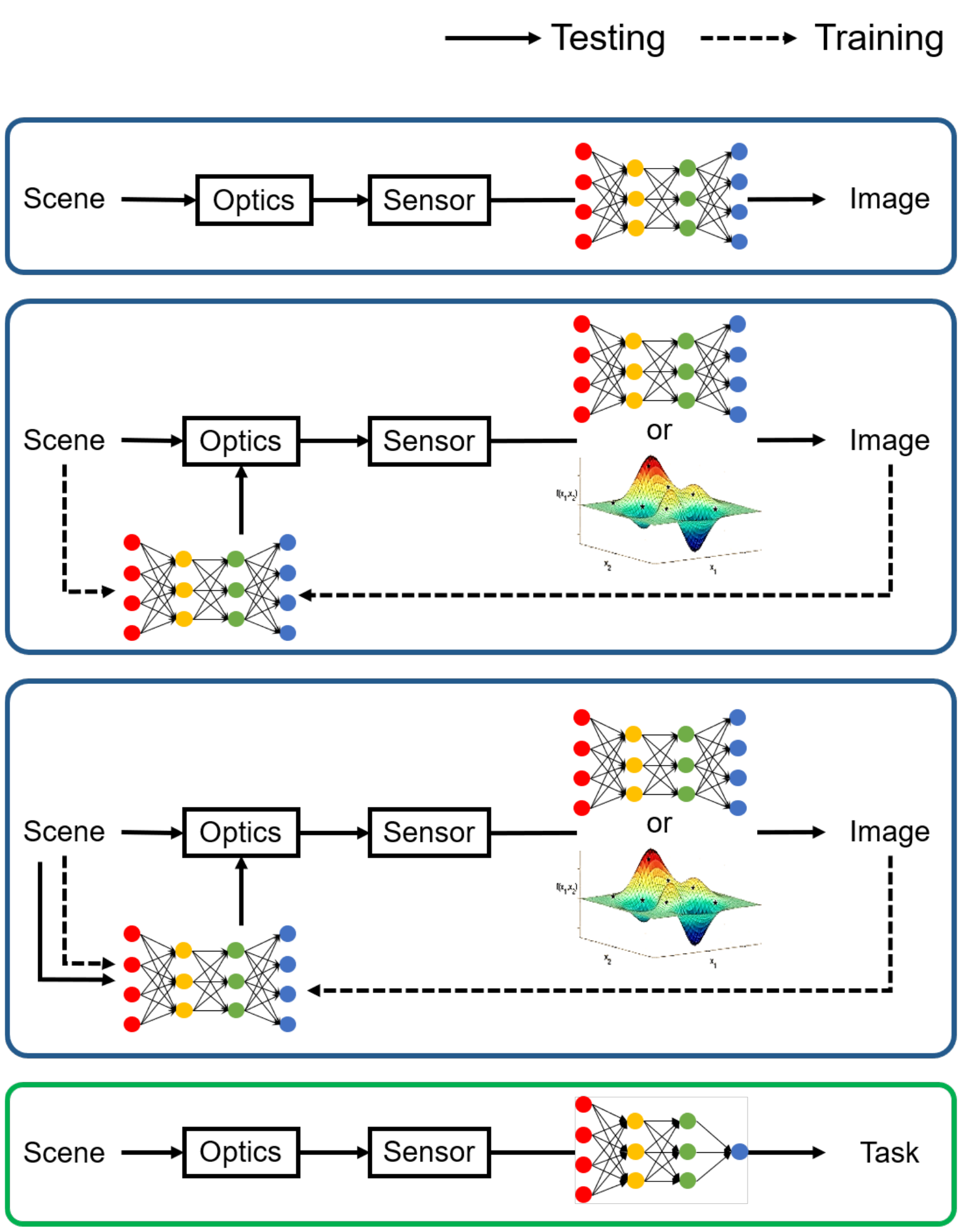}
\caption{Artificial intelligence for computational imaging: four categories described in sections \ref{sec:AIforCIfirst}, \ref{sec:AIforCIsecond}, \ref{sec:AIforCIthird} and \ref{sec:AIforCIlast}, respectively. From top to bottom, the first plot demonstrates that the AI algorithms only enhance the processing of the images captured by traditional imaging systems. The second plot shows that CI provides the quantitative modulation calculated by AI algorithm (to optimize the optics) in the imaging light path to obtain the encoded image, and then deep learning or optimization methods are used to obtain the required results from the measurements with rich information. The third 
plot shows that the AI algorithm provides the CI system with adaptive enhancement in the field of region of interest, parameters, and performance by analyzing the intrinsic connection between specific scenarios and tasks. The fourth plot shows that CI only captures information for specific tasks, which is subsequently applied to specific tasks by AI algorithms, reducing the required bandwidth.}
\label{fig:Fourtypes}
\end{figure}

As shown in the first plot of Fig.~\ref{fig:Fourtypes}, the most straightforward way and also the widely exploited  one is to use  AI to improve the quality and efficiency of CI systems. In this subsection, we use compressive imaging as an example to demonstrate the power of AI to improve the reconstruction quality of CI.
By incorporating spatial light modulation into the imaging system, compressive imaging can encompass high-dimensional visual information (video, {hyperspectral data cube}, light field, \etc.) into one or more snapshot(s). Such encoding schemes make extensive use of the intrinsic redundancy of visual information and can reconstruct the desired high-throughput data computationally. Thus, compressive imaging greatly reduces the recording and transmission bandwidth of high-dimensional imaging systems.
Since the decoding is an ill-posed problem, complex algorithms are required for high quality reconstruction. Researchers have spent great efforts on algorithm development (such as optimization methods using total variation (TV)~\cite{rudin1992nonlinear} and sparsity priors~\cite{hitomi2011video}\cite{Yuan15JSTSP} into generalized alternating projection (GAP)~\cite{Liao14GAP}\cite{Yuan16ICIP_GAP} and two-step iterative shrinkage/thresholding (TwIST)~\cite{Bioucas-Dias2007TwIST}, and new algorithms such as Gaussian mixture model (GMM)~\cite{Yang14GMM}\cite{Yang14GMMonline} and decompress-snapshot-compressive-imaging (DeSCI)~\cite{Liu18TPAMI}) and the performance is improved continuously during the past years. However, these algorithms are inferred under the optimization framework, and are usually either of limited accuracy or time consuming. 

Instead, the pre-trained deep networks can learn the image priors quite well and are of high computing efficiency. 
Taking the snapshot compressive imaging (SCI)~\cite{Yuan2021_SPM} as an example, SCI conducts specific encoding and acquisition from the scene, and takes the encoded single shot as the input of the decoding algorithm to reconstruct multiple interrelated images at the same time, such as continuous video frames or multi-spectral images from the same scene, with high compression ratio. At present, a variety of high-precision CNN structures have been proposed for SCI. Ma et al. \cite{ma2019deep} proposed a tensor-based  deep neural network, and achieved promising results. Similarly, Qiao et al. \cite{qiao2020deep} built a video SCI system using a digital micro-mirror device and developed both an end-to-end convolutional neural network (E2E-CNN) and a Plug-and-Play (PnP) algorithm for video reconstruction. The PnP framework~\cite{Yuan_2020_CVPR}\cite{Yuan2021_TPAMI_PnP} can serve as a baseline in video SCI reconstruction considering the trade-off among speed, accuracy and flexibility. PnP has also been used in other CI systems~\cite{kamilov2017plug}\cite{zheng2021deep}. With the development of deep learning, more optimization of the speed and logic of the SCI networks are proposed. By embedding Anderson acceleration into the network unit, a deep unrolling algorithm has been developed for SCI reconstruction~\cite{li2020end}. 
Considering the temporal correlation in video frames, the recurrent neural network has also been a tool~\cite{Cheng20ECCV_Birnat}\cite{Zheng2021_Patterns} for video SCI. For large-scale SCI reconstruction, the memory-efficient network and meta learning~\cite{Cheng2021_CVPR_ReverSCI}\cite{Wang2021_CVPR_MetaSCI} pave the way. To overcome the challenge of limited training data, the untrained neural networks is proposed most recently~\cite{Qiao2021_MicroCACTI}\cite{Meng2021_ICCV_self}, which also achieves high performance with the requirement of less data. Based on the optimized neural network, the application driven SCI meets the specific needs and achieves real-time and dynamic imaging. For example, neural network based reconstruction methods have been developed for spectral SCI systems~\cite{Miao19ICCV,Meng20ECCV_TSAnet,Huang2021_CVPR_GSMSCI}.
By introducing a diffractive optical element in front of the conventional image sensor, a compact hyperspectral imaging system \cite{jeon2019compact} has been built with CNN for high-quality  reconstruction, performing well in terms of spectral accuracy and spatial resolution.

In addition to reconstructing multiple images with a single measurement, 
deep learning has also been used to decode the high-dimensional data from multiple encoded measurements and achieved satisfying results \cite{gedalin2019deepcubenet,van2018compressed,Li2021_ICCV_Deblur}. A reinforcement learning based auto focus method has been proposed in \cite{wang2021deep}. 
Taking one step further, the sampling in video compressive imaging can be optimized to achieve sufficient spatio-temporal sampling of {sequential} frames at the maximal capturing speed~\cite{iliadis2016deepbinarymask}, and learning the sensing matrix to optimize the mask can further improve the quality of the compressive sensing reconstruction algorithm~\cite{spinoulas2015sampling}.

In addition to SCI, deep learning based reconstruction has recently been widely used in other CI systems, such as 3D imaging with deep sensor fusion~\cite{lindell2018single} and lensless imaging~\cite{khan2020flatnet}\cite{Yuan18OE}.


\begin{table*}[ht]\centering
\small
\caption{Reference list of AI for CI}
\label{table1}
 \resizebox{1\textwidth}{!}
 {
\begin{tabular}{c|l|c}
\hline
\rule{0pt}{10pt} \textbf{AI for CI}& \makecell[c]{\textbf{Optimization Method}}& \textbf{{References}} \\
\hline
\multirow{5}*{\makecell*[c]{\quad \\AI improves \\quality and efficiency \\of CI system}}&\rule{0pt}{10pt} \makecell[l]{Use deep learning to decode the {high-dimensional} data from multiple encoded measurements} & \cite{gedalin2019deepcubenet}\cite{van2018compressed}  \\
\cline{2-3}

 &\rule{0pt}{10pt}  \makecell[l]{Propose a deep neural network based on a standard tensor ADMM algorithm}&  \cite{ma2019deep}\\

\cline{2-3}
 & \rule{0pt}{15pt} \makecell[l]{Build a video CI system using a digital micro-mirror device and develop end-to-end \\convolutional neural networks for reconstruction} &  \cite{qiao2020deep} \\
 
\cline{2-3}
 & \rule{0pt}{15pt} \makecell[l]{Build a multispectral endomicroscopy CI system using coded aperture plus disperser and\\ develop end-to-end convolutional neural networks for reconstruction} &  \cite{Meng2020_OL_SHEM} \\
 
\cline{2-3}
 & \rule{0pt}{10pt} \makecell[l]{Propose a new auto-focus mechanisms based on reinforcement learning} &  \cite{wang2021deep} \\

\hline

\multirow{4}*{\makecell[c]{\quad \\AI optimizes \\structure and design\\ of CI system}}& \rule{0pt}{10pt}{{ Assist in designing} the physical layout of imaging system for compact imaging} & \cite{horstmeyer2017convolutional} \\
\cline{2-3}

 &\rule{0pt}{10pt}{ Perform end-to-end optimization of an optical system} & {\cite{chang2019deep}\cite{Metzler_2020_CVPR}\cite{sitzmann2018end,dun2019joint,han2011novel,Bergman:2020:DeepLiDAR}} \\
\cline{2-3}

 & \rule{0pt}{10pt}{ Introduce special optical elements, or replace the original ones for lightweight system}& \cite{peng2019learned,banerji2019diffractive,shedligeri2017data,wang2019measurement}\\
\cline{2-3}
 & \rule{0pt}{10pt}{ Show better results by novel end-to-end network frameworks via optimization}& \cite{jeon2019compact}\cite{wang2018hyperreconnet,zhang2020optimization,inagaki2018learning,akpinar2019learning} \\
\hline

\multirow{2}*{\makecell[c]{ AI promotes scene \\ adaptive CI system}}&\rule{0pt}{20pt}\makecell[l] {\ Propose the concept of pre-deep-learning that uses random projections for information\\ \ encoding, and reduces the number of required measurements to achieve high quality\\ \ imaging for specific tasks} & \cite{rao2013context,ashok2008compressive,abetamann2013compressive,ashok2008task}\\
\cline{2-3}

 &\rule{0pt}{15pt} \makecell[l]{Provide powerful means for scenes adaptive acquisition by upper-deep-learning combined\\ with public database and open source }& \cite{rawat2015context}\cite{rawat2016clicksmart} \\
\hline

\multirow{3}*{\makecell[c]{AI guides high-level \\ task of CI system}}&\rule{0pt}{15pt}\makecell[l]{\  Use scene's random-projections as coded measurements and embed the random features\\ \ into a bag-of-words model} & \cite{liu2012texture} \\
\cline{2-3}

 & \rule{0pt}{10pt}{ Introduce multiscale binary descriptor for texture classification and obtain improved robustness} & \cite{liu2016median}\\

\cline{2-3}
 &\rule{0pt}{10pt}{ Other application scenarios of task oriented CI} & \cite{bian2016robust}\cite{zhang2014fast}\\
\hline

\end{tabular}
}
\end{table*}

\subsection{AI Optimizes Structure and Design of CI System}
\label{sec:AIforCIsecond}

Instead of designing the system in an ad-hoc manner and adopting AI for {high-quality} or efficient reconstruction (previous subsection), recently, some researchers proposed closer integration between CI and AI. In particular, as shown in the second plot of Fig.~\ref{fig:Fourtypes}, by optimizing the encoding and decoding jointly, AI can assist in designing the physical layout of the system \cite{horstmeyer2017convolutional} for compact and high-performance imaging, such as extended depth of field, {auto focus} and high dynamic range.

One method is performing the end-to-end optimization\cite{chang2019deep} of an optical system using a deep neural network. A completely differentiable simulation model that maps the real source image to the (identical) reconstructed counterpart was proposed in \cite{sitzmann2018end}. This model jointly optimizes the optical parameters (parameters of the lens and the diffractive optical elements) and image processing algorithm using automatic differentiation, {which achieved achromatic extended depth of field and snapshot super-resolution.}  By collecting feature vectors consisting of focus value increment ratio and comparing them with the input feature vector, Han et al.~\cite{han2011novel} proposed a training based auto-focus method to efficiently adjust lens position. Another computational imaging method  achieved high-dynamic-range capture by jointly training an optical encoder and {algorithmic} decoder~\cite{Metzler_2020_CVPR}. Here the encoder is parameterized as the point spread function (PSF) of the lens and the decoder is a {CNN}. Similar ideas can also be applied on a variety of devices. For instance, a deep adaptive sampling LiDAR was developed by Bergman et al. \cite{Bergman:2020:DeepLiDAR}.

Different from above systems employing existing CNNs, improved deep networks have also been developed to achieve better results in some systems. For example, based on CNN, Wang et al. \cite{wang2018hyperreconnet} proposed a unified framework for coded aperture optimization and image reconstruction. Zhang et al. \cite{zhang2020optimization} proposed an optimization-inspired explainable deep network composed of sampling subnet, initialization subnet and recovery subnet. In this framework, all the parameters are learned in an end-to-end manner. Compared with existing state-of-the-art network-based methods, the OPINE-Net not only achieves high quality but also requires fewer parameters and less storage space, while maintaining a real-time running speed. This method also works in reconstructing a high-quality light field through a coded aperture camera \cite{inagaki2018learning} and depth of field extension \cite{akpinar2019learning}.

Another method is to introduce some special optical elements into the light path for higher performance, or replace the original ones with lower weight implementation. Using only a single thin-plate lens element, a novel lens design \cite{peng2019learned} and learned reconstruction architecture achieved a large field of view.
Banerji et al. \cite{banerji2019diffractive} proposed a multilevel diffractive lens  design based on deep learning, which drastically enhances the depth of focus. 
In addition, design of the aperture pattern plays an essential role in imaging systems. In view of this situation, a data driven approach  has been proposed to learn the optimal aperture pattern where coded aperture images are simulated from a training dataset of all-focus images and depth maps \cite{shedligeri2017data}.
For imaging under ultra-weak {illumination}, Wang et al. \cite{wang2019measurement} proposed a single photon compressive imaging system based on single photon counting technology. An adaptive spatio-spectral imaging system has been developed in~\cite{saragadam2021sassi}.

One significant progress of AI optimizing CI is the neural sensors proposed by Martel et al.~\cite{martel2020neural}, which can learn the  pixel exposures for HDR imaging and video compressive sensing with programmable sensors. This will inspire more research in designing advanced CI systems using emerging AI tools.

\subsection{AI Promotes Scene Adaptive CI System}
\label{sec:AIforCIthird}
Taking one step further, imaging systems should adapt to different scenes, where AI can play significant roles as shown in the third plot in Fig.~\ref{fig:Fourtypes}.
Conventional imaging systems use a uniform acquisition for varying environments and target tasks. This usually ignores the diverse results in improper camera settings, or records a large amount of data irrelevant to the task which are discarded in the successive processing.
Bearing this in mind, the imaging setting of an ideal CI system should actively adapt to the target scene, and exploit the task-relevant information~\cite{gupta2010flexible}.

Pre-deep-learning studies in this direction often use optimized random projections for information encoding, and attempt to reduce the number of required measurements for specific tasks. Rao et al. \cite{rao2013context} proposed a compressive optical foveated architecture which adapts the dictionary structure and compressive measurements to the target signal, by reducing the mutual coherence between the measurement and dictionary, and increasing the sparsity of representation coefficients. Compared to conventional Nyquist sampling and compressive sensing-based approaches, this method adaptively extracts task-relevant regions of interest (ROIs) and thus reduces meaningless measurements. Another task-specific compressive imaging system employs generalized Fisher discriminant projection bases to achieve optimized performance \cite{ashok2008compressive}. It analyzes the irrelevant performance of target detection tasks and achieves optimized task-specific performance. In addition, adaptive sensing has also been applied to ghost imaging \cite{abetamann2013compressive} and iris-recognition \cite{ashok2008task}.

Inspired by deep learning, machine vision now has become a more powerful approach for scene adaptive acquisition. 
As camera parameters and scene composition play a vital role in aesthetics of a captured image, researchers can now use publicly available photo database, social media tips, and open source information for reinforcement learning to improve the user experience~\cite{rawat2015context}. ClickSmart \cite{rawat2016clicksmart}, a viewpoint recommendation system can assist users in capturing high quality photographs at well-known tourist locations, based on the preview at the user's camera, current time and geolocation. Specifically, composition learning is defined according to factors affecting the exposure, geographic location, environmental conditions, and image type, and then provides adaptive viewpoint recommendation by learning from the large {high-quality} database. Moreover, the recommendation system also provides camera motion guidance for pan, tilt and zoom to the user for improving scene composition.
An end-to-end optimization of optics and image processing for achromatic extended depth of field and super-resolution imaging framework was proposed in~\cite{sitzmann2018end}. Towards correcting large-scale distortions in computational cameras, an adaptive optics approach was proposed in ~\cite{wang2018megapixel}.

\subsection{AI Guides High-level Task of CI System}
\label{sec:AIforCIlast}
As mentioned in the introduction, instead of capturing more data, CI systems will focus on preforming specific tasks and AI will help to maximize the performance, which is illustrated in the bottom plot of Fig.~\ref{fig:Fourtypes}.
Specifically, computational imaging optimizes the design of the imaging system by considering the afterwards processing. Towards this end, one can build systems optimized for specific high-level vision tasks, and conduct analysis directly from the coded measurements~\cite{Lu20SEC}. Such method can capture information tailored for the desired task, and thus is of improved performance and largely reduced bandwidth and memory requirements. Taking the properties of the target scene into system design, we can also get high flexibility to the environments, which is important for mobile platforms.

Task-oriented CI is an emerging ongoing direction and there only {exist} some preliminary studies during the writing of this article.
Texture is an important sensor property of nature scenes and provides informative visual cues for {high-level} tasks. 
Traditional texture classification algorithms on 2D images are widely studied but they occupy a large bandwidth, since object with a complicated texture is hard to compress. 
To achieve texture classification from a small number of measurements, an approach was proposed in \cite{liu2012texture} to use scene's random-projections as coded measurements and {embed} the random features into a bag-of-words model to perform texture classification. This approach brings significant improvements in classification accuracy and reductions in feature dimensionality.
Differently, Liu et al. \cite{liu2016median}  introduced a multiscale local binary patterns (LBP) descriptor for texture classification and obtained improved robustness.
The classification from coded measurements is also proved to be applicable for other kinds of data \cite{bian2016robust}, such as hyperspectral images. 
Besides texture classification, Zhang et al. \cite{zhang2014fast} proposed an appearance model tracking algorithm based on extracted multi-scale image features, which performs favorably against state-of-the-art methods in terms of efficiency, accuracy and robustness in target tracking.

\begin{figure*}[ht]
\centering
\includegraphics[width=\linewidth]{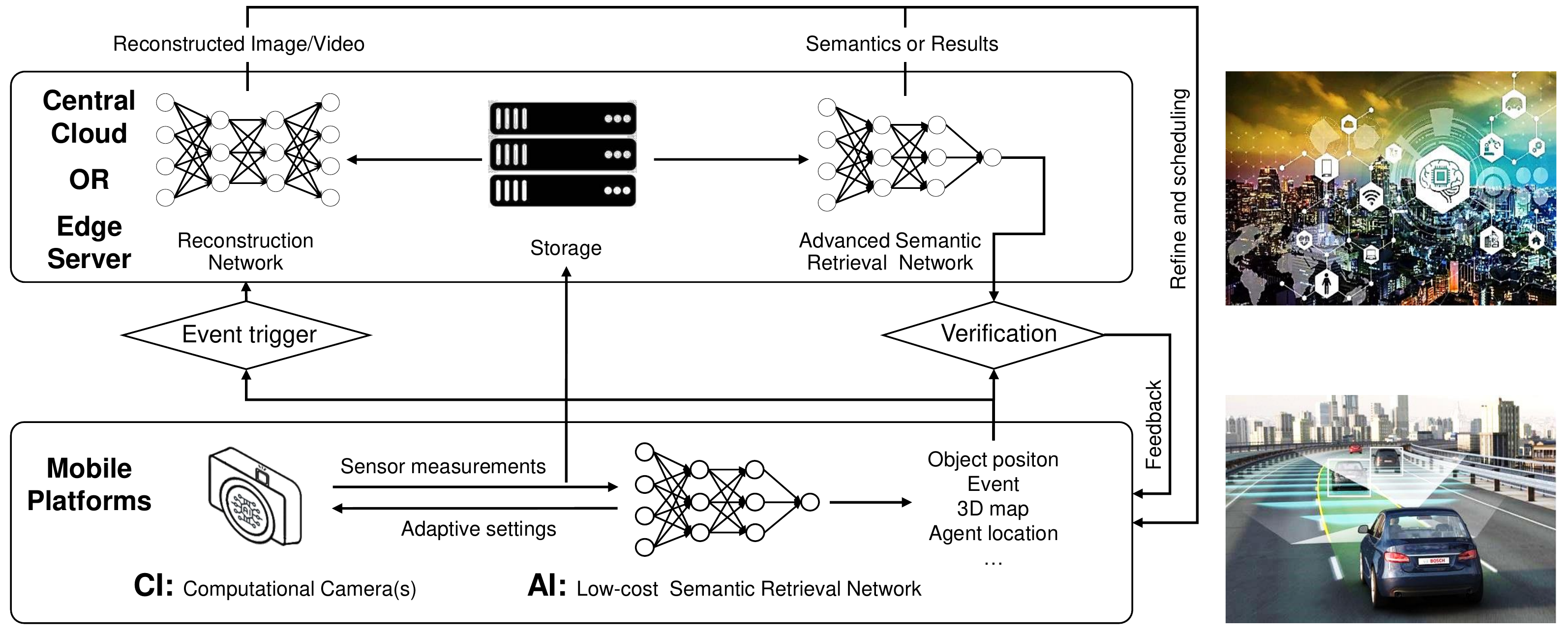}
\caption{{Proposed framework} of visual processing solutions for mobile platforms based on computational imaging, artificial intelligence, cloud computing (by the central cloud) and edge computing (by the edge server) architecture.}
\label{fig:framework}
\end{figure*}

\section{CI and AI: Outlook \label{Sec:Outlook}}

In the previous section, we have shown the different preliminary integration approaches of CI and AI, and the advantageous over systems using conventional cameras. However, in our daily life, due to the limitations of load capacity, space and budget, the current CI + AI methods still cannot provide ready machine vision technologies for common mobile platforms. Since combining CI and AI often requires a complex CI system and AI algorithms involving heavy calculations, how to integrate CI and AI compactly to form a general and computing efficient device has become an urgent challenge.
Fortunately, with the recent advances of mobile communication (5G), edge and cloud computing, big data analysis and deep learning based recognition and control algorithms, it is feasible now to tightly integrate CI and AI in an unprecedented manner.
In this section, using the example of self-driving vehicles, we propose a new framework in Fig.~\ref{fig:framework} for CI + AI as well as cloud computing and big data analysis, \etc.


For a vehicle equipped with computational cameras, the CI setup extracts effective visual information from the environment by encoded illumination and acquisition. Next, the artificial intelligence methods, in which deep learning now becomes the mainstream, are applied to process the collected data to retrieve semantic information, such as 3D geometry,  with high efficiency and accuracy. Then, the semantic information assists decision-making and actions of intelligent mobile platforms.

Nevertheless, processing at each node (mobile platforms such as cellphones and local vehicles) faces multiple challenges. Taking the self-driving vehicle as an example, each node (vehicle) needs to process a large amount of data from computational cameras and other sensors, such as radar and LiDAR, in real time. Besides, it also needs to combine these information for decision making and action. As mentioned before, the advantageous of CI + AI systems come at expenses of complex (sometimes heavy) setup and  high computational cost. Additionally, a large amount of data need to be stored for pattern recognition and learning, which demands a large amount of storage. Therefore, an autonomous driving device needs to be equipped with high-performance processing units and large storage, which might either exceed the load budget, or take too long response time that is fatal for mobile platforms.

Moreover, mobile vision systems usually in a  coexist  scenario and communications are required. Cloud computing might be a plausible option for the mutual communication. However, processing the enormous data from the big groups of devices is still challenging, and limited network bandwidth leads to serious network congestion.
To cope with this problem, the participation of various technologies mentioned in the introduction is required. Among them, the edge computing technology provides a promising combination of CI and AI together with the rapid development of communication technologies. With a top-down architecture, most of the data are consumed at the edge of the mesh structure and network traffic is distributed \cite{shi2016edge} and can be performed in parallel.

In the following, we propose a framework to deeply integrate CI and AI by using autonomous driving as an example to demonstrate the structure. After that, we extend the proposed framework to other mobile platforms.

\subsection{Motivation: Requirement of Mobile Vision Systems}

Since the mobile platform is usually of limited size and resources, the imaging system on it needs to be compact, lightweight, and low power consumption. Meanwhile, some mobile platforms have high requirements to conduct real-time operation, and thus build the rapid observation-orientation-decision-action loop. Therefore, both miniaturized imaging system and efficient computing are required to meet the needs of system scale, power consumption and timely response. For effective computing on mobile vision platforms, we need to develop lightweight algorithms or employ distributed computing techniques, in which 5G and edge/cloud computing are combined to build a multi-layer real-time computing architecture.

Towards this end, we propose a hierarchical structure of {edge computing} for real-time data processing and navigation of the mobile platforms shown in Fig. \ref{fig:framework}. In this structure, the central cloud or the edge server (such as a road side unit and cellular tower) communicates with each vehicle, forming a typical vehicle-to-infrastructure (V2I) scenario \cite{liu2019edge}. Under this proposed framework, the problems urging to be solved are distributed at two levels of the architecture.

\vspace{3mm}
\subsection{Mobile Platforms}
At the level of a {\em single vehicle}, reliable and low-cost tasks are required for {real-time} navigation under limited resources. Firstly, we need to develop specific (such as energy efficient) algorithms to finish various real-time tasks from the encoded measurements of CI cameras, which might be more difficult than those from conventional cameras. Recently, researches on vehicle detection and tracking algorithms have made significant progress. Girshick et al. \cite{girshick2015region, girshick2015fast, ren2015faster} proposed a series of general target detection algorithms based on convolutional neural networks. Among them, the detection method for road vehicles based on Faster R-CNN network \cite{ren2015faster} has made some breakthrough. Secondly, the limitations of the available computing facilities equipped with mobile platforms call for low-power AI algorithms, \ie, networks applicable on mobile platforms.
With the rapid development of convolutional neural network and the popularization of mobile devices, lightweight neural networks have been proposed with the advantages of the customized and lightweight structure, thus reducing the resource requirements on the specific platform.
For example, some methods build miniature neural networks based on parameter reconstruction to reduce the storage cost. Among the prominent representatives are MobileNet and ShuffleNet, which make it possible to run neural network models on mobile platforms or embedded devices. {MobileNet consists of a smaller parameter scale by using depth-wise separable convolution instead of standard convolution to achieve a trade-off between latency and accuracy  \cite{howard2017mobilenets, zhuinverted, howard2019searching}}. At the same time, it can realize many applications on mobile terminals such as target detection, target classification and face recognition. ShuffleNet is mainly characterized by high running speed, which improves the efficiency of implementing this model on hardware \cite{zhang2018shufflenet}\cite{ma2018shufflenet}. Differently, some methods achieve smaller sizes by trimming the existing network structure. In this research line, YOLO, YOLOV2 and YOLOV3 algorithms are a series of general target detection models proposed by Joseph et al. \cite{redmon2016you, redmon2017yolo9000, redmon2018yolov3}, among which Tiny YOLOV3 is a simplification of YOLOV3 model incorporating feature pyramid networks \cite{lin2017feature} and fully convolutional networks \cite{long2015fully}; it has a simpler model structure and higher detection accuracy, which is potential to be more suitable for real-time target detection tasks on mobile platforms, leading to effectively finishing various decision-making activities.
We anticipate that more specific task-driven networks will be developed with increasing deployment of CI systems.

\subsection{Edge Server and/or Central Cloud}
At the level of {\em edge cloud}, which can be a road side unit or a cellular tower having more power than the mobile platform, the data and action from the single vehicle are processed and transmitted to the edge cloud for optimized global control. Generally, the edge cloud computers execute communications that need to be dealt with immediately, such as real-time location and mapping \cite{mao2017survey}, or some resources demanding calculation infeasible on single vehicle.
The edge servers are mainly triggered by certain events at the bottom layer containing a set of mobile platforms (single vehicles).
For example, at some key time instants (when abnormal or emergent events happen), the edge servers conduct full frame reconstruction based on some measurements and characteristics from the platforms, and transmit the reconstruction to the central cloud.
In addition, the edge servers will also conduct cross-detection and verification for the acquisition and decision at single vehicles. By comparing the reconstructed results with the detection results obtained by an energy efficient algorithm on the mobile platform, the edges can provide better data {for} the central cloud for planning and traffic control.

At the level of {\em central cloud}, eventually, cloud computing technology provides a large amount of computing power, storage resources and databases to assist the complex large-scale computing tasks and make decisions. This will help the high-level management and large-scale planning.

{{Cloud Computing and Edge Computing}} play roles in a manifold way in the central cloud and edge server in addition to promote information processing and mechanical controlling through streaming with local processing platforms. {Firstly}, it can be used to control the settings of computational cameras for better acquisition \cite{zhang2009intelligent} as well as to manage large-scale multi-modality camera networks by compressive sensing to reduce the data deluge \cite{mitra2014toward}. 
Secondly, the modularized software framework composed of cloud computing and edge computing can effectively hand out multiple image processing tasks to distributed computing resources. For example, one can  only transmit the pre-processed image information to avoid insufficiency on storage, bandwidth, and computing power of mobile platforms \cite{bistry2010cloud}.
Thirdly, at the decision-making stage, vehicle cloud (and edge) computing technology has changed the ways of vehicle communication and underlying traffic management by sharing traffic resources via network for more effective traffic management and planning. This has wide applications in the field of intelligent transportation \cite{ahmad2016role}. Similarly, cloud (and edge) computing should also be applied to control drone fleet formation to improve real-time performance.

In summary, this architecture is designed for effective computations and resource distributions under energy and resource constraints \cite{liu2019edge}. Moreover, it is robust to recover from the failure of a part of the edge servers and vehicles. Cloud-based edge computing technology is important in all aspects of the new mobile vision framework, as Fig. \ref{fig:framework} depicts the close connections.
Each research topic depicted in the plot will play an important role in the integration of CI and AI, thus leading to the next revolution of machine vision.

\subsection{Extensions / Generalization to Related Systems}
We expect to deploy our proposed framework to intelligent mobile platforms in various fields in the near future. Self-driving car is the intelligent transportation vehicle that will most likely be put into use~\cite{Lu20SEC}; meanwhile, unmanned aerial vehicle (UAV), satellite and unmanned underwater vehicle will be applied in military or civilian scenes. Different mobile platforms have varying loads and requirements for imaging dynamic range, resolution and frame rate, to perform different tasks such as detection, tracking and navigation. 
In addition to the single mode, some mobile platforms are also required to cooperate with other similar platforms, such as drone swarms, which poses a more complex challenge to the integration of CI and AI on intelligent mobile platforms. By combining the latest communication technology such as 5G (or  6G) and big data technology, we expect the proposed framework to be extended on the mobile platform cluster such as the the autonomous vehicle fleets.

\vspace{3mm}
\noindent{\bf{\em{(1) Single Mobile Platform}}}

The working mode of a single mobile platform can be simplified to the pipeline from sensor input to the execution of decision-making. As shown in  Fig.~\ref{fig:framework}, the layered cloud/edge computing framework distributes and asynchronously executes computing tasks on both the platform and the cloud/edge server based on the response time and resource consumption. The final decision-making module collects the reconstruction and extracted semantics to guide the next operating step of the mobile platform, which is supported by multiple technologies, such as reinforcement learning and big data analysis.

{\bf Reinforcement Learning} (RL) can be applied to the overall scheduling (such as arrangement of road traffic) and the feedback control of a single robot (such as the vehicle behavior), thus finishing the closed loop for mobile vision systems. 
Specifically, RL can be used to design complex and hard-to-engineer behaviors through trial-and-error interactions with the environment, thus providing a framework and a set of tools for robotics (vehicles) \cite{kober2013reinforcement}.
{Generally speaking, in mobile platforms, RL helps to learn how to accomplish tasks that humans cannot demonstrate, and promote the platform to adapt to new, previously unknown external environments. Through these aspects, RL equips {vehicles} with previously missing skills \cite{kormushev2013reinforcement}.}

{One category of state-of-the-art RL algorithms in robotics is the policy-search RL methods, which avoid the high dimensionality of the traditional RL methods by using a smaller policy space to increase the convergence speed. Algorithms in this category include the policy-gradient algorithms \cite{peters2008natural}, Expectation-Maximization (EM) based algorithms \cite{kober2009learning} and stochastic optimization based algorithms \cite{theodorou2010generalized}, providing different performances when adapting to various environments. These RL algorithms are usually based on the input of the images or other sensor data from the  environment, and use a reward to optimize the strategy. Again taking self-driving as an example, an RL method using policy gradient iterations was proposed in  \cite{shalev2016safe}, which takes driving comfort as the expected goal and takes safety as a hard constraint, thereby decomposing the strategy into a learning mapping and an optimization problem. {Isele et al.} \cite{isele2018navigating} applied a branch of RL, specifically Q-learning, to provide an effective strategy to safely pass through {intersections without traffic signals} and to navigate in the event of occlusion. {Zhu et al.} \cite{zhu2018human} applied the depth deterministic policy gradient algorithm by using speed, acceleration and other information as input and constantly updating the reward function of the degree of deviation from historical experience data, and proposed an autonomous car-following planning model.}  

In summary, in the framework of RL, the agent, such as the self-driving car, achieves specific goals by learning from the environment. However, {challenges still exist} in smoothness, safety, scalability, adaptability, \etc \cite{kormushev2013reinforcement}, and hence RL needs dramatic efforts and researches to lead to its applications in mobile vision.

{\bf Big Data Analysis} is crucial for both action instructions to single mobile intelligent platforms and complex traffic or cluster dispatching by using large-scale mobile data to characterize and understand real-life phenomena in mobile vision, including individual traits and interaction patterns. 
As one of the most promising technologies to break the restrictions in current data management, big data analysis can help the central cloud in decision making for single vehicle from the following aspects \cite{han2015mobile}: 
\begin{itemize}
    \item  Process massive amounts of data efficiently for internal operations and interactions of mobile platforms involving large amounts of data traffic.
    \item  Accelerate decision-making procedure to reduce the risk level and promote the users’ experience, which is crucial for reliable and safe decisions of mobile platforms.
    \item  Combine with machine learning or data mining techniques to reveal new patterns and values in the data, thereby helping to predict and prepare strategies for possible future events.
    \item  Integrate and compactly code the large variety of data sources, including mobile sensor data, audio and video streaming data, and the database is still being updated online~\cite{yue2013cloud}.
\end{itemize}
As an example of applying big data analysis to assist AI computing under our framework of CI + AI, a method of image coding for mobile devices is designed to improve the performance of image compression based on Cloud Computing. Instead of performing image compression at the pixel scale, it does down-sampling and feature extraction respectively for the input image and encodes the features extracted with a local feature descriptor as feature vectors, which are decoded from the down-sampled image and matched with the large image database in the cloud to reconstruct these features. The down-sampled image is then used to stitch the retrieved image blocks together for the reconstruction of the compressed image. This method achieves higher image visual quality at thousands to one compression ratio and helps make safe and reliable decisions.
In addition, big data analysis is also expected to be employed in emerging 5G networks to improve the performance of mobile networks \cite{zheng2016big}.

\vspace{3mm}
\noindent{\bf{\em{(2) Mobile Platform Cluster}}}

Mobile-platform-cluster combines resources from single platform to perform specific large-scale tasks through cooperation, including traffic monitoring, earthquake and volcano detection, long-term military monitoring \etc \cite{jawhar2017communication}.
Similar to a single platform, the requirements of latency, delay and bandwidth need to be taken into account to fulfill specific tasks. For example, in the applications of precision agriculture \cite{tsouros2019review}\cite{adao2017hyperspectral} and environmental remote sensing \cite{niethammer2012uav}, the delay tolerance is high but requires high bandwidth \cite{jawhar2017communication}. These mobile platforms perform online coded capture and offline storage and processing such as image correction and matching \cite{niethammer2012uav} and other methods in the edge server. As a different example, in the applications of surveillance, one needs to conduct real-time object detection and tracking\cite{opromolla2019airborne}, which can be executed on the edge server and send specific action instructions back to the mobile platform, such as adjusting the speed according to the deviation from the reference \cite{sampedro2018image}, or the position of other platforms and objects \cite{sridhar2016target}. Such tasks demand low latency, but are not bandwidth starving.

In general, mobile-platform-cluster scheduling algorithms can be categorized into centralized and distributed ones considering the resource distribution, power consumption, collision avoidance, task distribution and surrounding environments. Centralized cluster scheduling is convenient to consider the location relationship among all mobile platforms and all tasks or coverage areas in the same coordinate system~\cite{dantu2011programming}, and make timely and rapid dynamic adjustment in case of failure of some nodes \cite{duan2020dynamic}, but the amount of calculation is large, which is suitable to be performed in the edge server according to our proposed framework. Under various constraints, different improved models have been proposed~\cite{jin2020uav,dai2019deploy,tanil2013collaborative,shima2006multiple,sharma2009collision}.
In distributed cluster scheduling, the positioning of peers and the identification of environment are the basis of autonomous navigation and scheduling. Therefore, the accurate peer position estimation is important and diverse algorithms have been proposed using different techniques~\cite{clark2017autonomous,saska2014autonomous,saska2016swarm,sanchez2020semantic,sanchez2019deep,saska2014swarms,schleich2013uav,messous2016network}. The position information is then used for route planning~\cite{sanchez2019real}, navigation~\cite{sampedro2018laser} and target tracking~\cite{atten2016uav}, where the RL have been used to improve the performance~\cite{yang2019application}\cite{wang2019reinforcement}. Wu et al. \cite{wu2019couav} have proposed an integrated platform using the collaborative algorithms of mobile platform cluster represented by UAV, including connection management and path planning; this paves the way for the implementation of our framework.

\section{CI + AI + X:  Win-Win Future by Collaboration \label{Sec:X}}

\begin{figure}[ht]
\centering
\includegraphics[width=0.9\linewidth]{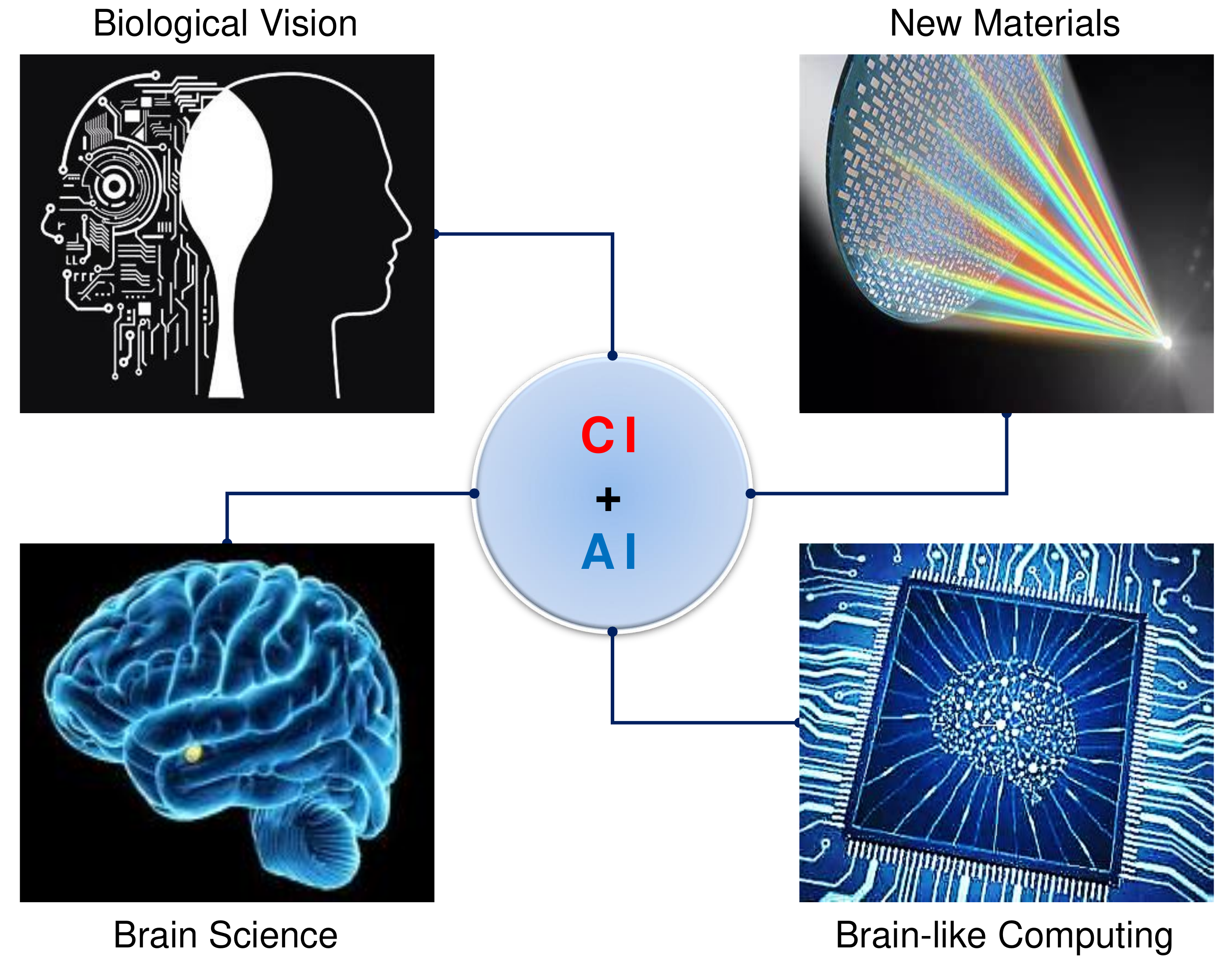}
\caption{Win-win collaborations among AI, CI and other fields. 
}
\label{fig:winwin_X}
\end{figure}

For the future of mobile vision systems, it is as important as strengthening ``CI + AI" integration to extending the integration to new fields for a win-win collaboration. On one hand, the CI acquisition, AI processing and their integration would benefit from a broad range of disciplines. For example, {the way that biological vision systems record and process visual information} is very delicate. Future mobile vision can draw on the principles of biological vision and brain science to meet the developing trend of high-precision, high-speed and low-cost next generation imaging systems (Fig.~\ref{fig:winwin_X}). The hardware development in chips, sensors, optical elements would promote compact lightweight implementations.
On the other hand, mobile vision systems are of wide applications and can serve as a testing bed for achieving strong artificial intelligence \cite{searle1980minds} in specific tasks.

\subsection{Biological Vision for CI + AI}
After millions of years  of evolution, biological vision systems are of delicate and  high performance, especially in terms of semantic perception. Thus, reverse engineering for biological vision systems or biomimetic intelligent imaging are of important guidance. 
As an inspiration, current research on computational imaging draws on the mechanisms of biological vision systems, including humans.

For example, borrowing the ``Region of Interest" strategy in human visual system, computational visual attention systems selectively detect regions of interest in the target scene, and have great applications in fields like computer vision, cognitive systems and mobile robotics \cite{frintrop2010computational}. The event camera draws inspiration from the transient information pathway of ``photoreceptor cells-bilevel cells-ganglion cells" and records the event of abrupt intensity changes, instead of {conducting} regular sampling in space and time domains. By directly performing detection from the event streams~\cite{barua2016direct}, such sensing mechanism obtains qualitative improvement in both perception speed and sensitivity \cite{kim2014simultaneous}\cite{mueggler2017event}, especially under {low-light} conditions.
As a silicon retina, an event stream from an event camera can be used to track accurate camera rotation while building a persistent and {high-quality} mosaic of a scene which is super-resolved and of high dynamic range \cite{kim2014simultaneous}.
Using only an event camera, depth estimation of features and six degrees of freedom  pose estimation can be realized, thus enhancing the 3D dynamic reconstruction \cite{kim2016real}.

{In addition}, evidence from neuroscience suggests that the human visual system uses a segmentation strategy based on identifying discontinuities and grouping them into contours and boundaries. This strategy provides better ideas for image segmentation, and some work shows that the perceptual performance of visual contour grouping can be improved through the mechanism of perceptual learning \cite{medathati2016bio}.
Moreover, the information about the object and self-motion in the human brain has realized a fast, feedforward hierarchical structure. The visual processing unit integrates the motion to calculate the local moving direction and speed, which has become an important reference for the research of optical flow~\cite{medathati2016bio}.

{Due to the incomplete understanding of biological vision, adopting it to machine vision is still limited to individual optical elements and the overall imaging mechanism still follows the traditional imaging scheme, and thus the system is of high complexity, large size and high power consumption, which limits the mobile applications.
{Based on} the structural and functional analysis of the biological vision system, brain-like intelligence {for} the machine vision system might be a good research direction for the next generation mobile vision.}

\subsection{Brain Science for CI + AI}

The information path of the human visual system eventually leads to the visual cortex in the brain. Since Hubel and Wiesel's ground-breaking study of information processing in the primary visual system of cats in 1959 \cite{hubel1998early}, researchers have initially understood the general characteristics of neuron responses at various levels from the retina to the higher-order visual cortex, establishing effective visual computing models of the brain. 
{Due to the huge advantages of the human brain in response speed, energy consumption, bandwidth usage and accuracy,} learning the information processing mechanism and getting inspirations {from} the structure of the brain might push forward the development of artificial intelligence. 

The available knowledge is that biological neurons have complex dendritic structures \cite{kumar2010spiking}, which is the structural basis for its ability to efficiently process visual information. {Existing artificial neural networks, including deep learning  can be recognized as an oversimplification of biological neural networks \cite{kriegeskorte2015deep}.} At the level of the neural circuit, each neuron is inter-connected with other neurons through thousands or even tens of thousands of synapses, where procedures similar to the back propagation in deep learning are performed to modify the synapses to improve the behaviours\cite{lillicrap2020backpropagation}, and the number of inhibitory synapses often significantly exceeds that of excitatory synapses. This forms a strong non-linearity as the physical foundation on which the brain can efficiently complete basic functions such as target detection, tracking, and recognition, and advanced functions such as visual reasoning and imagination. 
{Using neural networks to simulate the information processing strategy of biological neurons and synapses, brain-like computing has become the new generation of low-cost computing. }


\subsection{Brain-like Computing for CI + AI}
In the next generation of mobile vision, the major function of computers will transform from calculation to intelligent processing and information retrieval.
On one hand, intelligent computational imaging system is no longer for recording, but to provide robust, reliable guidance for intelligent applications.
This processing will be computing intensive and of huge power consumption\cite{hennessy2019new}. On the other hand, the mobile vision systems must meet the demanding requirement from real applications such as high throughput, high frame rate, weak illumination, \etc. and  of low computation cost to work under limited lightweight budget. Therefore, optoelectronic computing 
has become a hot topic due to its stronger computing power and potentially lower power consumption. 

For common convolutional neural networks, 
the number of parameters and nodes in CNNs increases dramatically with the performance requirement, thus the power consumption and memory demands grow correspondingly. Recently, the hybrid architecture combining optoelectronic computing and traditional neural networks is used to improve current network performance in terms of speed, accuracy, and energy consumption.
Benefiting from the broad bandwidth, high computing speed, high inter-connectivity, and inherent parallel processing characteristics of optical computing, the Optical Neural Network (ONN) performs expensive matrix multiplication of a fully connected layer by optics to increase the computation speed \cite{psaltis1988adaptive}.
Further, by placing an optical convolution layer for computational imaging systems at the front end of the CNN, the network performance can be maintained with significantly reduced energy consumption \cite{chang2018hybrid}. In addition to locally improving the computational performance of deep networks, a photonic neurosynaptic network based on wavelength division multiplexing technology has recently been proposed \cite{feldmann2019all}. Different from traditional computing architecture, this all-optical network processes information in a more brain-like manner, which enables fast, efficient, and low-energy computing.

Recently, a new type of brain-like computing architecture, hybrid Tianjic chip architecture that draws on the basic principles of brain science was proposed. As a brain-like computing chip that integrates different structures, Tianjic can support neural network models in both computer science and neuroscience \cite{pei2019towards}, such as artificial neural networks and spiking neural networks~\cite{gibson1999two}\cite{hodgkin1952currents}, to demonstrate their respective advantages. Compared with the current world-leading chips, Tianjic is flexible and expandable, meanwhile enjoying the high integration level, high speed, and large bandwidth.
{Inspired by this, brain-like computing technologies might significantly} aid self-driving, UAVs, and intelligent robots and also provide higher computing power and low-latency computing systems for the internet industry.

\subsection{New Materials for CI + AI}

Using coded acquisition scheme, CI systems often involve complex optical design and modulation. It is desirable to adopt lightweight modules on mobile platforms. For instance, metasurfaces are surfaces covered by ultra-thin plasmonic structures \cite{kildishev2013planar}. Such emerging materials have the advantageous of being thin but powerful of controlling the phase of light, and have been widely used in optical imaging.
Ni et al. \cite{ni2013metasurface} proved that the ultra-thin metasurface hologram with 30nm thickness can work in the visible light range. This technology can be used to specifically modulate the amplitude and phase to produce high-resolution low-noise images, which serves as the basis of new ultra-thin optical devices.
Further, Zheng et al. \cite{zheng2015metasurface} demonstrated that the geometric metasurfaces consisting of an array of plasmonic nanorods with spatially varying orientations can further improve the efficiency of holograms.
In addition, the gradient metasurface based on this structure can serve as a two-dimensional optical element, {such as ultra-thin gratings and lenses \cite{lin2014dielectric}.}
The combination of semiconductor manufacturing technology and metasurfaces can effectively improve the quality and efficiency of imaging and greatly save space. For example, by using metasurface, Holsteen et al.~\cite{holsteen2019light} improved the spatial resolution of the three-dimensional information collected by ordinary microscopes. The ultra-thin nature of metasurfaces also provides other advantages such as higher energy conversion efficiency \cite{ma2014acoustic}. In summary, the flat and highly integrated optical elements made of metasurface can achieve local control of phase, amplitude, and polarization, and have the properties of high efficiency and small size. Therefore, metasurfaces are expected to replace traditional optical devices in many fields including mobile vision or wearable optics to overcome some distortions or noise, as well as to improve imaging quality \cite{capasso2017metasurfaces}.

The fields we list above will play hybrid (sometimes complementary) roles to enhance the future mobile vision systems with CI and AI. For example, the researches on biological vision and brain science promote the recognition for biological system, thus help the development of brain-like computing, and eventually will be used in mobile vision system to improve the performance.
\section{Discussion}
{In current digital era, mobile vision is playing an important role.} 
Inspired by vast applications in computer vision and machine intelligence, computational imaging systems start to change the way of capturing information. Different from the style of ``capturing images first and processing afterwards" used in  conventional cameras, in CI, {more information can be captured in an indirect way.} By integrating AI into CI as well as the new techniques of 5G and beyond plus edge computing, we believe a new revolution of mobile vision is around the corner.

This study reviewed the history of mobile vision and presented a survey of using artificial intelligence to computational imaging. This combination inspires new designs, {new capabilities, and new applications} of imaging systems. 
As an example, a novel framework has been  proposed for self-driving vehicles, which is a representative application for this new revolution.
The integration of CI and AI will further inspire new tools for material analysis, diagnosis, healthcare, virtual/augmented/mixed reality, \etc.  


\bibliographystyle{IEEEtran}
\bibliography{reference1,reference0716,nlos,YmZhang,UAVs}

\begin{thebibliography}{100}
\providecommand{\url}[1]{#1}
\csname url@samestyle\endcsname
\providecommand{\newblock}{\relax}
\providecommand{\bibinfo}[2]{#2}
\providecommand{\BIBentrySTDinterwordspacing}{\spaceskip=0pt\relax}
\providecommand{\BIBentryALTinterwordstretchfactor}{4}
\providecommand{\BIBentryALTinterwordspacing}{\spaceskip=\fontdimen2\font plus
\BIBentryALTinterwordstretchfactor\fontdimen3\font minus
  \fontdimen4\font\relax}
\providecommand{\BIBforeignlanguage}[2]{{%
\expandafter\ifx\csname l@#1\endcsname\relax
\typeout{** WARNING: IEEEtran.bst: No hyphenation pattern has been}%
\typeout{** loaded for the language `#1'. Using the pattern for}%
\typeout{** the default language instead.}%
\else
\language=\csname l@#1\endcsname
\fi
#2}}
\providecommand{\BIBdecl}{\relax}
\BIBdecl

\bibitem{Boyle_Smith_CCD}
W.~S. {Boyle} and G.~E. {Smith}, ``Charge coupled semiconductor devices,''
  \emph{The Bell System Technical Journal}, vol.~49, no.~4, pp. 587--593, 1970.

\bibitem{Chidester1999Recording}
A.~Chidester, J.~Hinch, T.~C. Mercer, and K.~S. Schultz, ``{Recording
  automotive crash event data},'' in \emph{Transportation Recording: 2000 and
  Beyond, International Symposium on Transportation Recorders}, 1999, pp.
  85--98.

\bibitem{firstcamphoneweb}
J.~Callaham, ``The first camera phone was sold 20 years ago, and it’s not
  what you might expect,''
  \url{https://www.androidauthority.com/first-camera-phone-anniversary-993492/}.

\bibitem{Dalrymple2006Apple}
J.~Dalrymple, ``{Apple releases the MacBook},'' \emph{Macworld}, vol.~23,
  no.~7, pp. 20--21, 2006.

\bibitem{chen2014the}
S.~Chen, J.~Zhao, and Y.~Peng, ``The development of {TD-SCDMA} 3{G} to
  {TD-LTE}-advanced 4{G} from 1998 to 2013,'' \emph{IEEE Wireless
  Communications}, vol.~21, no.~6, pp. 167--176, 2014.

\bibitem{nilsson1984shakey}
N.~J. Nilsson, ``Shakey the robot,'' SRI International Menlo Park CA, Tech.
  Rep., 1984.

\bibitem{Jochem1995PANS}
T.~{Jochem}, D.~{Pomerleau}, B.~{Kumar}, and J.~{Armstrong}, ``{PANS: A
  portable navigation platform},'' in \emph{Intelligent Vehicles '95.
  Symposium}, 1995, pp. 107--112.

\bibitem{Altmann18Science}
Y.~Altmann, S.~McLaughlin, M.~J. Padgett, V.~K. Goyal, A.~O. Hero, and
  D.~Faccio, ``Quantum-inspired computational imaging,'' \emph{Science}, vol.
  361, no. 6403, 2018.

\bibitem{Mait18_AOP_CI}
J.~N. Mait, G.~W. Euliss, and R.~A. Athale, ``Computational imaging,''
  \emph{Advances in Optics and Photonics}, vol.~10, no.~2, pp. 409--483, Jun
  2018.

\bibitem{brady2020smart}
D.~J. Brady, M.~Hu, C.~Wang, X.~Yan, L.~Fang, Y.~Zhu, Y.~Tan, M.~Cheng, and
  Z.~Ma, ``Smart {cameras},'' \emph{arXiv preprint arXiv:2002.04705}, 2020.

\bibitem{lucas2018using}
A.~Lucas, M.~Iliadis, R.~Molina, and A.~K. Katsaggelos, ``Using deep neural
  networks for inverse problems in imaging: Beyond analytical methods,''
  \emph{IEEE Signal Processing Magazine}, vol.~35, no.~1, pp. 20--36, 2018.

\bibitem{rick2017one}
J.~Rick~Chang, C.-L. Li, B.~Poczos, B.~Vijaya~Kumar, and A.~C.
  Sankaranarayanan, ``{One network to solve them all--solving linear inverse
  problems using deep projection models},'' in \emph{IEEE International
  Conference on Computer Vision}, 2017, pp. 5888--5897.

\bibitem{Ongie2020_JSAIT}
G.~Ongie, A.~Jalal, C.~A. Metzler, R.~G. Baraniuk, A.~G. Dimakis, and
  R.~Willett, ``Deep learning techniques for inverse problems in imaging,''
  \emph{IEEE Journal on Selected Areas in Information Theory}, vol.~1, no.~1,
  pp. 39--56, 2020.

\bibitem{agrawal2016signal}
A.~Agrawal, R.~Baraniuk, P.~Favaro, and A.~Veeraraghavan, ``Signal processing
  for computational photography and displays [from the guest editors],''
  \emph{IEEE Signal Processing Magazine}, vol.~33, no.~5, pp. 12--15, 2016.

\bibitem{wetzstein2013plenoptic}
G.~Wetzstein, I.~Ihrke, and W.~Heidrich, ``On plenoptic multiplexing and
  reconstruction,'' \emph{{International Journal of Computer Vision}}, vol.
  101, no.~2, pp. 384--400, 2013.

\bibitem{sciammarella1982moire}
C.~A. Sciammarella, ``The moir{\'e} method—a review,'' \emph{Experimental
  Mechanics}, vol.~22, no.~11, pp. 418--433, 1982.

\bibitem{grinberg1994geometry}
V.~S. Grinberg, G.~W. Podnar, and M.~Siegel, ``Geometry of binocular imaging,''
  in \emph{Stereoscopic Displays and Virtual Reality Systems}, vol. 2177, 1994,
  pp. 56--65.

\bibitem{Sun17OE}
Y.~Sun, X.~Yuan, and S.~Pang, ``Compressive high-speed stereo imaging,''
  \emph{Optics Express}, vol.~25, no.~15, pp. 18\,182--18\,190, 2017.

\bibitem{chen2012depth}
L.~Chen, H.~Lin, and S.~Li, ``Depth image enhancement for kinect using region
  growing and bilateral filter,'' in \emph{International Conference on Pattern
  Recognition}, 2012, pp. 3070--3073.

\bibitem{Sun16OE}
Y.~Sun, X.~Yuan, and S.~Pang, ``High-speed compressive range imaging based on
  active illumination,'' \emph{Optics Express}, vol.~24, no.~20, pp.
  22\,836--22\,846, 2016.

\bibitem{Willomitzer_OE17}
F.~Willomitzer and G.~H\"{a}usler, ``Single-shot {3D} motion picture camera
  with a dense point cloud,'' \emph{Optics Express}, vol.~25, no.~19, pp.
  23\,451--23\,464, Sep 2017.

\bibitem{wu2020freecam3d}
Y.Wu, V.Boominathan, J.~X.Zhao, H.Kawasaki, A.Sankaranarayanan, and
  A.Veeraraghavan, ``Freecam{3D}: Snapshot structured light {3D} with
  freely-moving cameras,'' in \emph{European Conference on Computer Vision},
  2020, pp. 309--325.

\bibitem{ng2005light}
R.~Ng, M.~Levoy, M.~Br{\'e}dif, G.~Duval, M.~Horowitz, and P.~Hanrahan, ``Light
  field photography with a hand-held plenoptic camera,'' \emph{Computer Science
  Technical Report}, vol.~2, no.~11, pp. 1--11, 2005.

\bibitem{chang2016variable}
X.~J.Chang, I.Kauvar and G.Wetzstein, ``Variable aperture light field
  photography: overcoming the diffraction-limited spatio-angular resolution
  tradeoff,'' in \emph{IEEE Conference on Computer Vision and Pattern
  Recognition}, 2016, pp. 3737--3745.

\bibitem{tambe2013towards}
A.~S.Tambe and A.Agrawal, ``Towards motion aware light field video for dynamic
  scenes,'' in \emph{IEEE International Conference on Computer Vision}, 2013,
  pp. 1009--1016.

\bibitem{lehtinen2011temporal}
J.Lehtinen, T.Aila, J.Chen, S.Laine, and F.Durand, ``Temporal light field
  reconstruction for rendering distribution effects,'' in \emph{ACM SIGGRAPH
  papers}, 2011, pp. 1--12.

\bibitem{jayasuriya2015depth}
S.Jayasuriya, A.Pediredla, S.Sivaramakrishnan, A.Molnar, and A.Veeraraghavan,
  ``Depth fields: Extending light field techniques to time-of-flight imaging,''
  in \emph{2015 International Conference on 3D Vision}, 2015, pp. 1--9.

\bibitem{nayar1994shape}
S.~K. Nayar and Y.~Nakagawa, ``Shape from focus,'' \emph{IEEE Transactions on
  Pattern Analysis and Machine Intelligence}, vol.~16, no.~8, pp. 824--831,
  1994.

\bibitem{el2019solving}
M.~M.El.Helou and S.Susstrunk, ``Solving the depth ambiguity in
  single-perspective images,'' \emph{OSA Continuum}, vol.~2, no.~10, pp.
  2901--2913, 2019.

\bibitem{el2019closed}
M.~M.El.Helou and S.Ssstrunk, ``Closed-form solution to disambiguate defocus
  blur in single-perspective images,'' in \emph{Mathematics in Imaging}, 2019,
  pp. MM1D--2.

\bibitem{xiong1993depth}
Y.~Xiong and S.~A. Shafer, ``Depth from focusing and defocusing,'' in
  \emph{IEEE Conference on Computer Vision and Pattern Recognition}, 1993, pp.
  68--73.

\bibitem{subbarao1994depth}
M.~Subbarao and G.~Surya, ``Depth from defocus: A spatial domain approach,''
  \emph{International Journal of Computer Vision}, vol.~13, no.~3, pp.
  271--294, 1994.

\bibitem{chaudhuri2012depth}
S.~Chaudhuri and A.~N. Rajagopalan, \emph{Depth from defocus: a real aperture
  imaging approach}.\hskip 1em plus 0.5em minus 0.4em\relax Springer Science \&
  Business Media, 2012.

\bibitem{Zhou2009_ICCV}
C.~Zhou, S.~Lin, and S.~Nayar, ``Coded aperture pairs for depth from defocus,''
  in \emph{IEEE International Conference on Computer Vision}, 2009, pp.
  325--332.

\bibitem{levin2007image}
A.~Levin, R.~Fergus, F.~Durand, and W.~T. Freeman, ``Image and depth from a
  conventional camera with a coded aperture,'' \emph{ACM transactions on
  graphics}, vol.~26, no.~3, pp. 70--es, 2007.

\bibitem{kazmi2014indoor}
W.~Kazmi, S.~Foix, G.~Aleny{\`a}, and H.~J. Andersen, ``Indoor and outdoor
  depth imaging of leaves with time-of-flight and stereo vision sensors:
  Analysis and comparison,'' \emph{ISPRS journal of photogrammetry and remote
  sensing}, vol.~88, pp. 128--146, 2014.

\bibitem{heide2015doppler}
M.~F.Heide, W.Heidrich and G.Wetzstein, ``Doppler time-of-flight imaging,''
  \emph{ACM Transactions on Graphics}, vol.~34, no.~4, pp. 1--11, 2015.

\bibitem{shrestha2016computational}
W.~S.Shrestha, F.Heide and G.Wetzstein, ``Computational imaging with
  multi-camera time-of-flight systems,'' \emph{ACM Transactions on Graphics},
  vol.~35, no.~4, pp. 1--11, 2016.

\bibitem{marco2017deeptof}
J.~Marco, Q.~Hernandez, A.~Munoz, Y.~Dong, A.~Jarabo, M.~H. Kim, X.~Tong, and
  D.~Gutierrez, ``Deeptof: off-the-shelf real-time correction of multipath
  interference in time-of-flight imaging,'' \emph{ACM Transactions on
  Graphics}, vol.~36, no.~6, pp. 1--12, 2017.

\bibitem{su2018deep}
S.~Su, F.~Heide, G.~Wetzstein, and W.~Heidrich, ``Deep end-to-end
  time-of-flight imaging,'' in \emph{IEEE Conference on Computer Vision and
  Pattern Recognition}, 2018, pp. 6383--6392.

\bibitem{gong2016three}
W.~Gong, C.~Zhao, H.~Yu, M.~Chen, W.~Xu, and S.~Han, ``Three-dimensional ghost
  imaging lidar via sparsity constraint,'' \emph{Scientific Reports}, vol.~6,
  no.~1, p. 26133, 2016.

\bibitem{kirmani2014first}
A.~Kirmani, D.~Venkatraman, D.~Shin, A.~Cola{\c{c}}o, F.~N. Wong, J.~H.
  Shapiro, and V.~K. Goyal, ``First-photon imaging,'' \emph{Science}, vol. 343,
  no. 6166, pp. 58--61, 2014.

\bibitem{o2017reconstructing}
M.O'Toole, F.Heide, D.B.Lindell, S.~K.Zang, and G.Wetzstein, ``Reconstructing
  transient images from single-photon sensors,'' in \emph{IEEE Conference on
  Computer Vision and Pattern Recognition}, 2017, pp. 1539--1547.

\bibitem{lindell2018towards}
G.~D.B.Lindell, M.O'Toole, ``Towards transient imaging at interactive rates
  with single-photon detectors,'' in \emph{IEEE International Conference on
  Computational Photography}, 2018, pp. 1--8.

\bibitem{heide2018sub}
D.~F.Heide, S.Diamond and G.Wetzstein, ``Sub-picosecond photon-efficient 3d
  imaging using single-photon sensors,'' \emph{Scientific reports}, vol.~8,
  no.~1, pp. 1--8, 2018.

\bibitem{wu2018learning}
J.Wu, C.Zhang, X.Zhang, Z.Zhang, W.T.Freeman, and J.B.Tenenbaum, ``Learning
  shape priors for single-view {3D} completion and reconstruction,'' in
  \emph{European Conference on Computer Vision}, 2018, pp. 646--662.

\bibitem{sun2020spadnet}
O.~Z.Sun, D.B.Lindell and G.Wetzstein, ``Spadnet: deep rgb-spad sensor fusion
  assisted by monocular depth estimation,'' \emph{Optics Express}, vol.~28,
  no.~10, pp. 14\,948--14\,962, 2020.

\bibitem{chang2019deep}
J.~Chang and G.~Wetzstein, ``{Deep optics for monocular depth estimation and 3D
  object detection},'' in \emph{IEEE International Conference on Computer
  Vision}, 2019, pp. 10\,193--10\,202.

\bibitem{wu2019phasecam3d}
Y.~Wu, V.~Boominathan, H.~Chen, A.~Sankaranarayanan, and A.~Veeraraghavan,
  ``Phasecam{3D}—learning phase masks for passive single view depth
  estimation,'' in \emph{IEEE International Conference on Computational
  Photography}, 2019, pp. 1--12.

\bibitem{ba2019deep}
Y.~Ba, A.~R. Gilbert, F.~Wang, J.~Yang, R.~Chen, Y.~Wang, L.~Yan, B.~Shi, and
  A.~Kadambi, ``Deep shape from polarization,'' \emph{arXiv preprint
  arXiv:1903.10210}, 2019.

\bibitem{zhang2020multiscale}
J.~Zhang, T.~Zhu, A.~Zhang, X.~Yuan, Z.~Wang, S.~Beetschen, L.~Xu, X.~Lin,
  Q.~Dai, and L.~Fang, ``Multiscale-vr: Multiscale gigapixel 3d panoramic
  videography for virtual reality,'' in \emph{IEEE International Conference on
  Computational Photography}, 2020, pp. 1--12.

\bibitem{zhao2012ghost}
C.~Zhao, W.~Gong, M.~Chen, E.~Li, H.~Wang, W.~Xu, and S.~Han, ``Ghost imaging
  lidar via sparsity constraints,'' \emph{Applied Physics Letters}, vol. 101,
  no.~14, p. 141123, 2012.

\bibitem{levoy1996light}
M.~Levoy and P.~Hanrahan, ``Light field rendering,'' in \emph{Annual Conference
  on Computer Graphics and Interactive Techniques}, 1996, pp. 31--42.

\bibitem{Llull15Optica}
P.~Llull, X.~Yuan, L.~Carin, and D.~Brady, ``Image translation for single-shot
  focal tomography,'' \emph{Optica}, vol.~2, no.~9, pp. 822--825, 2015.

\bibitem{kadambi2017rethinking}
A.Kadambi and R.Raskar, ``Rethinking machine vision time of flight with ghz
  heterodyning,'' \emph{IEEE Access}, vol.~5, pp. 26\,211--26\,223, 2017.

\bibitem{hitomi2011video}
Y.~Hitomi, J.~Gu, M.~Gupta, T.~Mitsunaga, and S.~K. Nayar, ``Video from a
  single coded exposure photograph using a learned over-complete dictionary,''
  in \emph{IEEE International Conference on Computer Vision}, 2011, pp.
  287--294.

\bibitem{reddy2011p2c2}
D.~Reddy, A.~Veeraraghavan, and R.~Chellappa, ``{P2C2}: Programmable pixel
  compressive camera for high speed imaging,'' in \emph{IEEE Conference on
  Computer Vision and Pattern Recognition}, 2011, pp. 329--336.

\bibitem{llull2013coded}
P.~Llull, X.~Liao, X.~Yuan, J.~Yang, D.~Kittle, L.~Carin, G.~Sapiro, and D.~J.
  Brady, ``Coded aperture compressive temporal imaging,'' \emph{Optics
  Express}, vol.~21, no.~9, pp. 10\,526--10\,545, 2013.

\bibitem{deng2019sinusoidal}
C.~Deng, Y.~Zhang, Y.~Mao, J.~Fan, J.~Suo, Z.~Zhang, and Q.~Dai, ``Sinusoidal
  sampling enhanced compressive camera for high speed imaging,'' \emph{IEEE
  Transactions on Pattern Analysis and Machine Intelligence}, 2019.

\bibitem{liu2013efficient}
D.~Liu, J.~Gu, Y.~Hitomi, M.~Gupta, T.~Mitsunaga, and S.~K. Nayar, ``Efficient
  space-time sampling with pixel-wise coded exposure for high-speed imaging,''
  \emph{IEEE Transactions on Pattern Analysis and Machine Intelligence},
  vol.~36, no.~2, pp. 248--260, 2013.

\bibitem{gao2014single}
L.~Gao, J.~Liang, C.~Li, and L.~V. Wang, ``Single-shot compressed ultrafast
  photography at one hundred billion frames per second,'' \emph{Nature}, vol.
  516, no. 7529, pp. 74--77, 2014.

\bibitem{qi2020single}
D.~Qi, S.~Zhang, C.~Yang, Y.~He, F.~Cao, J.~Yao, P.~Ding, L.~Gao, T.~Jia,
  J.~Liang \emph{et~al.}, ``Single-shot compressed ultrafast photography: a
  review,'' \emph{Advanced Photonics}, vol.~2, no.~1, p. 014003, 2020.

\bibitem{liang2020single}
J.~Liang, P.~Wang, L.~Zhu, and L.~V. Wang, ``Single-shot stereo-polarimetric
  compressed ultrafast photography for light-speed observation of
  high-dimensional optical transients with picosecond resolution,''
  \emph{Nature Communications}, vol.~11, no.~1, pp. 1--10, 2020.

\bibitem{velten2013femto}
A.~Velten, D.~Wu, A.~Jarabo, B.~Masia, C.~Barsi, C.~Joshi, E.~Lawson,
  M.~Bawendi, D.~Gutierrez, and R.~Raskar, ``Femto-photography: capturing and
  visualizing the propagation of light,'' \emph{ACM Transactions on Graphics},
  vol.~32, no.~4, pp. 1--8, 2013.

\bibitem{antipa2019video}
N.~Antipa, P.~Oare, E.~Bostan, R.~Ng, and L.~Waller, ``Video from stills:
  Lensless imaging with rolling shutter,'' in \emph{IEEE International
  Conference on Computational Photography}, 2019, pp. 1--8.

\bibitem{agrawal2010optimal}
A.~Agrawal, M.~Gupta, A.~Veeraraghavan, and S.~G. Narasimhan, ``Optimal coded
  sampling for temporal super-resolution,'' in \emph{IEEE Conference on
  Computer Vision and Pattern Recognition}, 2010, pp. 599--606.

\bibitem{Qiao2020_CACTI}
M.~Qiao, X.~Liu, and X.~Yuan, ``Snapshot spatial--temporal compressive
  imaging,'' \emph{Optical Letters}, vol.~45, no.~7, pp. 1659--1662, 2020.

\bibitem{Yuan14CVPR}
X.~Yuan, P.~Llull, X.~Liao, J.~Yang, D.~J. Brady, G.~Sapiro, and L.~Carin,
  ``Low-cost compressive sensing for color video and depth,'' in \emph{IEEE
  Conference on Computer Vision and Pattern Recognition}, 2014, pp. 3318--3325.

\bibitem{Yuan16BOE}
X.~Yuan and S.~Pang, ``Structured illumination temporal compressive
  microscopy,'' \emph{Biomedical Optics Express}, vol.~7, no.~3, pp. 746--758,
  2016.

\bibitem{donoho2006compressed}
D.~L. Donoho, ``Compressed sensing,'' \emph{IEEE Transactions on Information
  Theory}, vol.~52, no.~4, pp. 1289--1306, 2006.

\bibitem{Candes06TIT}
E.~J. {Candes}, J.~{Romberg}, and T.~{Tao}, ``Robust uncertainty principles:
  Exact signal reconstruction from highly incomplete frequency information,''
  \emph{IEEE Transactions on Information Theory}, vol.~52, no.~2, pp. 489--509,
  2006.

\bibitem{zhao2019hyperspectral}
Y.~Zhao, H.~Guo, Z.~Ma, X.~Cao, T.~Yue, and X.~Hu, ``{Hyperspectral imaging
  with random printed mask},'' in \emph{IEEE Conference on Computer Vision and
  Pattern Recognition}, 2019, pp. 10\,149--10\,157.

\bibitem{cao2011prism}
X.~Cao, H.~Du, X.~Tong, Q.~Dai, and S.~Lin, ``A prism-mask system for
  multispectral video acquisition,'' \emph{IEEE Transactions on Pattern
  Analysis and Machine Intelligence}, vol.~33, no.~12, pp. 2423--2435, 2011.

\bibitem{ma2014acquisition}
C.~Ma, X.~Cao, X.~Tong, Q.~Dai, and S.~Lin, ``Acquisition of high spatial and
  spectral resolution video with a hybrid camera system,'' \emph{{International
  Journal of Computer Vision}}, vol. 110, no.~2, pp. 141--155, 2014.

\bibitem{cao2016computational}
X.~Cao, T.~Yue, X.~Lin, S.~Lin, X.~Yuan, Q.~Dai, L.~Carin, and D.~J. Brady,
  ``Computational snapshot multispectral cameras: Toward dynamic capture of the
  spectral world,'' \emph{IEEE Signal Processing Magazine}, vol.~33, no.~5, pp.
  95--108, 2016.

\bibitem{arad2018ntire}
O.-S. B.Arad and R.Timofte, ``Ntire 2018 challenge on spectral reconstruction
  from rgb images,'' in \emph{IEEE Conference on Computer Vision and Pattern
  Recognition Workshops}, 2018, pp. 929--938.

\bibitem{goodfellow2020generative}
I.~Goodfellow, J.~Pouget-Abadie, M.~Mirza, B.~Xu, D.~Warde-Farley, S.~Ozair,
  A.~Courville, and Y.~Bengio, ``Generative adversarial networks,''
  \emph{Communications of the ACM}, vol.~63, no.~11, pp. 139--144, 2020.

\bibitem{wagadarikar2008single}
A.~Wagadarikar, R.~John, R.~Willett, and D.~Brady, ``Single disperser design
  for coded aperture snapshot spectral imaging,'' \emph{Applied Optics},
  vol.~47, no.~10, pp. B44--B51, 2008.

\bibitem{mohan2008agile}
A.~Mohan, R.~Raskar, and J.~Tumblin, ``Agile spectrum imaging: Programmable
  wavelength modulation for cameras and projectors,'' in \emph{Computer
  Graphics Forum}, vol.~27, no.~2, 2008, pp. 709--717.

\bibitem{lin2014dual}
X.~Lin, G.~Wetzstein, Y.~Liu, and Q.~Dai, ``Dual-coded compressive
  hyperspectral imaging,'' \emph{Optics Letters}, vol.~39, no.~7, pp.
  2044--2047, 2014.

\bibitem{Yuan15JSTSP}
X.~Yuan, T.-H. Tsai, R.~Zhu, P.~Llull, D.~J. Brady, and L.~Carin,
  ``{Compressive hyperspectral imaging with side information},'' \emph{IEEE
  Journal of Selected Topics in Signal Processing}, vol.~9, no.~6, pp.
  964--976, 2015.

\bibitem{Meng2020_OL_SHEM}
Z.~Meng, M.~Qiao, J.~Ma, Z.~Yu, K.~Xu, and X.~Yuan, ``Snapshot multispectral
  endomicroscopy,'' \emph{Optical Letters}, vol.~45, no.~14, pp. 3897--3900,
  2020.

\bibitem{lohmann1996space}
A.~W. Lohmann, R.~G. Dorsch, D.~Mendlovic, Z.~Zalevsky, and C.~Ferreira,
  ``Space--bandwidth product of optical signals and systems,'' \emph{The
  Journal of the Optical Society of America A}, vol.~13, no.~3, pp. 470--473,
  1996.

\bibitem{cossairt2011gigapixel}
O.~S. Cossairt, D.~Miau, and S.~K. Nayar, ``Gigapixel computational imaging,''
  in \emph{IEEE International Conference on Computational Photography}, 2011,
  pp. 1--8.

\bibitem{brady2012multiscale}
D.~J. Brady, M.~E. Gehm, R.~A. Stack, D.~L. Marks, D.~S. Kittle, D.~R. Golish,
  E.~Vera, and S.~D. Feller, ``Multiscale gigapixel photography,''
  \emph{Nature}, vol. 486, no. 7403, pp. 386--389, 2012.

\bibitem{yuan2017multiscale}
X.~Yuan, L.~Fang, Q.~Dai, D.~J. Brady, and Y.~Liu, ``Multiscale gigapixel
  video: A cross resolution image matching and warping approach,'' in
  \emph{IEEE International Conference on Computational Photography}, 2017, pp.
  1--9.

\bibitem{nayar2000high}
S.~K. Nayar and T.~Mitsunaga, ``High dynamic range imaging: Spatially varying
  pixel exposures,'' in \emph{IEEE Conference on Computer Vision and Pattern
  Recognition}, vol.~1, 2000, pp. 472--479.

\bibitem{Metzler_2020_CVPR}
C.~A. Metzler, H.~Ikoma, Y.~Peng, and G.~Wetzstein, ``{Deep optics for
  single-shot high-dynamic-range imaging},'' in \emph{IEEE Conference on
  Computer Vision and Pattern Recognition}, 2020.

\bibitem{sun2020learning}
Q.Sun, E.Tseng, Q.Fu, W.Heidrich, and F.Heide, ``Learning rank-1 diffractive
  optics for single-shot high dynamic range imaging,'' in \emph{IEEE Conference
  on Computer Vision and Pattern Recognition}, 2020, pp. 1386--1396.

\bibitem{reinhard2010high}
E.~Reinhard, W.~Heidrich, P.~Debevec, S.~Pattanaik, G.~Ward, and K.~Myszkowski,
  \emph{High dynamic range imaging: Acquisition, display, and image-based
  lighting}.\hskip 1em plus 0.5em minus 0.4em\relax Morgan Kaufmann, 2010.

\bibitem{cvetkovic2010automatic}
S.~Cvetkovic, H.~Jellema, and P.~H. de~With, ``Automatic level control for
  video cameras towards hdr techniques,'' \emph{EURASIP Journal on Image and
  Video Processing}, vol. 2010, pp. 1--30, 2010.

\bibitem{Nayar03_HDR}
S.~{Nayar} and N.~{Branzoi}, ``Adaptive dynamic range imaging: Optical control
  of pixel exposures over space and time,'' in \emph{IEEE International
  Conference on Computer Vision}, vol.~2, 2003, pp. 1168--1175.

\bibitem{banterle2017advanced}
F.~Banterle, A.~Artusi, K.~Debattista, and A.~Chalmers, \emph{Advanced high
  dynamic range imaging}.\hskip 1em plus 0.5em minus 0.4em\relax CRC Press,
  2017.

\bibitem{rahman2011using}
M.~T. Rahman, N.~Kehtarnavaz, and Q.~R. Razlighi, ``Using image entropy maximum
  for auto exposure,'' \emph{Journal of Electronic Imaging}, vol.~20, no.~1, p.
  013007, 2011.

\bibitem{barakat2008minimal}
N.~Barakat, A.~N. Hone, and T.~E. Darcie, ``Minimal-bracketing sets for
  high-dynamic-range image capture,'' \emph{IEEE Transactions on Image
  Processing}, vol.~17, no.~10, pp. 1864--1875, 2008.

\bibitem{martel2020neural}
J.~N. Martel, L.~Mueller, S.~J. Carey, P.~Dudek, and G.~Wetzstein, ``{Neural
  sensors: Learning pixel exposures for HDR imaging and video compressive
  sensing with programmable sensors},'' \emph{IEEE Transactions on Pattern
  Analysis and Machine Intelligence}, vol.~42, no.~7, pp. 1642--1653, 2020.

\bibitem{zhao2015unbounded}
H.~Zhao, B.~Shi, C.~Fernandez-Cull, S.-K. Yeung, and R.~Raskar, ``Unbounded
  high dynamic range photography using a modulo camera,'' in \emph{IEEE
  International Conference on Computational Photography}, 2015.

\bibitem{faccio2020non}
D.~Faccio, A.~Velten, and G.~Wetzstein, ``Non-line-of-sight imaging,''
  \emph{Nature Reviews Physics}, vol.~2, no.~6, pp. 318--327, 2020.

\bibitem{velten2012recovering}
A.~Velten, T.~Willwacher, O.~Gupta, A.~Veeraraghavan, M.~G. Bawendi, and
  R.~Raskar, ``Recovering three-dimensional shape around a corner using
  ultrafast time-of-flight imaging,'' \emph{Nature Communications}, vol.~3,
  no.~1, pp. 1--8, 2012.

\bibitem{o2018confocal}
D.~M.O’Toole and G.Wetzstein, ``Confocal non-line-of-sight imaging based on
  the light-cone transform,'' \emph{Nature}, vol. 555, no. 7696, pp. 338--341,
  2018.

\bibitem{lindell2019wave}
G.~D.B.Lindell and M.O'Toole, ``Wave-based non-line-of-sight imaging using fast
  fk migration,'' \emph{ACM Transactions on Graphics}, vol.~38, no.~4, pp.
  1--13, 2019.

\bibitem{ahn2019convolutional}
B.Ahn, A.Dave, A.Veeraraghavan, I.Gkioulekas, and A.Sankaranarayanan,
  ``Convolutional approximations to the general non-line-of-sight imaging
  operator,'' in \emph{IEEE International Conference on Computer Vision}, 2019,
  pp. 7889--7899.

\bibitem{musarra2019non}
G.~Musarra, A.~Lyons, E.~Conca, Y.~Altmann, F.~Villa, F.~Zappa, M.~J. Padgett,
  and D.~Faccio, ``Non-line-of-sight three-dimensional imaging with a
  single-pixel camera,'' \emph{Physical Review Applied}, vol.~12, no.~1, p.
  011002, 2019.

\bibitem{musarra20193d}
G.~Musarra, A.~Lyons, E.~Conca, F.~Villa, F.~Zappa, Y.~Altmann, and D.~Faccio,
  ``3d rgb non-line-of-sight single-pixel imaging,'' in \emph{Imaging Systems
  and Applications}, 2019, pp. IM2B--5.

\bibitem{la2018error}
M.~La~Manna, F.~Kine, E.~Breitbach, J.~Jackson, T.~Sultan, and A.~Velten,
  ``Error backprojection algorithms for non-line-of-sight imaging,'' \emph{IEEE
  transactions on pattern analysis and machine intelligence}, vol.~41, no.~7,
  pp. 1615--1626, 2018.

\bibitem{heide2019non}
F.Heide, M.O’Toole, K.Zang, D.B.Lindell, S.Diamond, and G.Wetzstein,
  ``Non-line-of-sight imaging with partial occluders and surface normals,''
  \emph{ACM Transactions on Graphics}, vol.~38, no.~3, pp. 1--10, 2019.

\bibitem{tsai2019beyond}
C.-Y. Tsai, A.~C. Sankaranarayanan, and I.~Gkioulekas, ``Beyond volumetric
  albedo--a surface optimization framework for non-line-of-sight imaging,'' in
  \emph{IEEE Conference on Computer Vision and Pattern Recognition}, 2019, pp.
  1545--1555.

\bibitem{thrampoulidis2018exploiting}
C.~Thrampoulidis, G.~Shulkind, F.~Xu, W.~T. Freeman, J.~H. Shapiro,
  A.~Torralba, F.~N. Wong, and G.~W. Wornell, ``Exploiting occlusion in
  non-line-of-sight active imaging,'' \emph{IEEE Transactions on Computational
  Imaging}, vol.~4, no.~3, pp. 419--431, 2018.

\bibitem{rapp2020seeing}
J.~Rapp, C.~Saunders, J.~Tachella, J.~Murray-Bruce, Y.~Altmann, J.-Y.
  Tourneret, S.~McLaughlin, R.~M. Dawson, F.~N. Wong, and V.~K. Goyal, ``Seeing
  around corners with edge-resolved transient imaging,'' \emph{Nature
  Communications}, vol.~11, no.~1, pp. 1--10, 2020.

\bibitem{tanaka2020polarized}
K.~Tanaka, Y.~Mukaigawa, and A.~Kadambi, ``Polarized non-line-of-sight
  imaging,'' in \emph{IEEE Conference on Computer Vision and Pattern
  Recognition}, 2020, pp. 2136--2145.

\bibitem{ye2021compressed}
J.-T. Ye, X.~Huang, Z.-P. Li, and F.~Xu, ``Compressed sensing for active
  non-line-of-sight imaging,'' \emph{Optics Express}, vol.~29, no.~2, pp.
  1749--1763, 2021.

\bibitem{wu2021non}
C.~Wu, J.~Liu, X.~Huang, Z.-P. Li, C.~Yu, J.-T. Ye, J.~Zhang, Q.~Zhang, X.~Dou,
  V.~K. Goyal \emph{et~al.}, ``Non--line-of-sight imaging over 1.43 km,''
  \emph{National Academy of Sciences}, vol. 118, no.~10, 2021.

\bibitem{maeda2019thermal}
T.~Maeda, Y.~Wang, R.~Raskar, and A.~Kadambi, ``Thermal non-line-of-sight
  imaging,'' in \emph{IEEE International Conference on Computational
  Photography}, 2019, pp. 1--11.

\bibitem{lindell2019acoustic}
G.~D.B.Lindell and V.Koltun, ``Acoustic non-line-of-sight imaging,'' in
  \emph{IEEE Conference on Computer Vision and Pattern Recognition}, 2019, pp.
  6780--6789.

\bibitem{satat2017object}
G.~Satat, M.~Tancik, O.~Gupta, B.~Heshmat, and R.~Raskar, ``Object
  classification through scattering media with deep learning on time resolved
  measurement,'' \emph{Optics Express}, vol.~25, no.~15, pp. 17\,466--17\,479,
  2017.

\bibitem{isogawa2020optical}
M.~Isogawa, Y.~Yuan, M.~O'Toole, and K.~M. Kitani, ``Optical non-line-of-sight
  physics-based 3d human pose estimation,'' in \emph{IEEE Conference on
  Computer Vision and Pattern Recognition}, 2020, pp. 7013--7022.

\bibitem{metzler2019keyhole}
D.~C.A.Metzler and G.Wetzstein, ``Keyhole imaging: Non-line-of-sight imaging
  and tracking of moving objects along a single optical path at long standoff
  distances,'' \emph{arXiv e-prints}, pp. arXiv--1912, 2019.

\bibitem{tancik2018nlos}
M.~Tancik, T.~Swedish, G.~Satat, and R.~Raskar, ``Data-driven non-line-of-sight
  imaging with a traditional camera,'' in \emph{Imaging Systems and
  Applications}, 2018, pp. IW2B--6.

\bibitem{pandharkar2011estimating}
R.~Pandharkar, A.~Velten, A.~Bardagjy, E.~Lawson, M.~Bawendi, and R.~Raskar,
  ``Estimating motion and size of moving non-line-of-sight objects in cluttered
  environments,'' in \emph{IEEE Conference on Computer Vision and Pattern
  Recognition}, 2011, pp. 265--272.

\bibitem{katz2012looking}
O.~Katz, E.~Small, and Y.~Silberberg, ``Looking around corners and through thin
  turbid layers in real time with scattered incoherent light,'' \emph{Nature
  Photonics}, vol.~6, no.~8, pp. 549--553, 2012.

\bibitem{liu2019non}
X.~Liu, I.~Guill{\'e}n, M.~La~Manna, J.~H. Nam, S.~A. Reza, T.~H. Le,
  A.~Jarabo, D.~Gutierrez, and A.~Velten, ``Non-line-of-sight imaging using
  phasor-field virtual wave optics,'' \emph{Nature}, vol. 572, no. 7771, pp.
  620--623, 2019.

\bibitem{raskar2020seeing}
R.~Raskar, A.~Velten, S.~Bauer, and T.~Swedish, ``Seeing around corners using
  time of flight,'' in \emph{ACM SIGGRAPH Courses}, 2020, pp. 1--97.

\bibitem{bedri2014seeing}
H.~Bedri, M.~Feigin, M.~Everett, I.~Filho, G.~L. Charvat, and R.~Raskar,
  ``Seeing around corners with a mobile phone? synthetic aperture audio
  imaging,'' in \emph{ACM SIGGRAPH Posters}, 2014, pp. 1--1.

\bibitem{mao2018aim}
W.~Mao, M.~Wang, and L.~Qiu, ``Aim: Acoustic imaging on a mobile,'' in
  \emph{Annual International Conference on Mobile Systems, Applications, and
  Services}, 2018, pp. 468--481.

\bibitem{su2015acoustic}
F.~Su and C.~Joslin, ``Acoustic imaging using a 64-node microphone array and
  beamformer system,'' in \emph{IEEE International Symposium on Signal
  Processing and Information Technology}, 2015, pp. 168--173.

\bibitem{raskar2014looking}
R.~Raskar, ``Looking around corners with trillion frames per second imaging and
  projects in eye-diagnostics on a mobile phone,'' in \emph{Laser Applications
  to Chemical, Security and Environmental Analysis}, 2014, pp. JTu1A--2.

\bibitem{katz2014non}
O.~Katz, P.~Heidmann, M.~Fink, and S.~Gigan, ``Non-invasive single-shot imaging
  through scattering layers and around corners via speckle correlations,''
  \emph{Nature Photonics}, vol.~8, no.~10, pp. 784--790, 2014.

\bibitem{kirmani2011looking}
A.~Kirmani, T.~Hutchison, J.~Davis, and R.~Raskar, ``Looking around the corner
  using ultrafast transient imaging,'' \emph{International journal of Computer
  Vision}, vol.~95, no.~1, pp. 13--28, 2011.

\bibitem{buttafava2015non}
M.~Buttafava, J.~Zeman, A.~Tosi, K.~Eliceiri, and A.~Velten,
  ``Non-line-of-sight imaging using a time-gated single photon avalanche
  diode,'' \emph{Optics Express}, vol.~23, no.~16, pp. 20\,997--21\,011, 2015.

\bibitem{xin2019theory}
S.~Xin, S.~Nousias, K.~N. Kutulakos, A.~C. Sankaranarayanan, S.~G. Narasimhan,
  and I.~Gkioulekas, ``A theory of fermat paths for non-line-of-sight shape
  reconstruction,'' in \emph{IEEE Conference on Computer Vision and Pattern
  Recognition}, 2019, pp. 6800--6809.

\bibitem{reza2019phasor}
S.~A. Reza, M.~La~Manna, S.~Bauer, and A.~Velten, ``Phasor field waves: A
  huygens-like light transport model for non-line-of-sight imaging
  applications,'' \emph{Optics Express}, vol.~27, no.~20, pp. 29\,380--29\,400,
  2019.

\bibitem{reza2018physical}
S.~A. Reza, M.~La~Manna, and A.~Velten, ``A physical light transport model for
  non-line-of-sight imaging applications,'' \emph{arXiv preprint
  arXiv:1802.01823}, 2018.

\bibitem{lei2019direct}
X.~Lei, L.~He, Y.~Tan, K.~X. Wang, X.~Wang, Y.~Du, S.~Fan, and Z.~Yu, ``Direct
  object recognition without line-of-sight using optical coherence,'' in
  \emph{IEEE Conference on Computer Vision and Pattern Recognition}, 2019, pp.
  11\,737--11\,746.

\bibitem{reza2018imaging}
S.~A. Reza, M.~La~Manna, and A.~Velten, ``Imaging with phasor fields for
  non-line-of sight applications,'' in \emph{Computational Optical Sensing and
  Imaging}, 2018, pp. CM2E--7.

\bibitem{kiran2018non}
S.~kiran Doddalla and G.~C. Trichopoulos, ``Non-line of sight terahertz imaging
  from a single viewpoint,'' in \emph{IEEE/MTT-S International Microwave
  Symposium-IMS}, 2018, pp. 1527--1529.

\bibitem{la2020non}
M.~La~Manna, J.-H. Nam, S.~A. Reza, and A.~Velten, ``Non-line-of-sight-imaging
  using dynamic relay surfaces,'' \emph{Optics Express}, vol.~28, no.~4, pp.
  5331--5339, 2020.

\bibitem{liu2020role}
X.~Liu and A.~Velten, ``The role of wigner distribution function in
  non-line-of-sight imaging,'' in \emph{IEEE International Conference on
  Computational Photography}, 2020, pp. 1--12.

\bibitem{guillen2020effect}
I.~Guill{\'e}n, X.~Liu, A.~Velten, D.~Gutierrez, and A.~Jarabo, ``On the effect
  of reflectance on phasor field non-line-of-sight imaging,'' in \emph{ICASSP
  IEEE International Conference on Acoustics, Speech and Signal Processing},
  2020, pp. 9269--9273.

\bibitem{kirmani2009looking}
A.~Kirmani, T.~Hutchison, J.~Davis, and R.~Raskar, ``Looking around the corner
  using transient imaging,'' in \emph{IEEE International Conference on Computer
  Vision}, 2009, pp. 159--166.

\bibitem{lindell2020computational}
D.~B. Lindell, M.~O'Toole, S.~Narasimhan, and R.~Raskar, ``Computational
  time-resolved imaging, single-photon sensing, and non-line-of-sight
  imaging,'' in \emph{ACM SIGGRAPH Courses}, 2020, pp. 1--119.

\bibitem{liu2020phasor}
X.~Liu, S.~Bauer, and A.~Velten, ``Phasor field diffraction based
  reconstruction for fast non-line-of-sight imaging systems,'' \emph{Nature
  Communications}, vol.~11, no.~1, pp. 1--13, 2020.

\bibitem{tancik2018data}
M.~Tancik, T.~Swedish, G.~Satat, and R.~Raskar, ``Data-driven non-line-of-sight
  imaging with a traditional camera,'' in \emph{Imaging Systems and
  Applications}, 2018, pp. IW2B--6.

\bibitem{pandharkar2011hidden}
R.~R.~P. Pandharkar, ``Hidden object doppler: estimating motion, size and
  material properties of moving non-line-of-sight objects in cluttered
  environments,'' Ph.D. dissertation, Massachusetts Institute of Technology,
  2011.

\bibitem{seidel2020two}
S.~W. Seidel, J.~Murray-Bruce, Y.~Ma, C.~Yu, W.~T. Freeman, and V.~K. Goyal,
  ``Two-dimensional non-line-of-sight scene estimation from a single edge
  occluder,'' \emph{IEEE Transactions on Computational Imaging}, vol.~7, pp.
  58--72, 2020.

\bibitem{geng2021recent}
R.~Geng, Y.~Hu, and Y.~Chen, ``Recent advances on non-line-of-sight imaging:
  Conventional physical models, deep learning, and new scenes,'' \emph{arXiv
  preprint arXiv:2104.13807}, 2021.

\bibitem{moses2004wide}
R.~L. Moses, L.~C. Potter, and M.~Cetin, ``Wide-angle {SAR} imaging,'' in
  \emph{Algorithms for Synthetic Aperture Radar Imagery XI}, vol. 5427, 2004,
  pp. 164--175.

\bibitem{sun2020mimo}
S.~Sun, A.~P. Petropulu, and H.~V. Poor, ``Mimo radar for advanced
  driver-assistance systems and autonomous driving: Advantages and
  challenges,'' \emph{IEEE Signal Processing Magazine}, vol.~37, no.~4, pp.
  98--117, 2020.

\bibitem{mansour2018sparse}
U.~K. H.Mansour, D.Liu and P.T.Boufounos, ``Sparse blind deconvolution for
  distributed radar autofocus imaging,'' \emph{IEEE Transactions on
  Computational Imaging}, vol.~4, no.~4, pp. 537--551, 2018.

\bibitem{sun20214d}
S.~Sun and Y.~D. Zhang, ``4d automotive radar sensing for autonomous vehicles:
  A sparsity-oriented approach,'' \emph{IEEE Journal of Selected Topics in
  Signal Processing}, vol.~15, no.~4, pp. 879--891, 2021.

\bibitem{amin2014through}
M.~G. Amin and F.~Ahmad, ``Through-the-wall radar imaging: theory and
  applications,'' in \emph{Academic Press Library in Signal Processing}.\hskip
  1em plus 0.5em minus 0.4em\relax Elsevier, 2014, vol.~2, pp. 857--909.

\bibitem{amin2016radar}
M.~G. Amin, Y.~D. Zhang, F.~Ahmad, and K.~D. Ho, ``Radar signal processing for
  elderly fall detection: The future for in-home monitoring,'' \emph{IEEE
  Signal Processing Magazine}, vol.~33, no.~2, pp. 71--80, 2016.

\bibitem{wu2014compressive}
Q.~Wu, Y.~D. Zhang, F.~Ahmad, and M.~G. Amin, ``Compressive-sensing-based
  high-resolution polarimetric through-the-wall radar imaging exploiting target
  characteristics,'' \emph{IEEE Antennas and Wireless Propagation Letters},
  vol.~14, pp. 1043--1047, 2014.

\bibitem{wu2014through}
Q.~Wu, Y.~D. Zhang, M.~G. Amin, and F.~Ahmad, ``Through-the-wall radar imaging
  based on modified bayesian compressive sensing,'' in \emph{IEEE China Summit
  \& International Conference on Signal and Information Processing}, 2014, pp.
  232--236.

\bibitem{wang2019human}
M.~Wang, Y.~D. Zhang, and G.~Cui, ``Human motion recognition exploiting radar
  with stacked recurrent neural network,'' \emph{Digital Signal Processing},
  vol.~87, pp. 125--131, 2019.

\bibitem{zhang2018latern}
Z.~Zhang, Z.~Tian, and M.~Zhou, ``Latern: Dynamic continuous hand gesture
  recognition using fmcw radar sensor,'' \emph{IEEE Sensors Journal}, vol.~18,
  no.~8, pp. 3278--3289, 2018.

\bibitem{zhang2019udeep}
Z.~Zhang, Z.~Tian, Y.~Zhang, M.~Zhou, and B.~Wang, ``u-deephand: Fmcw
  radar-based unsupervised hand gesture feature learning using deep
  convolutional auto-encoder network,'' \emph{IEEE Sensors Journal}, vol.~19,
  no.~16, pp. 6811--6821, 2019.

\bibitem{hazra2018robust}
S.~Hazra and A.~Santra, ``Robust gesture recognition using millimetric-wave
  radar system,'' \emph{IEEE Sensors Letters}, vol.~2, no.~4, pp. 1--4, 2018.

\bibitem{kotaru2015spot}
M.~Kotaru, K.~Joshi, D.~Bharadia, and S.~Katti, ``Spotfi: Decimeter level
  localization using wifi,'' in \emph{ACM Conference on Special Interest Group
  on Data Communication}, 2015, pp. 269--282.

\bibitem{wang2017biloc}
X.~Wang, L.~Gao, and S.~Mao, ``Biloc: Bi-modal deep learning for indoor
  localization with commodity 5ghz wifi,'' \emph{IEEE Access}, vol.~5, pp.
  4209--4220, 2017.

\bibitem{wang2017csi}
X.~Wang, L.~Gao, S.~Mao, and S.~Pandey, ``Csi-based fingerprinting for indoor
  localization: A deep learning approach,'' \emph{IEEE Transactions on
  Vehicular Technology}, vol.~66, no.~1, pp. 763--776, 2017.

\bibitem{tomiyasu1978tutorial}
K.~Tomiyasu, ``Tutorial review of synthetic-aperture radar ({SAR}) with
  applications to imaging of the ocean surface,'' \emph{IEEE}, vol.~66, no.~5,
  pp. 563--583, 1978.

\bibitem{willett2013sparsity}
R.~M. Willett, M.~F. Duarte, M.~A. Davenport, and R.~G. Baraniuk, ``Sparsity
  and structure in hyperspectral imaging: Sensing, reconstruction, and target
  detection,'' \emph{IEEE Signal Processing Magazine}, vol.~31, no.~1, pp.
  116--126, 2013.

\bibitem{holloway2017savi}
J.~Holloway, Y.~Wu, M.~K. Sharma, O.~Cossairt, and A.~Veeraraghavan, ``{SAVI}:
  Synthetic apertures for long-range, subdiffraction-limited visible imaging
  using {F}ourier ptychography,'' \emph{Science Advances}, vol.~3, no.~4, p.
  e1602564, 2017.

\bibitem{abbas2019wideep}
M.~Abbas, M.~Elhamshary, H.~Rizk, M.~Torki, and M.~Youssef, ``Wideep:
  Wifi-based accurate and robust indoor localization system using deep
  learning,'' in \emph{IEEE International Conference on Pervasive Computing and
  Communications}, 2019, pp. 1--10.

\bibitem{mitra2014performance}
O.~K.Mitra and A.Veeraraghavan, ``Performance limits for computational
  photography,'' in \emph{Fringe 2013}.\hskip 1em plus 0.5em minus 0.4em\relax
  Springer, 2014, pp. 663--670.

\bibitem{Yuan_2021_Innovation}
X.~Yuan and S.~Han, ``Single-pixel neutron imaging with artificial
  intelligence: breaking the barrier in multi-parameter imaging, sensitivity
  and spatial resolution,'' \emph{The Innovation}, p. 100100, 2021.

\bibitem{rudin1992nonlinear}
L.~I. Rudin, S.~Osher, and E.~Fatemi, ``Nonlinear total variation based noise
  removal algorithms,'' \emph{Physica D: Nonlinear Phenomena}, vol.~60, no.
  1-4, pp. 259--268, 1992.

\bibitem{Liao14GAP}
X.~Liao, H.~Li, and L.~Carin, ``Generalized alternating projection for
  weighted-$\ell_{2,1}$ minimization with applications to model-based
  compressive sensing,'' \emph{SIAM Journal on Imaging Sciences}, vol.~7,
  no.~2, pp. 797--823, 2014.

\bibitem{Yuan16ICIP_GAP}
X.~Yuan, ``Generalized alternating projection based total variation
  minimization for compressive sensing,'' in \emph{International Conference on
  Image Processing}, 2016, pp. 2539--2543.

\bibitem{Bioucas-Dias2007TwIST}
J.~M. Bioucas-Dias and M.~A. Figueiredo, ``A new {TwIST: T}wo-step iterative
  shrinkage/thresholding algorithms for image restoration,'' \emph{IEEE
  Transactions on Image Processing}, vol.~16, no.~12, pp. 2992--3004, 2007.

\bibitem{Yang14GMM}
J.~Yang, X.~Yuan, X.~Liao, P.~Llull, G.~Sapiro, D.~J. Brady, and L.~Carin,
  ``{Video compressive sensing using gaussian mixture models},'' \emph{IEEE
  Transaction on Image Processing}, vol.~23, no.~11, pp. 4863--4878, 2014.

\bibitem{Yang14GMMonline}
J.~Yang, X.~Liao, X.~Yuan, P.~Llull, D.~J. Brady, G.~Sapiro, and L.~Carin,
  ``{Compressive sensing by learning a gaussian mixture model from
  measurements},'' \emph{IEEE Transaction on Image Processing}, vol.~24, no.~1,
  pp. 106--119, 2015.

\bibitem{Liu18TPAMI}
Y.~Liu, X.~Yuan, J.~Suo, D.~Brady, and Q.~Dai, ``{Rank minimization for
  snapshot compressive imaging},'' \emph{IEEE Transactions on Pattern Analysis
  and Machine Intelligence}, vol.~41, no.~12, pp. 2990--3006, 2019.

\bibitem{Yuan2021_SPM}
X.~{Yuan}, D.~J. {Brady}, and A.~K. {Katsaggelos}, ``Snapshot compressive
  imaging: Theory, algorithms, and applications,'' \emph{IEEE Signal Processing
  Magazine}, vol.~38, no.~2, pp. 65--88, 2021.

\bibitem{ma2019deep}
J.~Ma, X.-Y. Liu, Z.~Shou, and X.~Yuan, ``Deep tensor {ADMM-Net} for snapshot
  compressive imaging,'' in \emph{International Conference on Computer Vision},
  2019, pp. 10\,223--10\,232.

\bibitem{qiao2020deep}
M.~Qiao, Z.~Meng, J.~Ma, and X.~Yuan, ``Deep learning for video compressive
  sensing,'' \emph{APL Photonics}, vol.~5, no.~3, p. 030801, 2020.

\bibitem{Yuan_2020_CVPR}
X.~Yuan, Y.~Liu, J.~Suo, and Q.~Dai, ``Plug-and-play algorithms for large-scale
  snapshot compressive imaging,'' in \emph{IEEE Conference on Computer Vision
  and Pattern Recognition}, 2020, pp. 1444--1454.

\bibitem{Yuan2021_TPAMI_PnP}
X.~Yuan, Y.~Liu, J.~Suo, F.~Durand, and Q.~Dai, ``Plug-and-play algorithms for
  video snapshot compressive imaging,'' \emph{IEEE Transactions on Pattern
  Analysis and Machine Intelligence}, pp. 1--1, 2021.

\bibitem{kamilov2017plug}
H.~U.S~Kamilov and B.Wohlberg, ``A plug-and-play priors approach for solving
  nonlinear imaging inverse problems,'' \emph{IEEE Signal Processing Letters},
  vol.~24, no.~12, pp. 1872--1876, 2017.

\bibitem{zheng2021deep}
Z.~Siming, L.~Yang, M.~Ziyi, Q.~Mu, T.~Zhishen, Y.~Xiaoyu, H.~Shensheng, and
  Y.~Xin, ``Deep plug-and-play priors for spectral snapshot compressive
  imaging,'' \emph{Photonics Research}, vol.~9, no.~2, pp. B18--B29, 2021.

\bibitem{li2020end}
Y.Li, M.Qi, R.Gulve, M.Wei, R.Genov, K.N.Kutulakos, and W.Heidrich,
  ``End-to-end video compressive sensing using anderson-accelerated unrolled
  networks,'' in \emph{IEEE International Conference on Computational
  Photography}, 2020, pp. 1--12.

\bibitem{Cheng20ECCV_Birnat}
Z.~Cheng, R.~Lu, Z.~Wang, H.~Zhang, B.~Chen, Z.~Meng, and X.~Yuan, ``Birnat:
  Bidirectional recurrent neural networks with adversarial training for video
  snapshot compressive imaging,'' in \emph{European Conference on Computer
  Vision}, 2020.

\bibitem{Zheng2021_Patterns}
S.~Zheng, C.~Wang, X.~Yuan, and H.~L. Xin, ``Super-compression of large
  electron microscopy time series by deep compressive sensing learning,''
  \emph{Patterns}, vol.~2, no.~7, p. 100292, 2021.

\bibitem{Cheng2021_CVPR_ReverSCI}
Z.~Cheng, B.~Chen, G.~Liu, H.~Zhang, R.~Lu, Z.~Wang, and X.~Yuan,
  ``Memory-efficient network for large-scale video compressive sensing,'' in
  \emph{IEEE Conference on Computer Vision and Pattern Recognition}, June 2021,
  pp. 16\,246--16\,255.

\bibitem{Wang2021_CVPR_MetaSCI}
Z.~Wang, H.~Zhang, Z.~Cheng, B.~Chen, and X.~Yuan, ``Metasci: Scalable and
  adaptive reconstruction for video compressive sensing,'' in \emph{IEEE
  Conference on Computer Vision and Pattern Recognition}, June 2021, pp.
  2083--2092.

\bibitem{Qiao2021_MicroCACTI}
M.~Qiao, X.~Liu, and X.~Yuan, ``Snapshot temporal compressive microscopy using
  an iterative algorithm with untrained neural networks,'' \emph{Optics
  Letters}, vol.~46, no.~8, pp. 1888--1891, Apr 2021.

\bibitem{Meng2021_ICCV_self}
Z.~Meng, Z.~Yu, K.~Xu, and X.~Yuan, ``Self-supervised neural networks for
  spectral snapshot compressive imaging,'' in \emph{IEEE Conference on Computer
  Vision}, October 2021.

\bibitem{Miao19ICCV}
X.~Miao, X.~Yuan, Y.~Pu, and V.~Athitsos, ``$\lambda$-net: Reconstruct
  hyperspectral images from a snapshot measurement,'' in \emph{IEEE Conference
  on Computer Vision}, 2019.

\bibitem{Meng20ECCV_TSAnet}
Z.~Meng, J.~Ma, and X.~Yuan, ``End-to-end low cost compressive spectral imaging
  with spatial-spectral self-attention,'' in \emph{European Conference on
  Computer Vision}, 2020.

\bibitem{Huang2021_CVPR_GSMSCI}
T.~Huang, W.~Dong, X.~Yuan, J.~Wu, and G.~Shi, ``Deep gaussian scale mixture
  prior for spectral compressive imaging,'' in \emph{IEEE Conference on
  Computer Vision and Pattern Recognition}, June 2021, pp. 16\,216--16\,225.

\bibitem{jeon2019compact}
D.~S. Jeon, S.-H. Baek, S.~Yi, Q.~Fu, X.~Dun, W.~Heidrich, and M.~H. Kim,
  ``{Compact snapshot hyperspectral imaging with diffracted rotation},''
  \emph{ACM Transactions on Graphics}, vol.~38, no.~4, pp. 1--13, 2019.

\bibitem{gedalin2019deepcubenet}
D.~Gedalin, Y.~Oiknine, and A.~Stern, ``{DeepCubeNet}: Reconstruction of
  spectrally compressive sensed hyperspectral images with deep neural
  networks,'' \emph{Optics Express}, vol.~27, no.~24, pp. 35\,811--35\,822,
  2019.

\bibitem{van2018compressed}
D.~Van~Veen, A.~Jalal, M.~Soltanolkotabi, E.~Price, S.~Vishwanath, and A.~G.
  Dimakis, ``Compressed sensing with deep image prior and learned
  regularization,'' \emph{arXiv preprint arXiv:1806.06438}, 2018.

\bibitem{Li2021_ICCV_Deblur}
X.~Li, J.~Suo, W.~Zhang, X.~Yuan, and Q.~Dai, ``Universal and flexible optical
  aberration correction using deep-prior based deconvolution,'' in \emph{IEEE
  Conference on Computer Vision}, October 2021.

\bibitem{wang2021deep}
C.~Wang, Q.~Huang, M.~Cheng, Z.~Ma, and D.~J. Brady, ``Deep learning for camera
  autofocus,'' \emph{IEEE Transactions on Computational Imaging}, vol.~7, pp.
  258--271, 2021.

\bibitem{iliadis2016deepbinarymask}
M.~Iliadis, L.~Spinoulas, and A.~K. Katsaggelos, ``Deepbinarymask: Learning a
  binary mask for video compressive sensing,'' \emph{arXiv preprint
  arXiv:1607.03343}, 2016.

\bibitem{spinoulas2015sampling}
L.~Spinoulas, O.~Cossairt, and A.~K. Katsaggelos, ``Sampling optimization for
  on-chip compressive video,'' in \emph{IEEE International Conference on Image
  Processing}, 2015, pp. 3329--3333.

\bibitem{lindell2018single}
M.~D.B.Lindell and G.Wetzstein, ``Single-photon {3D} imaging with deep sensor
  fusion.'' \emph{ACM Transactions on Graphics}, vol.~37, no.~4, pp. 113--1,
  2018.

\bibitem{khan2020flatnet}
S.S.Khan, V.Sundar, V.Boominathan, A.Veeraraghavan, and K.Mitra, ``Flatnet:
  Towards photorealistic scene reconstruction from lensless measurements,''
  \emph{IEEE Transactions on Pattern Analysis and Machine Intelligence}, 2020.

\bibitem{Yuan18OE}
X.~Yuan and Y.~Pu, ``Parallel lensless compressive imaging via deep
  convolutional neural networks,'' \emph{Optics Express}, vol.~26, no.~2, pp.
  1962--1977, Jan 2018.

\bibitem{horstmeyer2017convolutional}
R.~Horstmeyer, R.~Y. Chen, B.~Kappes, and B.~Judkewitz, ``Convolutional neural
  networks that teach microscopes how to image,'' \emph{arXiv preprint
  arXiv:1709.07223}, 2017.

\bibitem{sitzmann2018end}
V.~Sitzmann, S.~Diamond, Y.~Peng, X.~Dun, S.~Boyd, W.~Heidrich, F.~Heide, and
  G.~Wetzstein, ``End-to-end optimization of optics and image processing for
  achromatic extended depth of field and super-resolution imaging,'' \emph{ACM
  Transactions on Graphics}, vol.~37, no.~4, pp. 1--13, 2018.

\bibitem{dun2019joint}
X.~Dun, Z.~Wang, and Y.~Peng, ``{ Joint-designed achromatic diffractive optics
  for full-spectrum computational imaging},'' in \emph{Optoelectronic Imaging
  and Multimedia Technology VI}, vol. 11187, 2019, p. 111870I.

\bibitem{han2011novel}
J.-W. Han, J.-H. Kim, H.-T. Lee, and S.-J. Ko, ``A novel training based
  auto-focus for mobile-phone cameras,'' \emph{IEEE Transactions on Consumer
  Electronics}, vol.~57, no.~1, pp. 232--238, 2011.

\bibitem{Bergman:2020:DeepLiDAR}
A.~W. Bergman, D.~B. Lindell, and G.~Wetzstein, ``{Deep adaptive LiDAR:
  End-to-end optimization of sampling and depth completion at low sampling
  rates},'' \emph{IEEE International Conference on Computational Photography},
  2020.

\bibitem{peng2019learned}
Y.~Peng, Q.~Sun, X.~Dun, G.~Wetzstein, W.~Heidrich, and F.~Heide, ``{Learned
  large field-of-view imaging with thin-plate optics},'' \emph{ACM Transactions
  on Graphics}, vol.~38, no.~6, p. 219, 2019.

\bibitem{banerji2019diffractive}
S.~Banerji, M.~Meem, A.~Majumder, B.~Sensale-Rodriguez, and R.~Menon,
  ``{Diffractive flat lens enables extreme depth-of-focus imaging},''
  \emph{arXiv preprint arXiv:1910.07928}, 2019.

\bibitem{shedligeri2017data}
P.~A. Shedligeri, S.~Mohan, and K.~Mitra, ``Data driven coded aperture design
  for depth recovery,'' in \emph{International Conference on Image Processing},
  2017, pp. 56--60.

\bibitem{wang2019measurement}
H.~Wang, Q.~Yan, B.~Li, C.~Yuan, and Y.~Wang, ``{Measurement matrix
  construction for large-area single photon compressive imaging},''
  \emph{Sensors}, vol.~19, no.~3, p. 474, 2019.

\bibitem{wang2018hyperreconnet}
L.~Wang, T.~Zhang, Y.~Fu, and H.~Huang, ``Hyperreconnet: Joint coded aperture
  optimization and image reconstruction for compressive hyperspectral
  imaging,'' \emph{IEEE Transactions on Image Processing}, vol.~28, no.~5, pp.
  2257--2270, 2018.

\bibitem{zhang2020optimization}
J.~Zhang, C.~Zhao, and W.~Gao, ``{Optimization-inspired compact deep
  compressive sensing},'' \emph{IEEE Journal of Selected Topics in Signal
  Processing}, 2020.

\bibitem{inagaki2018learning}
Y.~Inagaki, Y.~Kobayashi, K.~Takahashi, T.~Fujii, and H.~Nagahara, ``Learning
  to capture light fields through a coded aperture camera,'' in \emph{European
  Conference on Computer Vision}, 2018, pp. 418--434.

\bibitem{akpinar2019learning}
U.~Akpinar, E.~Sahin, and A.~Gotchev, ``{Learning optimal phase-coded aperture
  for depth of field extension},'' in \emph{International Conference on Image
  Processing}, 2019, pp. 4315--4319.

\bibitem{rao2013context}
S.~Rao, K.-Y. Ni, and Y.~Owechko, ``Context and task-aware knowledge-enhanced
  compressive imaging,'' in \emph{Unconventional Imaging and Wavefront
  Sensing}, vol. 8877, 2013, p. 88770E.

\bibitem{ashok2008compressive}
A.~Ashok, P.~K. Baheti, and M.~A. Neifeld, ``Compressive imaging system design
  using task-specific information,'' \emph{Applied Optics}, vol.~47, no.~25,
  pp. 4457--4471, 2008.

\bibitem{abetamann2013compressive}
M.~A$\beta$mann and M.~Bayer, ``Compressive adaptive computational ghost
  imaging,'' \emph{Scientific Reports}, vol.~3, no.~1, pp. 1--5, 2013.

\bibitem{ashok2008task}
A.~Ashok, ``A task-specific approach to computational imaging system design,''
  Ph.D. dissertation, The University of Arizona, 2008.

\bibitem{rawat2015context}
Y.~S. Rawat and M.~S. Kankanhalli, ``Context-aware photography learning for
  smart mobile devices,'' \emph{ACM Transactions on Multimedia Computing,
  Communications, and Applications}, vol.~12, no.~1s, pp. 1--24, 2015.

\bibitem{rawat2016clicksmart}
------, ``Clicksmart: A context-aware viewpoint recommendation system for
  mobile photography,'' \emph{IEEE Transactions on Circuits and Systems for
  Video Technology}, vol.~27, no.~1, pp. 149--158, 2016.

\bibitem{liu2012texture}
L.~Liu and P.~Fieguth, ``Texture classification from random features,''
  \emph{IEEE Transactions on Pattern Analysis and Machine Intelligence},
  vol.~34, no.~3, pp. 574--586, 2012.

\bibitem{liu2016median}
L.~Liu, S.~Lao, P.~W. Fieguth, Y.~Guo, X.~Wang, and M.~Pietik{\"a}inen,
  ``Median robust extended local binary pattern for texture classification,''
  \emph{IEEE Transactions on Image Processing}, vol.~25, no.~3, pp. 1368--1381,
  2016.

\bibitem{bian2016robust}
X.~Bian, C.~Chen, Y.~Xu, and Q.~Du, ``Robust hyperspectral image classification
  by multi-layer spatial-spectral sparse representations,'' \emph{Remote
  Sensing}, vol.~8, no.~12, p. 985, 2016.

\bibitem{zhang2014fast}
K.~Zhang, L.~Zhang, and M.-H. Yang, ``Fast compressive tracking,'' \emph{IEEE
  Transactions on Pattern Analysis and Machine Intelligence}, vol.~36, no.~10,
  pp. 2002--2015, 2014.

\bibitem{saragadam2021sassi}
V.Saragadam, M.DeZeeuw, R.G.Baraniuk, A.Veeraraghavan, and A.Sankaranarayanan,
  ``Sassi---super-pixelated adaptive spatio-spectral imaging,'' \emph{IEEE
  Transactions on Pattern Analysis and Machine Intelligence}, 2021.

\bibitem{gupta2010flexible}
M.~Gupta, A.~Agrawal, A.~Veeraraghavan, and S.~G. Narasimhan, ``Flexible voxels
  for motion-aware videography,'' in \emph{European Conference on Computer
  Vision}, 2010, pp. 100--114.

\bibitem{wang2018megapixel}
X.~C.Wang, Q.Fu and W.Heidrich, ``Megapixel adaptive optics: towards correcting
  large-scale distortions in computational cameras,'' \emph{ACM Transactions on
  Graphics}, vol.~37, no.~4, pp. 1--12, 2018.

\bibitem{Lu20SEC}
S.~Lu, X.~Yuan, and W.~Shi, ``An integrated framework for compressive imaging
  processing on {CAVs},'' in \emph{ACM/IEEE Symposium on Edge Computing},
  November 2020.

\bibitem{shi2016edge}
W.~Shi, J.~Cao, Q.~Zhang, Y.~Li, and L.~Xu, ``Edge computing: Vision and
  challenges,'' \emph{IEEE Internet of Things Journal}, vol.~3, no.~5, pp.
  637--646, 2016.

\bibitem{liu2019edge}
S.~Liu, L.~Liu, J.~Tang, B.~Yu, Y.~Wang, and W.~Shi, ``Edge computing for
  autonomous driving: Opportunities and challenges,'' \emph{IEEE}, vol. 107,
  no.~8, pp. 1697--1716, 2019.

\bibitem{girshick2015region}
R.~Girshick, J.~Donahue, T.~Darrell, and J.~Malik, ``Region-based convolutional
  networks for accurate object detection and segmentation,'' \emph{IEEE
  Transactions on Pattern Analysis and Machine Intelligence}, vol.~38, no.~1,
  pp. 142--158, 2015.

\bibitem{girshick2015fast}
R.~Girshick, ``Fast {R-CNN},'' in \emph{IEEE International Conference on
  Computer Vision}, 2015, pp. 1440--1448.

\bibitem{ren2015faster}
S.~Ren, K.~He, R.~Girshick, and J.~Sun, ``Faster {R-CNN}: Towards real-time
  object detection with region proposal networks,'' in \emph{Advances in Neural
  Information Processing Systems}, 2015, pp. 91--99.

\bibitem{howard2017mobilenets}
A.~G. Howard, M.~Zhu, B.~Chen, D.~Kalenichenko, W.~Wang, T.~Weyand,
  M.~Andreetto, and H.~Adam, ``Mobilenets: Efficient convolutional neural
  networks for mobile vision applications,'' \emph{arXiv preprint
  arXiv:1704.04861}, 2017.

\bibitem{zhuinverted}
M.~S. A. H.~M. Zhu and A.~Z. L.-C. Chen, ``Inverted residuals and linear
  bottlenecks: Mobile networks for classification, detection and
  segmentation,'' \emph{arXiv preprint arXiv:1801.04381}, 2018.

\bibitem{howard2019searching}
A.~Howard, M.~Sandler, G.~Chu, L.-C. Chen, B.~Chen, M.~Tan, W.~Wang, Y.~Zhu,
  R.~Pang, V.~Vasudevan, Q.~V. Le, and H.~Adam, ``Searching for mobilenetv3,''
  in \emph{International Conference on Computer Vision}, 2019, pp. 1314--1324.

\bibitem{zhang2018shufflenet}
X.~Zhang, X.~Zhou, M.~Lin, and J.~Sun, ``Shufflenet: An extremely efficient
  convolutional neural network for mobile devices,'' in \emph{IEEE Conference
  on Computer Vision and Pattern Recognition}, 2018, pp. 6848--6856.

\bibitem{ma2018shufflenet}
N.~Ma, X.~Zhang, H.-T. Zheng, and J.~Sun, ``Shufflenet v2: Practical guidelines
  for efficient cnn architecture design,'' in \emph{European Conference on
  Computer Vision}, 2018, pp. 116--131.

\bibitem{redmon2016you}
J.~Redmon, S.~Divvala, R.~Girshick, and A.~Farhadi, ``You only look once:
  Unified, real-time object detection,'' in \emph{IEEE Conference on Computer
  Vision and Pattern Recognition}, 2016, pp. 779--788.

\bibitem{redmon2017yolo9000}
J.~Redmon and A.~Farhadi, ``{YOLO}9000: Better, faster, stronger,'' in
  \emph{IEEE Conference on Computer Vision and Pattern Recognition}, 2017, pp.
  7263--7271.

\bibitem{redmon2018yolov3}
------, ``{YOLO}v3: An incremental improvement,'' \emph{arXiv preprint
  arXiv:1804.02767}, 2018.

\bibitem{lin2017feature}
T.-Y. Lin, P.~Doll{\'a}r, R.~Girshick, K.~He, B.~Hariharan, and S.~Belongie,
  ``Feature pyramid networks for object detection,'' in \emph{IEEE Conference
  on Computer Vision and Pattern Recognition}, 2017, pp. 2117--2125.

\bibitem{long2015fully}
J.~Long, E.~Shelhamer, and T.~Darrell, ``Fully convolutional networks for
  semantic segmentation,'' in \emph{IEEE Conference on Computer Vision and
  Pattern Recognition}, 2015, pp. 3431--3440.

\bibitem{mao2017survey}
Y.~Mao, C.~You, J.~Zhang, K.~Huang, and K.~B. Letaief, ``A survey on mobile
  edge computing: The communication perspective,'' \emph{IEEE Communications
  Surveys \& Tutorials}, vol.~19, no.~4, pp. 2322--2358, 2017.

\bibitem{zhang2009intelligent}
L.~Zhang, S.~Malki, and L.~Spaanenburg, ``Intelligent camera cloud computing,''
  in \emph{International Symposium on Circuits and Systems}, 2009, pp.
  1209--1212.

\bibitem{mitra2014toward}
K.~Mitra, A.~Veeraraghavan, A.~C. Sankaranarayanan, and R.~G. Baraniuk,
  ``Toward compressive camera networks,'' \emph{Computer}, vol.~47, no.~5, pp.
  52--59, 2014.

\bibitem{bistry2010cloud}
H.~Bistry and J.~Zhang, ``A cloud computing approach to complex robot vision
  tasks using smart camera systems,'' in \emph{IEEE/RSJ International
  Conference on Intelligent Robots and Systems}, 2010, pp. 3195--3200.

\bibitem{ahmad2016role}
I.~Ahmad, R.~M. Noor, I.~Ali, and M.~A. Qureshi, ``The role of vehicular cloud
  computing in road traffic management: A survey,'' in \emph{International
  Conference on Future Intelligent Vehicular Technologies}, 2016, pp. 123--131.

\bibitem{kober2013reinforcement}
J.~Kober, J.~A. Bagnell, and J.~Peters, ``Reinforcement learning in robotics: A
  survey,'' \emph{The International Journal of Robotics Research}, vol.~32,
  no.~11, pp. 1238--1274, 2013.

\bibitem{kormushev2013reinforcement}
P.~Kormushev, S.~Calinon, and D.~G. Caldwell, ``Reinforcement learning in
  robotics: Applications and real-world challenges,'' \emph{Robotics}, vol.~2,
  no.~3, pp. 122--148, 2013.

\bibitem{peters2008natural}
J.~Peters and S.~Schaal, ``Natural actor-critic,'' \emph{Neurocomputing},
  vol.~71, no. 7-9, pp. 1180--1190, 2008.

\bibitem{kober2009learning}
J.~Kober and J.~Peters, ``Learning motor primitives for robotics,'' in
  \emph{IEEE International Conference on Robotics and Automation}, 2009, pp.
  2112--2118.

\bibitem{theodorou2010generalized}
E.~Theodorou, J.~Buchli, and S.~Schaal, ``A generalized path integral control
  approach to reinforcement learning,'' \emph{The Journal of Machine Learning
  Research}, vol.~11, no. 104, pp. 3137--3181, 2010.

\bibitem{shalev2016safe}
S.~Shalev-Shwartz, S.~Shammah, and A.~Shashua, ``Safe, multi-agent,
  reinforcement learning for autonomous driving,'' \emph{arXiv preprint
  arXiv:1610.03295}, 2016.

\bibitem{isele2018navigating}
D.~Isele, R.~Rahimi, A.~Cosgun, K.~Subramanian, and K.~Fujimura, ``Navigating
  occluded intersections with autonomous vehicles using deep reinforcement
  learning,'' in \emph{IEEE International Conference on Robotics and
  Automation}, 2018, pp. 2034--2039.

\bibitem{zhu2018human}
M.~Zhu, X.~Wang, and Y.~Wang, ``Human-like autonomous car-following model with
  deep reinforcement learning,'' \emph{Transportation Research Part C: Emerging
  Technologies}, vol.~97, pp. 348--368, 2018.

\bibitem{han2015mobile}
Q.~Han, S.~Liang, and H.~Zhang, ``Mobile cloud sensing, big data, and 5{G}
  networks make an intelligent and smart world,'' \emph{IEEE Network}, vol.~29,
  no.~2, pp. 40--45, 2015.

\bibitem{yue2013cloud}
H.~Yue, X.~Sun, J.~Yang, and F.~Wu, ``Cloud-based image coding for mobile
  devices—toward thousands to one compression,'' \emph{IEEE Transactions on
  Multimedia}, vol.~15, no.~4, pp. 845--857, 2013.

\bibitem{zheng2016big}
K.~Zheng, Z.~Yang, K.~Zhang, P.~Chatzimisios, K.~Yang, and W.~Xiang, ``Big
  data-driven optimization for mobile networks toward 5{G},'' \emph{IEEE
  Network}, vol.~30, no.~1, pp. 44--51, 2016.

\bibitem{jawhar2017communication}
I.~Jawhar, N.~Mohamed, J.~Al-Jaroodi, D.~P. Agrawal, and S.~Zhang,
  ``Communication and networking of uav-based systems: Classification and
  associated architectures,'' \emph{Journal of Network and Computer
  Applications}, vol.~84, pp. 93--108, 2017.

\bibitem{tsouros2019review}
D.~C. Tsouros, S.~Bibi, and P.~G. Sarigiannidis, ``A review on {UAV}-based
  applications for precision agriculture,'' \emph{Information}, vol.~10,
  no.~11, p. 349, 2019.

\bibitem{adao2017hyperspectral}
T.~Ad{\~a}o, J.~Hru{\v{s}}ka, L.~P{\'a}dua, J.~Bessa, E.~Peres, R.~Morais, and
  J.~J. Sousa, ``Hyperspectral imaging: A review on uav-based sensors, data
  processing and applications for agriculture and forestry,'' \emph{Remote
  Sensing}, vol.~9, no.~11, p. 1110, 2017.

\bibitem{niethammer2012uav}
U.~Niethammer, M.~James, S.~Rothmund, J.~Travelletti, and M.~Joswig,
  ``Uav-based remote sensing of the super-sauze landslide: Evaluation and
  results,'' \emph{Engineering Geology}, vol. 128, pp. 2--11, 2012.

\bibitem{opromolla2019airborne}
R.~Opromolla, G.~Inchingolo, and G.~Fasano, ``Airborne visual detection and
  tracking of cooperative uavs exploiting deep learning,'' \emph{Sensors},
  vol.~19, no.~19, p. 4332, 2019.

\bibitem{sampedro2018image}
C.~Sampedro, A.~Rodriguez-Ramos, I.~Gil, L.~Mejias, and P.~Campoy,
  ``Image-based visual servoing controller for multirotor aerial robots using
  deep reinforcement learning,'' in \emph{IEEE/RSJ International Conference on
  Intelligent Robots and Systems}, 2018, pp. 979--986.

\bibitem{sridhar2016target}
V.~Sridhar and A.~Manikas, ``Target tracking with a flexible {UAV} cluster
  array,'' in \emph{IEEE Globecom Workshops}, 2016, pp. 1--6.

\bibitem{dantu2011programming}
K.~Dantu, B.~Kate, J.~Waterman, P.~Bailis, and M.~Welsh, ``Programming
  micro-aerial vehicle swarms with karma,'' in \emph{9th ACM Conference on
  Embedded Networked Sensor Systems}, 2011, pp. 121--134.

\bibitem{duan2020dynamic}
T.~Duan, W.~Wang, T.~Wang, X.~Chen, and X.~Li, ``Dynamic tasks scheduling model
  of {UAV} cluster based on flexible network architecture,'' \emph{IEEE
  Access}, vol.~8, pp. 115\,448--115\,460, 2020.

\bibitem{jin2020uav}
Y.~Jin, Z.~Qian, and W.~Yang, ``{UAV} cluster-based video surveillance system
  optimization in heterogeneous communication of smart cities,'' \emph{IEEE
  Access}, vol.~8, pp. 55\,654--55\,664, 2020.

\bibitem{dai2019deploy}
H.~Dai, H.~Zhang, M.~Hua, C.~Li, Y.~Huang, and B.~Wang, ``How to deploy
  multiple uavs for providing communication service in an unknown region?''
  \emph{IEEE Wireless Communications Letters}, vol.~8, no.~4, pp. 1276--1279,
  2019.

\bibitem{tanil2013collaborative}
{\c{C}}.~Tanil, C.~Warty, and E.~Obiedat, ``Collaborative mission planning for
  {UAV} cluster to optimize relay distance,'' in \emph{IEEE Aerospace
  Conference}, 2013, pp. 1--11.

\bibitem{shima2006multiple}
T.~Shima, S.~J. Rasmussen, A.~G. Sparks, and K.~M. Passino, ``Multiple task
  assignments for cooperating uninhabited aerial vehicles using genetic
  algorithms,'' \emph{Computers \& Operations Research}, vol.~33, no.~11, pp.
  3252--3269, 2006.

\bibitem{sharma2009collision}
R.~Sharma and D.~Ghose, ``Collision avoidance between {UAV} clusters using
  swarm intelligence techniques,'' \emph{International Journal of Systems
  Science}, vol.~40, no.~5, pp. 521--538, 2009.

\bibitem{clark2017autonomous}
R.~A. Clark, G.~Punzo, C.~N. MacLeod, G.~Dobie, R.~Summan, G.~Bolton, S.~G.
  Pierce, and M.~Macdonald, ``Autonomous and scalable control for remote
  inspection with multiple aerial vehicles,'' \emph{Robotics and Autonomous
  Systems}, vol.~87, pp. 258--268, 2017.

\bibitem{saska2014autonomous}
M.~Saska, J.~Chudoba, L.~P{\v{r}}eu{\v{c}}il, J.~Thomas, G.~Loianno,
  A.~T{\v{r}}e{\v{s}}{\v{n}}{\'a}k, V.~Von{\'a}sek, and V.~Kumar, ``Autonomous
  deployment of swarms of micro-aerial vehicles in cooperative surveillance,''
  in \emph{International Conference on Unmanned Aircraft Systems}, 2014, pp.
  584--595.

\bibitem{saska2016swarm}
M.~Saska, V.~Von{\'a}sek, J.~Chudoba, J.~Thomas, G.~Loianno, and V.~Kumar,
  ``Swarm distribution and deployment for cooperative surveillance by
  micro-aerial vehicles,'' \emph{Journal of Intelligent \& Robotic Systems},
  vol.~84, no.~1, pp. 469--492, 2016.

\bibitem{sanchez2020semantic}
J.~L. Sanchez-Lopez, M.~Castillo-Lopez, and H.~Voos, ``Semantic situation
  awareness of ellipse shapes via deep learning for multirotor aerial robots
  with a 2d lidar,'' in \emph{International Conference on Unmanned Aircraft
  Systems}, 2020, pp. 1014--1023.

\bibitem{sanchez2019deep}
J.~L. Sanchez-Lopez, C.~Sampedro, D.~Cazzato, and H.~Voos, ``Deep learning
  based semantic situation awareness system for multirotor aerial robots using
  lidar,'' in \emph{International Conference on Unmanned Aircraft Systems},
  2019, pp. 899--908.

\bibitem{saska2014swarms}
M.~Saska, J.~Vakula, and L.~P{\v{r}}eu{\'c}il, ``Swarms of micro aerial
  vehicles stabilized under a visual relative localization,'' in \emph{IEEE
  International Conference on Robotics and Automation}, 2014, pp. 3570--3575.

\bibitem{schleich2013uav}
J.~Schleich, A.~Panchapakesan, G.~Danoy, and P.~Bouvry, ``Uav fleet area
  coverage with network connectivity constraint,'' in \emph{11th ACM
  international symposium on Mobility management and wireless access}, 2013,
  pp. 131--138.

\bibitem{messous2016network}
M.-A. Messous, S.-M. Senouci, and H.~Sedjelmaci, ``Network connectivity and
  area coverage for uav fleet mobility model with energy constraint,'' in
  \emph{IEEE Wireless Communications and Networking Conference}, 2016, pp.
  1--6.

\bibitem{sanchez2019real}
J.~L. Sanchez~Lopez, M.~Wang, M.~A. Olivares~Mendez, M.~Molina, and H.~Voos,
  ``A real-time 3d path planning solution for collision-free navigation of
  multirotor aerial robots in dynamic environments,'' \emph{Journal of
  Intelligent and Robotic Systems}, vol.~93, no. 1-2, pp. 33--53, 2019.

\bibitem{sampedro2018laser}
C.~Sampedro, H.~Bavle, A.~Rodriguez-Ramos, P.~De~La~Puente, and P.~Campoy,
  ``Laser-based reactive navigation for multirotor aerial robots using deep
  reinforcement learning,'' in \emph{IEEE/RSJ International Conference on
  Intelligent Robots and Systems}, 2018, pp. 1024--1031.

\bibitem{atten2016uav}
C.~Atten, L.~Channouf, G.~Danoy, and P.~Bouvry, ``Uav fleet mobility model with
  multiple pheromones for tracking moving observation targets,'' in
  \emph{European Conference on the Applications of Evolutionary Computation},
  2016, pp. 332--347.

\bibitem{yang2019application}
J.~Yang, X.~You, G.~Wu, M.~M. Hassan, A.~Almogren, and J.~Guna, ``Application
  of reinforcement learning in {UAV} cluster task scheduling,'' \emph{Future
  generation computer systems}, vol.~95, pp. 140--148, 2019.

\bibitem{wang2019reinforcement}
T.~Wang, R.~Qin, Y.~Chen, H.~Snoussi, and C.~Choi, ``A reinforcement learning
  approach for uav target searching and tracking,'' \emph{Multimedia Tools and
  Applications}, vol.~78, no.~4, pp. 4347--4364, 2019.

\bibitem{wu2019couav}
W.~Wu, Z.~Huang, F.~Shan, Y.~Bian, K.~Lu, Z.~Li, and J.~Wang, ``Couav: a
  cooperative uav fleet control and monitoring platform,'' \emph{arXiv preprint
  arXiv:1904.04046}, 2019.

\bibitem{searle1980minds}
J.~R. Searle, ``Minds, brains, and programs,'' \emph{Behavioral and Brain
  Sciences}, vol.~3, no.~3, pp. 417--424, 1980.

\bibitem{frintrop2010computational}
S.~Frintrop, E.~Rome, and H.~I. Christensen, ``Computational visual attention
  systems and their cognitive foundations: A survey,'' \emph{{Association for
  Computing Machinery} Transactions on Applied Perception}, vol.~7, no.~1, pp.
  1--39, 2010.

\bibitem{barua2016direct}
S.~Barua, Y.~Miyatani, and A.~Veeraraghavan, ``Direct face detection and video
  reconstruction from event cameras,'' in \emph{IEEE Winter Conference on
  Applications of Computer Vision}, 2016, pp. 1--9.

\bibitem{kim2014simultaneous}
H.~Kim, A.~Handa, R.~Benosman, S.~H. Ieng, and A.~J. Davison, ``Simultaneous
  mosaicing and tracking with an event camera,'' in \emph{British Machine
  Vision Conference}, 2014.

\bibitem{mueggler2017event}
E.~Mueggler, H.~Rebecq, G.~Gallego, T.~Delbruck, and D.~Scaramuzza, ``The
  event-camera dataset and simulator: Event-based data for pose estimation,
  visual odometry, and {SLAM},'' \emph{The International Journal of Robotics
  Research}, vol.~36, no.~2, pp. 142--149, 2017.

\bibitem{kim2016real}
H.~Kim, S.~Leutenegger, and A.~J. Davison, ``Real-time {3D} reconstruction and
  {6-DoF} tracking with an event camera,'' in \emph{European Conference on
  Computer Vision}, 2016, pp. 349--364.

\bibitem{medathati2016bio}
N.~K. Medathati, H.~Neumann, G.~S. Masson, and P.~Kornprobst, ``Bio-inspired
  computer vision: Towards a synergistic approach of artificial and biological
  vision,'' \emph{Computer Vision and Image Understanding}, vol. 150, pp.
  1--30, 2016.

\bibitem{hubel1998early}
D.~H. Hubel and T.~N. Wiesel, ``Early exploration of the visual cortex,''
  \emph{Neuron}, vol.~20, no.~3, pp. 401--412, 1998.

\bibitem{kumar2010spiking}
A.~Kumar, S.~Rotter, and A.~Aertsen, ``Spiking activity propagation in neuronal
  networks: Reconciling different perspectives on neural coding,'' \emph{Nature
  Reviews Neuroscience}, vol.~11, no.~9, pp. 615--627, 2010.

\bibitem{kriegeskorte2015deep}
N.~Kriegeskorte, ``Deep neural networks: A new framework for modeling
  biological vision and brain information processing,'' \emph{Annual Review of
  Vision Science}, vol.~1, no.~1, pp. 417--446, 2015.

\bibitem{lillicrap2020backpropagation}
T.~P. Lillicrap, A.~Santoro, L.~Marris, C.~J. Akerman, and G.~Hinton,
  ``Backpropagation and the brain,'' \emph{Nature Reviews Neuroscience},
  vol.~21, no.~6, p. 335–346, 2020.

\bibitem{hennessy2019new}
J.~L. Hennessy and D.~A. Patterson, ``A new golden age for computer
  architecture,'' \emph{Communications of the {Association for Computing
  M}achinery}, vol.~62, no.~2, pp. 48--60, 2019.

\bibitem{psaltis1988adaptive}
D.~Psaltis, D.~Brady, and K.~Wagner, ``Adaptive optical networks using
  photorefractive crystals,'' \emph{Applied Optics}, vol.~27, no.~9, pp.
  1752--1759, 1988.

\bibitem{chang2018hybrid}
J.~Chang, V.~Sitzmann, X.~Dun, W.~Heidrich, and G.~Wetzstein, ``Hybrid
  optical-electronic convolutional neural networks with optimized diffractive
  optics for image classification,'' \emph{Scientific Reports}, vol.~8, no.~1,
  pp. 1--10, 2018.

\bibitem{feldmann2019all}
J.~Feldmann, N.~Youngblood, C.~D. Wright, H.~Bhaskaran, and W.~Pernice,
  ``All-optical spiking neurosynaptic networks with self-learning
  capabilities,'' \emph{Nature}, vol. 569, no. 7755, pp. 208--214, 2019.

\bibitem{pei2019towards}
J.~Pei, L.~Deng, S.~Song, M.~Zhao, Y.~Zhang, S.~Wu, G.~Wang, Z.~Zou, Z.~Wu,
  W.~He, F.~Chen, N.~Deng, S.~Wu, Y.~Wang, Y.~Wu, Z.~Yang, C.~Ma, G.~Li,
  W.~Han, H.~Li, H.~Wu, R.~Zhao, Y.~Xie, and L.~Shi, ``Towards artificial
  general intelligence with hybrid tianjic chip architecture,'' \emph{Nature},
  vol. 572, no. 7767, pp. 106--111, 2019.

\bibitem{gibson1999two}
J.~R. Gibson, M.~Beierlein, and B.~W. Connors, ``Two networks of electrically
  coupled inhibitory neurons in neocortex,'' \emph{Nature}, vol. 402, no. 6757,
  pp. 75--79, 1999.

\bibitem{hodgkin1952currents}
A.~L. Hodgkin and A.~F. Huxley, ``Currents carried by sodium and potassium ions
  through the membrane of the giant axon of {L}oligo,'' \emph{The Journal of
  Physiology}, vol. 116, no.~4, pp. 449--472, 1952.

\bibitem{kildishev2013planar}
A.~V. Kildishev, A.~Boltasseva, and V.~M. Shalaev, ``Planar photonics with
  metasurfaces,'' \emph{Science}, vol. 339, no. 6125, p. 1232009, 2013.

\bibitem{ni2013metasurface}
X.~Ni, A.~V. Kildishev, and V.~M. Shalaev, ``Metasurface holograms for visible
  light,'' \emph{Nature Communications}, vol.~4, no.~1, pp. 1--6, 2013.

\bibitem{zheng2015metasurface}
G.~Zheng, H.~M{\"u}hlenbernd, M.~Kenney, G.~Li, T.~Zentgraf, and S.~Zhang,
  ``Metasurface holograms reaching 80\% efficiency,'' \emph{Nature
  Nanotechnology}, vol.~10, no.~4, pp. 308--312, 2015.

\bibitem{lin2014dielectric}
D.~Lin, P.~Fan, E.~Hasman, and M.~L. Brongersma, ``Dielectric gradient
  metasurface optical elements,'' \emph{{Science}}, vol. 345, no. 6194, pp.
  298--302, 2014.

\bibitem{holsteen2019light}
A.~L. Holsteen, D.~Lin, I.~Kauvar, G.~Wetzstein, and M.~L. Brongersma, ``A
  light-field metasurface for high-resolution single-particle tracking,''
  \emph{Nano Letters}, vol.~19, no.~4, pp. 2267--2271, 2019.

\bibitem{ma2014acoustic}
G.~Ma, M.~Yang, S.~Xiao, Z.~Yang, and P.~Sheng, ``Acoustic metasurface with
  hybrid resonances,'' \emph{Nature Materials}, vol.~13, no.~9, pp. 873--878,
  2014.

\bibitem{capasso2017metasurfaces}
F.~Capasso, ``Metasurfaces: From quantum cascade lasers to flat optics,'' in
  \emph{International Conference on Infrared, Millimeter, and Terahertz Waves},
  2017, pp. 1--3.

\end{thebibliography}

\begin{IEEEbiography}[{\includegraphics[width=1in,height=1.25in,clip,keepaspectratio]{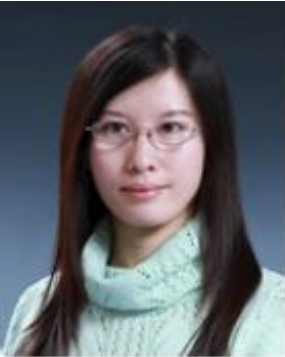}}]{Jinli Suo}
received the B.S. degree in computer science from Shandong University, Shandong, China, in 2004 and the Ph.D. degree from the Graduate University of Chinese Academy of Sciences, Beijing, China, in 2010. She is currently an Associate Professor with the Department of Automation, Tsinghua University, Beijing. Her current research interests include computer vision, computational photography, and statistical learning.
\end{IEEEbiography}

\begin{IEEEbiography}[{\includegraphics[width=1in,height=1.25in,clip,keepaspectratio]{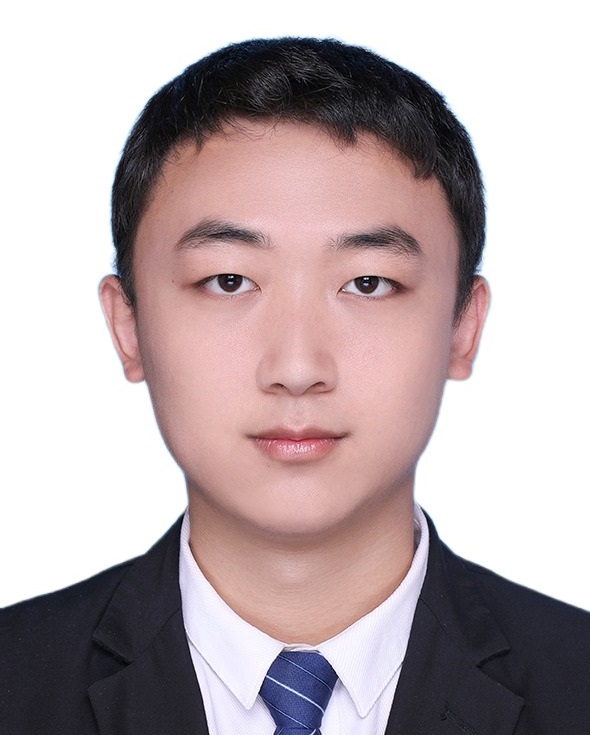}}]{Weihang Zhang}
received the BS degree in automation from Tsinghua University, Beijing, China, in 2019. He is currently working toward the Ph.D. degree in control theory and engineering in the Department of Automation, Tsinghua University, Beijing, China. His research interests include computational imaging and microscopy.
\end{IEEEbiography}

\begin{IEEEbiography}[{\includegraphics[width=1in,height=1.25in,clip,keepaspectratio]{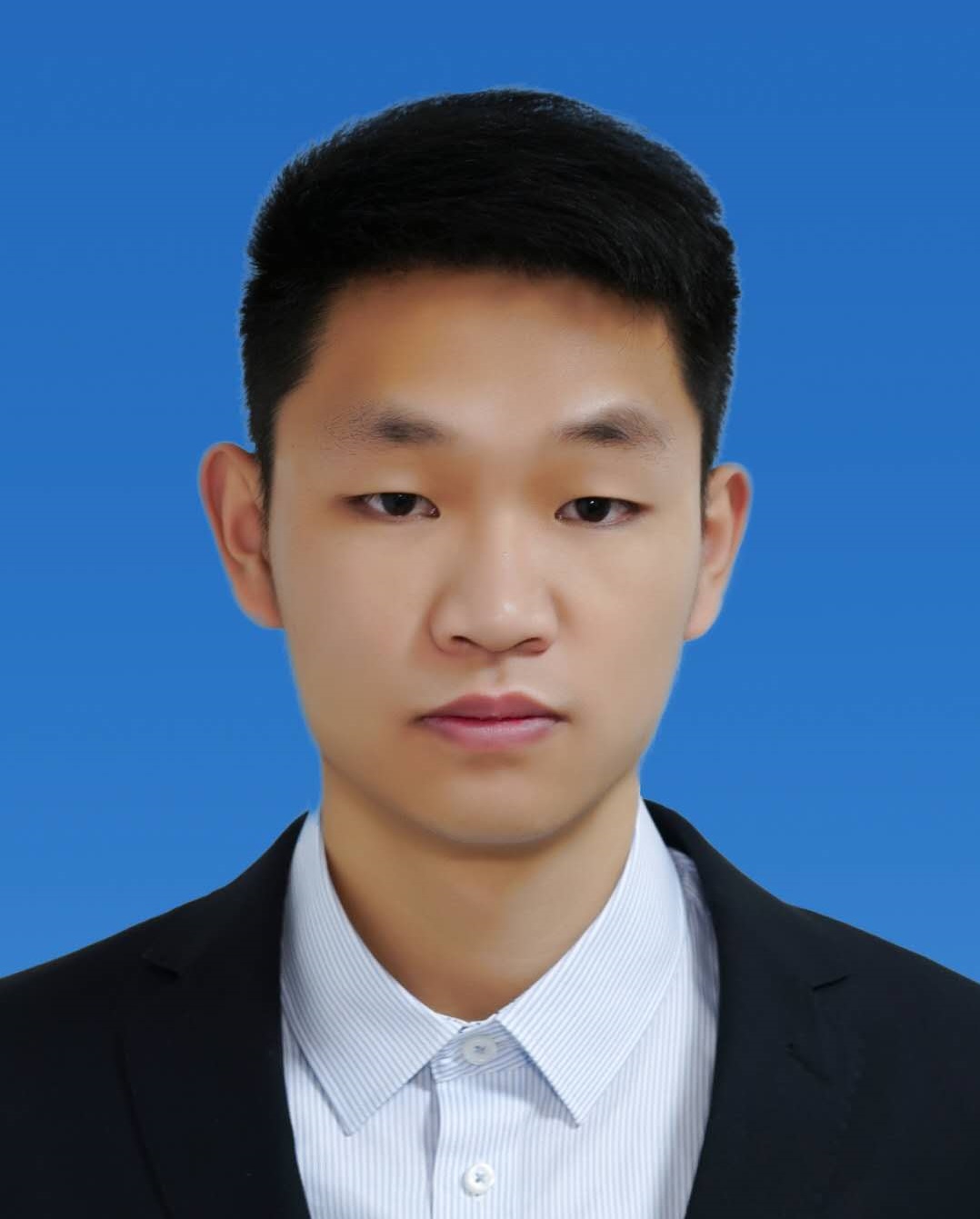}}]{Jin Gong}
received the BS degree in electronic information engineering from Xidian University, Shaanxi, China, in 2020. He is currently working towards the Ph.D. degree in control science and engineering in the Department of Automation, Tsinghua University, Beijing, China. His research interests include computational imaging.
\end{IEEEbiography}

\begin{IEEEbiography}[{\includegraphics[width=1in,height=1.25in,clip,keepaspectratio]{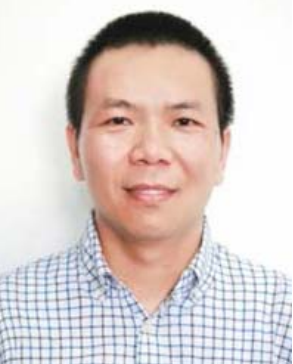}}]{Xin Yuan}
(SM’16) received the BEng and MEng degrees from Xidian University, in 2007 and 2009, respectively, and the PhD from the Hong Kong Polytechnic University, in 2012. He is currently a video analysis and coding lead researcher at Bell Labs, Murray Hill, NJ. Prior to this, he had been a post-doctoral associate with the Department of Electrical and Computer Engineering, Duke University, from 2012 to 2015, where he was working on compressive sensing and machine learning. 
He has been the Associate Editor of {\em Pattern Recognition} (2019-), {\em International Journal of Pattern Recognition and Artificial Intelligence} (2020-) and {\em Chinese Optics Letters} (2021-). He is leading the special issue of "Deep Learning for High Dimensional Sensing" in the {\em IEEE Journal of Selected Topics in Signal Processing} in 2021.
\end{IEEEbiography}

\begin{IEEEbiography}[{\includegraphics[width=1in,height=1.25in,clip,keepaspectratio]{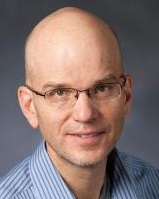}}]{David J. Brady}
(F’09) received the B.A. in physics and math from Macalester College, St Paul, MN, USA, in 1984, and the M.S. and Ph.D. degrees in applied physics from the California Institute of
Technology, Pasadena, CA, in 1986 and 1990, respectively. He was on the faculty of the University
of Illinois from 1990 until moving to Duke University, Durham, NC, USA, in 2001. He was a Professor of Electrical and Computer engineering at Duke University, Durham, NC, USA, and Duke Kunshan University, Kunshan, China, where he led the Duke Imaging and Spectroscopy Program (DISP), and the Camputer Lab, respectively from 2001 to 2020. He joins the Wyant College of Optical Sciences, University of Arizona as the J. W. and H. M. Goodman Endowed Chair Professor in 2021. He is the author of the book Optical Imaging and Spectroscopy (Hoboken, NJ, USA: Wiley). His research interests include computational imaging, gigapixel imaging, and compressive tomography. He is a Fellow of IEEE, OSA, and SPIE, and he won the 2013 SPIE Dennis Gabor Award for his work on compressive holography. 
\end{IEEEbiography}

\begin{IEEEbiography}[{\includegraphics[width=1in,height=1.25in,clip,keepaspectratio]{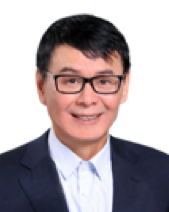}}]{Qionghai Dai}
(SM’05) received the ME and PhD degrees in computer science and automation from Northeastern University, Shenyang, China, in 1994 and 1996, respectively. He has been the faculty member of Tsinghua University since 1997. He is currently a professor with the Department of Automation, Tsinghua University, Beijing,
China, and an adjunct professor in the School of Life Science, Tsinghua University. His research areas
include computational photography and microscopy, computer vision and graphics, brain science, and video
communication. He is an associate editor of the JVCI, the IEEE TNNLS, and the IEEE TIP. He is an academician of the Chinese Academy of Engineering. He is a senior member of the IEEE.
\end{IEEEbiography}

\end{document}